\documentclass[pdflatex,sn-mathphys-num]{sn-jnl}% Math and Physical Sciences Numbered Reference Style
\usepackage{graphicx}%
\usepackage{multirow}%
\usepackage{amsmath,amssymb,amsfonts}%
\usepackage{amsthm}%
\usepackage{mathrsfs}%
\usepackage[title]{appendix}%
\usepackage[dvipsnames]{xcolor}%
\usepackage{textcomp}%
\usepackage{manyfoot}%
\usepackage{booktabs}%
\usepackage{algorithm}%
\usepackage{algorithmicx}%
\usepackage{algpseudocode}%
\usepackage{subcaption}
\usepackage{listings}%
\usepackage{longtable}
\usepackage{makecell}
\usepackage{array}
\usepackage{url}
\usepackage{caption}
\DeclareMathAlphabet{\pazocal}{OMS}{zplm}{m}{n}

\theoremstyle{thmstyleone}%
\theoremstyle{thmstyletwo}%

\theoremstyle{thmstylethree}%

\begin{document}

\newcommand{\Xp}{\pazocal{X}}

\title[Silhouette-Driven Instance-Weighted $k$-means]{Silhouette-Driven Instance-Weighted $k$-means}

\author*[1,2]{\fnm{Aggelos} \sur{Semoglou}}\email{a.semoglou@athenarc.gr}

\author[3]{\fnm{Aristidis} \sur{Likas}}\email{arly@cs.uoi.gr}

\author[1,2]{\fnm{John} \sur{Pavlopoulos}}\email{annis@aueb.gr}

\affil*[1]{
%\orgdiv{Dept. of Informatics},
\orgname{Athens University of Economics and Business}, \orgaddress{\country{Greece}}}

\affil[2]{\orgname{Archimedes, Athena Research Center, Greece}}

\affil[3]{
%\orgdiv{Dept. of Computer Science and Engineering},
\orgname{University of Ioannina}, \orgaddress{\country{Greece}}}

\abstract{Clustering is a fundamental unsupervised learning task with applications across a wide range of domains. Popular algorithms such as \emph{k}-means are efficient and widely used, but can be sensitive to outliers, ambiguous boundary points, and heterogeneous cluster geometry, which may distort centroid estimates and yield suboptimal partitions. We introduce \textsc{K-Sil}, a silhouette-driven \emph{k}-means variant that, at each iteration, weights points using a centroid-margin proxy for the silhouette score, emphasizing confidently assigned instances while down-weighting borderline or noisy regions. Centroid updates take the form of a softmax-weighted mean, and an adaptive temperature automatically calibrates the sharpness of the weight distribution using a cluster-balanced, macro-averaged, silhouette criterion. Under standard separation conditions, we establish a local convergence result for the induced weighted centroid updates. Experiments on 15 real-world datasets spanning tabular, biomedical, text, and image representations show consistent gains in internal validation metrics and typical improvements in external validation metrics over \emph{k}-means and competitive instance-weighted baselines.}




\maketitle

\section{Introduction}\label{sec:intro}

Clustering, the process of organizing data into meaningful groups, is a cornerstone of
unsupervised learning with broad applications in pattern recognition and data mining~\cite{jain1999data, kaufman2009finding}.
Among clustering algorithms, \emph{k}-means~\cite{macqueen1967} remains widely adopted due to its simplicity, scalability, and strong performance when clusters are well separated~\cite{ikotun2022}. However, the same property that makes \emph{k}-means efficient, updating each centroid as an unweighted mean of its assigned points, can make it sensitive in realistic settings~\cite{bubeck2009}. Under uncertain boundary assignments, noise and outliers, or heterogeneous within-cluster structure, the arithmetic mean can be pulled toward ambiguous regions, propagating early mistakes and yielding suboptimal partitions~\cite{Jain2008DataC5}. This motivates a long line of work on robust and weighted variants of \emph{k}-means, where each sample's influence is adjusted to reduce sensitivity to unreliable observations~\cite{xu2005survey}.

A central challenge in instance-weighted centroid methods is deciding \emph{which} points are reliable and \emph{how strongly} they should influence centroid updates. Many approaches rely on density scores and outlier estimates~\cite{lofkmeans}, or introduce additional modeling assumptions, which may increase complexity or require careful tuning. In contrast, \emph{k}-means iterations already compute a natural geometric signal of assignment confidence: distances to the assigned centroid and to competing centroids. This perspective suggests the following research question:

\medskip
\noindent\textit{How can geometric signals of assignment confidence be converted, at each iteration, into a principled weighting distribution that shapes centroid updates, while preserving the efficient centroid-based structure of $k$-means?}
\medskip

To address this research question we introduce \textsc{K-Sil}, a silhouette-driven, instance-weighted \emph{k}-means variant that converts centroid-distance confidence signals into a weighting mechanism. At each iteration, \textsc{K-Sil} computes for every point a centroid-based approximation of the silhouette score~\cite{wangsilapprox, rousseeuw1987, dudek2020silhouette}. These scores are then mapped through an exponential transformation, producing positive instance weights that define a softmax-weighted centroid update, a convex combination of the points in each cluster. This acts like a within-cluster attention mechanism, giving more influence to confidently assigned points. As centroids move, the weights are recomputed, yielding an iterative reweighting process that increasingly emphasizes well-separated points while reducing the influence of ambiguous boundary points.

A practical consideration is calibrating the sharpness of the weighting.
Overly flat weights recover standard \emph{k}-means, while overly peaked weights can make updates excessively selective. To handle this trade-off without manual tuning, we introduce an adaptive temperature that adjusts the reweighting based on changes in a cluster-balanced (macro-averaged) silhouette score~\cite{pavlopoulos2024} across iterations. When clustering quality improves, the temperature sharpens the weights, as the geometry becomes more reliable and confident points can be emphasized; when it plateaus or falls, it flattens the weights to allow more exploratory updates.

We validate \textsc{K-Sil} with both theory and experiments. Our theoretical analysis studies its local behavior in well-separated regimes and shows local convergence of the induced weighted centroid iterations. Empirically, on a diverse set of 15 real-world datasets, \textsc{K-Sil} consistently improves internal clustering quality (e.g., silhouette, Davies--Bouldin) and typically increases external validation metrics (e.g., NMI, ARI) compared to standard \emph{$k$}-means and other instance-weighted baselines.

\noindent Our code on \textsc{K-Sil} is available at: \url{https://github.com/semoglou/ksil_clustering}.

\section{Related work}\label{sec:related}
\medskip
\textbf{Centroid-based clustering.}

\noindent Clustering algorithms include centroid-based partitioning (e.g., \emph{k}-means)~\cite{macqueen1967}, density-based methods (e.g., DBSCAN)~\cite{ester1996}, hierarchical clustering~\cite{mullner2011modern}, model-based approaches (e.g., GMMs)~\cite{reynolds2009gaussian}, and spectral techniques~\cite{ng2002spectral}. In this study, we focus on improving centroid-based partitioning via instance-level weighting derived from internal confidence signals.
\emph{k}-means alternates between nearest-centroid assignment and updates centroids by unweighted means, minimizing within-cluster variance. While efficient, it is sensitive to initialization, noise, and outliers. Initialization methods, such as \emph{k}-means++~\cite{arthur2007kmeans++} and seeding schemes~\cite{bachem2016afkmc2} improve stability, while variants such as global \emph{k}-means~\cite{likas2003global, vardakas2022global} aim to avoid poor local minima. However, these approaches typically preserve the same mean-based centroid update, so they do not directly control the influence of unreliable points during optimization. 

\medskip
\noindent \textbf{Silhouette and internal validation.}

\noindent Internal validation indices provide feedback on cluster cohesion and separation (e.g., Davies--Bouldin, Calinski--Harabasz)~\cite{arbelaitz2013extensive} without ground-truth cluster labels. Among them, the silhouette coefficient~\cite{rousseeuw1987,dudek2020silhouette} assigns each point a score by contrasting its average intra-cluster dissimilarity with its minimum average inter-cluster dissimilarity. Computing silhouette scores is expensive due to their reliance on pairwise distances, motivating simplified and centroid-based approximations that preserve the same intuition while scaling efficiently~\cite{wangsilapprox}. Beyond evaluation, silhouette has also motivated refinement methods and objectives that explicitly incorporate silhouette-based criteria~\cite{bombina2024,batool2021clustering}. Recent work further shows that the choice of silhouette aggregation, micro (global) vs.\ macro (per-cluster), can substantially affect outcomes, with macro-averaging being particularly suitable when one aims to control cluster-level behavior~\cite{pavlopoulos2024}.
 
\medskip
\noindent \textbf{Weighted \emph{k}-means variants.}

\noindent Weighted variants of \emph{k}-means address the limitations of the unweighted mean update by modulating either feature importance or instance influence. Feature-weighted methods (e.g., WK-Means and EWKM)~\cite{huang2005,jing2007} learn per-feature weights to reduce the effect of irrelevant dimensions, while related schemes such as AWA~\cite{chan2004optimization} use variance-driven weighting to improve robustness under noisy features. Recent feature-weighted variants also use silhouette to update dimension weights during \emph{k}-means iterations (e.g., WKBSC)~\cite{lai2024}.
More closely related to our setting are instance-weighted approaches, which control how strongly individual points influence centroid updates. A prominent line of work derives weights from outlier analysis estimated via local neighborhood structure, as in LOF-based and iterative iLOF-based \emph{k}-means~\cite{lofkmeans, ilof}. 
Similarly, OWKMeans introduces an object-weighted \emph{k}-means objective, updating per-point weights and penalties from distance and silhouette-width signals, with higher penalties for poorly separated or overlapping points~\cite{owkmeans}. \textsc{K-Sil} follows this line by turning silhouette-based confidence into within-cluster weighting distributions that are adaptively sharpened over iterations and directly shape the centroid updates.

\section{Methodology}\label{sec:method}
\subsection{Data and notation}\label{subsec:notation}
Let $\Xp = \left\{x_1, x_2, \dots,x_n \right\} \subset \mathbb{R}^d$ be fixed data points. A clustering into $k \in \mathbb{N} \ (k>1)$ distinct clusters $\left\{C_1, C_2, \dots, C_k \right\}$ is described by (\textit{i}) the corresponding centroids $\mu$:
\begin{equation}\label{eq:centers}
\mu = (\mu_1, \ \mu_2, ..., \ \mu_k) \in (\mathbb{R}^d)^k \cong \mathbb{R}^{kd},
\end{equation}
and (\textit{ii}) the cluster labels (indices) $c(i)$ for each data point $x_i \in \Xp$: 
\begin{equation}\label{eq:labels}
c:\left\{1, \dots, n \right\} \rightarrow \left\{1, \dots,k \right\},  \text{ } c(i) \text{: the cluster index for point } x_i\in \Xp.
\end{equation}
Given centers $\mu$, we use the nearest–centroid assignment: 
\begin{equation}\label{eq:assignment}
c(\mu)(i) \in \arg \min_{1 \le j \le k} \|x_i - \mu_j \| \ \forall x_i \in \Xp 
\end{equation}
with a fixed and deterministic tie-breaking rule (e.g. smallest index).
We write the index set of cluster $j$ as $C_j(\mu) = \left\{i: c(\mu)(i)=j \right\}$, and its size as $n_j(\mu) = | C_j(\mu)|. $

\subsection{Centroid initialization}\label{subsec:initialization}
Given a prescribed number of clusters $k$, \textsc{K-Sil} starts from an initial set of centroids $\mu_0 = (\mu_{0,1}, \dots, \mu_{0,k}) \in (\mathbb{R}^d)^k$. In practice, $\mu_0$ can be obtained either by selecting $k$ data points at random from $\Xp$ or by using a more structured seeding strategy such as $k$-means++ initialization. Once $\mu_0$ is chosen, the initial labels are assigned by the nearest–centroid rule in Eq.~\ref{eq:assignment}. 
The subsequent sections (\S\ref{subsec:sil}--\S\ref{subsec:conv}) describe how, given current centroids and cluster labels, \textsc{K-Sil} computes silhouette scores, instance weights, and updates centroids.

\subsection{Centroid-margin silhouette proxy}\label{subsec:sil}
At each iteration of \textsc{K-Sil}, given the current centroids and labels, we first compute silhouette-based scores for all points. These
scores are then used (\textit{i}) to define silhouette-based instance weights (\S\ref{subsec:weights_update}), and (\textit{ii}) to provide a global measure of clustering quality that controls how strongly \textsc{K-Sil} emphasizes high-confidence points in subsequent updates (\S\ref{subsec:temperature}). 

Classical silhouette scores are defined using average intra- and inter-cluster distances between points. For efficiency, \textsc{K-Sil} uses a centroid-based proxy: each point’s silhouette is computed solely from its distance to cluster centroids. Given centers $\mu = (\mu_1, \mu_2 \dots, \mu_k)$ and the associated labels $c(\mu)$, we define for each point $x_i \in \Xp$ its intra-cluster distance proxy (own centroid distance) $a_i = a_i(\mu)$ and its inter-cluster distance proxy (nearest other centroid distance) $b_i = b_i(\mu)$:\footnote{In the standard silhouette definition, $a_i$ is the average distance from $x_i$ to all other points in its assigned cluster, and $b_i$ is the minimum (over clusters $j\neq c(i)$) of the average distance from $x_i$ to points in cluster $j$.}  
\begin{equation}\label{eq:intra_inter}
a_i(\mu) = \|x_i - \mu_{c(\mu)(i)} \|, \quad
b_i(\mu) = \min_{j \neq c(\mu)(i)} \|x_i - \mu_j\|.
\end{equation}
The per-point silhouette proxy is then: 
\begin{equation}\label{eq:silhouette_score}
    s_i(\mu) = \frac{b_i(\mu) - a_i(\mu)}{\max{\{a_i(\mu), \ b_i(\mu)\}}}.
\end{equation}
In our centroid-distance proxy, $s_i(\mu)$ is best understood as a normalized centroid--margin: it compares the distance from $x_i$ to its assigned centroid with the distance to the nearest other centroid. Values $\approx 1$ indicate points that lie well inside their assigned cluster, whereas values near $0$ correspond to points close to a centroid decision boundary. Moreover \emph{the proxy scores are constrained to be nonnegative}. By the assignment rule in Eq.~\ref{eq:assignment},
$
a_i(\mu) = \|x_i - \mu_{c(\mu)(i)}\|
\le \|x_i - \mu_j\| \quad \forall j,
$
and in particular $a_i(\mu) \le b_i(\mu)$. We therefore have $\max\{a_i(\mu), b_i(\mu)\} = b_i(\mu)$ and thus 
$
s_i(\mu) = \frac{b_i(\mu) - a_i(\mu)}{b_i(\mu)} \in [0,1].
$
If $a_i(\mu) > b_i(\mu)$ were to occur, then $c(\mu)(i)$ could not be a minimizer in Eq.~\ref{eq:assignment}, which gives a contradiction. Additionally, we aggregate the pointwise silhouette scores into a macro-averaged, cluster-level score by averaging the mean silhouette of each cluster ($n_j\!\ge\!1$; see empty-cluster re-initialization strategy in Appendix~\ref{app:empty}):
\begin{equation}\label{eq:macro_silhouette}
S(\mu) = \frac{1}{k}\sum_{j=1}^{k}\frac{1}{n_j(\mu)}\sum_{c(\mu)(i)=j}s_i(\mu) \in [0,1].
\end{equation}
This macro-averaging is particularly useful in the presence of unbalanced cluster sizes. Unlike the global mean silhouette score $\frac{1}{n}\sum_{x_i}s_i$ (the standard micro-averaged score used by \texttt{sklearn}), which is dominated by large clusters, $S(\mu)$ gives each cluster equal weight in the overall score. As a result, it is sensitive to poorly separated or low-quality clusters even when they contain relatively few points, and thus provides a more balanced measure of clustering quality for guiding the adaptive steps of \textsc{K-Sil} (\S\ref{subsec:temperature}).

\subsection{Instance weights and centroid update}\label{subsec:weights_update}
Given centers $\mu$ and a temperature parameter $\tau > 0$ (\S\ref{subsec:temperature}), \textsc{K-Sil} uses the silhouette scores $\{s_i(\mu)\}$ to define instance weights $w_i$ and update the cluster centroids as weighted means. Each instance $x_i \in \Xp$ is assigned a silhouette-based weight:
\begin{equation}\label{eq:weights}
w_i(\mu, \tau) = \exp\left\{ \tau s_i(\mu) \right\} >0.
\end{equation}
Points with higher silhouette $s_i(\mu)$ (i.e.\ better cluster fit) thus receive exponentially higher weights, while ambiguous or poorly fitted points are down-weighted. Given centroids $\mu = (\mu_j)_{j=1}^k$ (Eq.~\ref{eq:centers}), the (updated) weighted centroid of cluster $j$, $\mu_j$, is: 
\begin{equation}\label{eq:update}
H_j(\mu, \tau) = \frac{\sum_{i \in C_j(\mu)}w_i(\mu, \tau)x_i}{\sum_{i \in C_j(\mu)}w_i(\mu, \tau)} =  \mathrm{softmax}\bigg(\tau \bigg[s_i(\mu)\bigg]_{i\in C_j(\mu)}\bigg)^{\!\top} \bigg[x_i\bigg]_{i\in C_j(\mu)},
\end{equation}
where the softmax is taken over the index set $C_j(\mu)$ and
$[x_i]_{i \in C_j(\mu)}$ denotes the stacked vector of points in cluster $j$. Thus, $H_j(\mu,\tau)$ is a convex combination of the points in $C_j(\mu)$, with weights proportional to $\exp\{\tau s_i(\mu)\}$ and normalized to one via softmax.

Geometrically, the silhouette scores $s_i(\mu)$ quantify how well each point fits its assigned cluster, and the softmax transformation turns these scores into a probability distribution within each cluster (over $C_j(\mu)$). 
This is analogous to within-cluster attention: points with high silhouette receive most of the probability mass and therefore ``pull'' the centroid $H_j(\mu,\tau)$ toward well-supported regions of the cluster, while low-silhouette boundary or misclustered points have little influence. This makes the centroid update more robust than an unweighted mean and allows \textsc{K-Sil} to focus its updates on high-confidence points.
We collect the cluster-wise centroid updates into:
\begin{equation}\label{eq:mu+1}
H(\mu, \tau) = \big(H_1(\mu, \tau), \ H_2(\mu, \tau), \dots, \ H_k(\mu, \tau)\big) \in \mathbb{R}^{kd},
\end{equation}
which defines the centroid update step of one \textsc{K-Sil} iteration for a temperature $\tau$.

\subsection{Temperature parameter}\label{subsec:temperature}
\begin{figure}[htbp]
    \vspace{-4mm}
    \centering
    \includegraphics[width=\textwidth, trim=0 0 0 7mm, clip]{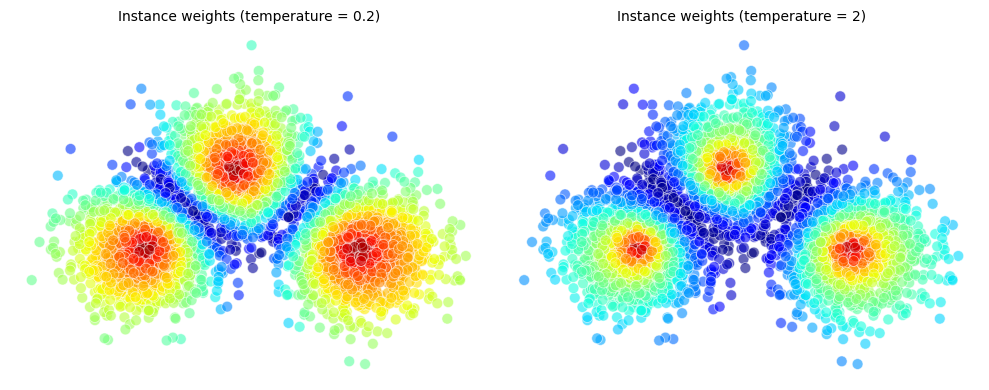}
    \caption{Effect of temperature $\tau$ on weights (Eq.~\ref{eq:weights}), on synthetic data ($k=3$). \textbf{Left}: $\tau = 0.2$ (flatter weighting). \textbf{Right}: $\tau = 2$ (more peaked weighting) ($\uparrow$ \textcolor{red}{red}, $\downarrow$ \textcolor{blue}{blue}).}
    \label{fig:tempdiffs}
    \vspace{-6mm}
\end{figure}

The temperature $\tau > 0$ controls the sharpness of the silhouette-based weights in Eq.~\ref{eq:weights}. For small $\tau$, we have
$\exp\{\tau s_i(\mu)\} \approx 1$ for all $i$, so the weights
$w_i(\mu,\tau)$ are nearly uniform and $H_j(\mu,\tau)$ is close to the standard (unweighted) cluster mean. As the temperature $\tau$ increases, the softmax in Eq.~\ref{eq:update} becomes more peaked: points with larger silhouette $s_i(\mu)$ (well inside the cluster) receive disproportionately large weights, while low-silhouette points contribute very little. In the limit of large $\tau$, the centroid update is dominated by the highest-silhouette points in each cluster. Thus, $\tau$ acts as a ``focus'' parameter that interpolates between uniform averaging and an update driven almost exclusively by the most confidently assigned points (Fig.~\ref{fig:tempdiffs}).

\noindent In \textsc{K-Sil}, the temperature is initialized at a neutral value, $\tau_0 = 1 \text{ or }2$ (Appendix~\ref{app:inittau} describes an optional rule selecting $\tau_0\in\{1,2\}$ from the initial $s_i(\mu_0)$ distribution), and then updated adaptively based on the evolution of the macro silhouette $S(\mu)$ (Eq.~\ref{eq:macro_silhouette}) across iterations. Let $S_t = S(\mu_t)$ denote the macro silhouette at iteration $t$.\\ For $t \ge 1$ we first compute a normalized score drift by comparing $S_t$ and $S_{t-1}$: 
\begin{equation}\label{eq:r}
\tilde r(S_t,S_{t-1}) = \frac{S_t - S_{t-1}}{1-S_{t-1} + \varepsilon} , \quad \varepsilon>0 \text{ small for stability}.
\end{equation}
Since $S_t \in [0,1]$ (\S\ref{subsec:sil}), the denominator $1 - S_{t-1}$ approximately measures the remaining ``headroom'' to the best possible value $(1)$. Thus, $\tilde r(S_t,S_{t-1})(<1)$ encodes the signed change in $S_t$ relative to the largest improvement still possible from $S_{t-1}$: values $\tilde r(S_t,S_{t-1}) \approx 1$ correspond to improvements that nearly exhaust this headroom, while small $|\tilde r(S_t,S_{t-1})|$ indicate only minor changes. To prevent overly large temperature reactions when $S_t$ drops, we use
$R(S_t,S_{t-1})=\max\{-1,\tilde r(S_t,S_{t-1})\}\in[-1,1]$. 
Next, we fix a temperature lower bound $\tau_{\min} >0$ and define an upper bound $\tau_{\max}(t)$ per iteration using the current cluster sizes. Let $n_j(\mu_t) = |C_j(\mu_t)|$ be the size of cluster $j$ at iteration $t$ (assuming $n_j(\mu_t)\ge 3$ $\forall j,t$, see Appendix~\ref{app:empty}), and set $m_{\max}(\mu_t) = \max_{1\le j \le k}n_j(\mu_t)$, the size of the largest cluster at iteration $t$.
Within a cluster, weight ratios scale exponentially: $w_i/w_h=\exp\{\tau(s_i-s_h)\}$ for $x_i, x_h$ in the same cluster.
To keep this contrast from becoming excessively large as cluster size grows, we cap $\tau$ using a worst-case reference configuration in a cluster of size $m$, where one point has $s=1$ and the remaining $m-1$ points have $s=0$.
In this configuration the cluster's mean silhouette is $1/m$, so the best point exceeds the mean by $1-1/m$.
We choose $\tau_{\max}$ so that the corresponding weight spread relative to this mean is at most $m^2$ (a quadratic cap that prevents the exponentiated weights from becoming excessively large as $m$ grows):
$\exp\Big\{\tau\Big(1-\frac{1}{m}\Big)\Big\}\le m^2$,
which yields $\tau\le \frac{2\log m}{(m-1)/m}$.
We therefore set $\tau_{\max}(t)$ using $m=m_{\max}(\mu_t)$, i.e., the bound is calibrated by the largest cluster size, and define:
\begin{equation}\label{eq:tau_max_new}
\tau_{\max}(t)
=\frac{2\,m_{\max}(\mu_t)\,\log m_{\max}(\mu_t)}{m_{\max}(\mu_t)-1}.
\end{equation}
This cap is a stabilization safeguard on the temperature dynamics, and it scales only logarithmically with $m_{\max}(\mu_t)$.
The raw temperature update is a multiplicative, exponentiated-gradient
step: 
\begin{equation}\label{eq:tauraw}
\tau_{t+1}^{\mathrm{raw}} = \tau_t \exp\left\{\eta\, R(S_t, S_{t-1}) \right\},
\end{equation}
where $\eta$ is a learning rate parameter (default: 0.2; Appendix Fig.~\ref{fig:ablationlr} shows that, under the temperature bounds, performance is largely insensitive to $\eta$ over $\{0.1,0.2,0.3\}$, making further tuning typically unnecessary).
Thus $ \tau_t e^{-\eta} \le \tau_{t+1}^{\mathrm{raw}} \le \tau_t e^{\eta}$ and 
the new temperature $\tau_{t+1}$ is obtained by clipping $\tau_{t+1}^{\mathrm{raw}}$,
so that $\tau_{\min} \le \tau_{t+1} \le \tau_{\max}(t) \ \forall t$.
In regimes where the cluster sizes stabilize, $\tau_{\max}(t)$ also stabilizes, and $\tau$ remains in a fixed interval $[\tau_{\min}, \tau_{\max}^\star]$. 
This adaptive scheme has an intuitive interpretation: if $S_t$ increases, the algorithm gradually raises $\tau$, sharpening the weights and letting the centroids focus more on high-quality points. If $S_t$ decreases, the algorithm lowers $\tau$, flattening the weights and allowing for broader, more exploratory centroid updates. We summarize this update as an abstract map:
\begin{equation}\label{eq:Psi_map}
    \Psi_\eta(\tau_t, S_t, S_{t-1}) = \tau_{t+1} = \mathrm{clip}_{[\tau_{\min},\,\tau_{\max}(t)]}
       \bigl[\tau_t \exp\{\eta\, R(S_t, S_{t-1})\}\bigr].
\end{equation}

\subsection{Convergence and stopping criterion}\label{subsec:conv}
\textsc{K-Sil} alternates between (\textit{i}) computing centroid-proxy silhouettes $s_i(\mu)$ and the macro averaged silhouette score $S(\mu)$ (\S\ref{subsec:sil}, Eqs.~\ref{eq:silhouette_score}--\ref{eq:macro_silhouette}), (\textit{ii}) updating the temperature $\tau$ using the adaptive rule $\Psi_\eta$ (\S\ref{subsec:temperature}, Eq.~\ref{eq:r}--\ref{eq:Psi_map}), (\textit{iii}) updating centroids using the weighted map $H$ (\S\ref{subsec:weights_update}, Eqs.~\ref{eq:update}--\ref{eq:mu+1}), and (\textit{iv}) assigning labels with the nearest-centroid $c$ (\S\ref{subsec:notation}, Eq.~\ref{eq:assignment}) with a deterministic tie-breaking; if any cluster becomes empty, its centroid is re-initialized using the safeguard in Appendix~\ref{app:empty}. These updates (outlined in Alg.~\ref{alg:ksil}) are repeated until either a maximum number of iterations $T_{\max}$ is reached (default $T_{\max}=100$) or when the centroids stabilize. Specifically, letting $\mu_{t,j}$ be a centroid $j$ at iteration $t$, we compute the average centroid movement:
\begin{equation}\label{eq:avgmov}
\Delta_t = \frac{1}{k}\sum_{j=1}^{k}\|\mu_{t+1,j} - \mu_{t,j}\|,
\end{equation}
and declare convergence when $\Delta_t < \mathrm{tol}$ for a tolerance level $\mathrm{tol}>0$ ($10^{-4}$ in default). \footnote{In all experiments reported (\S\ref{subsec:eval}), \textsc{K-Sil} satisfied the centroid-stabilization criterion $\Delta_t < \mathrm{tol}$ before reaching the iteration cap $T_{\max}$ (Fig.~\ref{fig:convergence}; Appendix~\ref{app:empirical} Figs.~\ref{fig:convergence_app_a}--\ref{fig:convergence_app_d}). Thus, $T_{\max}$ serves only as a safeguard to guarantee termination in pathological cases.}

\vspace{-5mm}
\begin{algorithm}[H]
\caption{\textsc{K-Sil} clustering}
\label{alg:ksil}
\begin{algorithmic}[]
\Require Dataset $\Xp = \{x_i\}_{i=1}^n \subset \mathbb{R}^d$, number of clusters $k>1$, learning rate $\eta > 0$ (default: 0.2), maximum number of iterations $T_{\max}$ (default: $100$), centroid movement tolerance (convergence threshold) $\mathrm{tol}$ (default: $10^{-4}$)
\Ensure Cluster centroids $\mu$ and labels $c$

\State \textbf{Initialization} (\S\ref{subsec:initialization})
\State Choose initial centroids $\mu_0 = (\mu_{0,1}, \dots, \mu_{0,k})$ (e.g.\ random points in $\Xp$)
\State Assign initial labels $c_0 \gets c(\mu_0)$ using nearest--centroid rule (Eq.~\ref{eq:assignment})
\State Set initial temperature $\tau_0 \gets 1.0$ (Appendix~\ref{app:inittau}) and $S_{-1} \gets$ \texttt{None}

\For{$t = 0, 1,\dots,T_{\max}-1$}
    \State \textbf{Silhouette computation} (\S\ref{subsec:sil})
    \State Given $(\mu_t, c_t)$, compute $s_i(\mu_t)$ and $S_t = S(\mu_t)$
           (Eqs.~\ref{eq:intra_inter}--\ref{eq:macro_silhouette})

    \State \textbf{if} $S_{t-1} =$ \texttt{None} \textbf{then} set $S_{t-1} \gets S_t$ so the first update uses $S_t = S_{t-1}$

    \State \textbf{Temperature update} (\S\ref{subsec:temperature})
    \State Update temperature $\tau_{t+1} \leftarrow \Psi_\eta(\tau_t, S_t, S_{t-1})$ (Eq.~\ref{eq:Psi_map})

    \State \textbf{Weight computation and centroid update} (\S\ref{subsec:weights_update})
    \State Compute weights $w_i(\mu_t,\tau_{t+1}) \leftarrow \exp\{\tau_{t+1}\, s_i(\mu_t)\}$ for all $i$
           (Eq.~\ref{eq:weights})

    \State Update centroids by the weighted centroid map
           $\mu_{t+1} \leftarrow H(\mu_t,\tau_{t+1})$ (Eqs.~\ref{eq:update}--\ref{eq:mu+1})
    \State Reassign labels $c_{t+1} \gets c(\mu_{t+1})$ using nearest--centroid
           assignment (Eq.~\ref{eq:assignment})
    \State \textbf{if} any cluster is empty \textbf{then} re-initialize its centroid (as in Appendix~\ref{app:empty})

    \State \textbf{Convergence check} (\S\ref{subsec:conv})
    \State Compute the average centroid movement
           $\Delta_t \leftarrow \frac{1}{k}\sum_{j=1}^k \|\mu_{t+1,j} - \mu_{t,j}\|$ (Eq.~\ref{eq:avgmov})

    \State \textbf{if} $\Delta_t< \mathrm{tol}$ \textbf{then} \textbf{return} centroids $\mu_{t+1}$ and labels $c_{t+1}$

    \State Set $S_{t-1} \gets S_t$ \text{ for the next iteration}
\EndFor

\State \textbf{return} final centroids $\mu_{T_{\max}}$ and labels $c_{T_{\max}}$
\end{algorithmic}
\end{algorithm}

\section{Theoretical Analysis}\label{sec:theory}
\subsection{Local convergence}\label{subsec:localconv}
We study the local behavior of \textsc{K-Sil} around a well-separated configuration.
Recall that one iteration updates $\tau_{t+1}=\Psi_\eta(\tau_t,S_t,S_{t-1}),\
\mu_{t+1}=H(\mu_t,\tau_{t+1}),\
S_t=S(\mu_t),$ (\S\ref{subsec:conv}, Alg.~\ref{alg:ksil}) with labels given by nearest-centroid assignment $c(\mu_t)$ (Eq.~\ref{eq:assignment}). We list Assumptions A.1, A.2 and Lemmas L.1–L.4 used in our analysis, along with the local convergence theorem. Full proofs and detailed derivations are provided in Appendix~\ref{app:theory}.

\medskip
\noindent\textbf{Norm.}
For $\mu\in\mathbb{R}^{kd}$ (Eq.~\ref{eq:centers}) we use the block-max norm
$\|\mu\|_{\infty,2}=\max_{1\le j\le k}\|\mu_j\|_2.$

\medskip
\noindent Fix centers $\mu_\star\in\mathbb{R}^{kd}$ and let $c_\star=c(\mu_\star)$ denote the induced nearest--centroid labels.
Throughout the local analysis we assume the induced clustering has no empty clusters  and we study iterates initialized in a neighborhood of $\mu_\star$.

\medskip
\noindent\textbf{A.1 (Geometric separation at $\mathbf{\mu_\star}$).}
There exists $r>0$ such that for each cluster $j$ and all $i$ with $c_\star(i)=j$,
$\|x_i-\mu_{\star,j}\|\le r,
\ \text{and}\
\|\mu_{\star,j}-\mu_{\star,\ell}\|\ge 5r\quad\forall j\neq \ell.$ Thus, each cluster lies inside a ball of radius $r$ around its centroid, and distinct centroids are at least $5r$ apart. This is an idealized but standard local margin assumption used to simplify the analysis by ruling out nearly overlapping clusters near $\mu_\star$.

\medskip
\noindent\textbf{A.2 (Local bounded-temperature).}
We analyze the dynamics in a neighborhood where:
(i) $\tau$ remains bounded: $\tau_t\in[\tau_{\min},\bar\tau]$ for all $t$, for some $\bar\tau>0$;
(ii) the drift lower clipping is inactive along iterates, i.e. $\tilde r(S_t,S_{t-1})>-1$ so that
$R(S_t,S_{t-1})=\tilde r(S_t,S_{t-1})$;
(iii) the ``headroom'' denominator is bounded below:
there exists $\alpha>0$ such that $1-S_t+\epsilon\ge \alpha$ for all $t$.
((i) is ensured in practice by Eq.~\ref{eq:Psi_map}; (ii)--(iii) are mild local regularity conditions ruling out
saturation of the drift normalization.)

\medskip
\noindent\textbf{L.1 (Stability near $\mathbf{\mu_\star}$).}
Let $U_\mu:=\{\mu\in\mathbb{R}^{kd}:\ \|\mu-\mu_\star\|_{\infty,2}\le r\}$  be a neighborhood of $\mu_\star$.
Under A.1, for every $\mu\in U_\mu \Rightarrow c(\mu)=c_\star$.
Moreover, $H(\mu,\tau)\in U_\mu \ \forall \tau \ge 0$.

\medskip
\noindent\textbf{L.2 (Silhouette Lipschitz bound).}
For $\mu,\mu'\in U_\mu$, each $s_i(\mu)$ and $S(\mu)$ satisfy $|s_i(\mu)-s_i(\mu')|\le L_s\|\mu-\mu'\|_{\infty,2}$, $|S(\mu)-S(\mu')|\le L_s\|\mu-\mu'\|_{\infty,2}$ with $L_s=\frac{5}{9r}$.

\medskip
\noindent\textbf{L.3 (Softmax Lipschitz).}
$\forall u,u'\in\mathbb{R}^m$, $\|\mathrm{softmax}(u)-\mathrm{softmax}(u')\|_1 \le \|u-u'\|_\infty$.

\medskip
\noindent\textbf{L.4 (Lipschitz bounds for the centroid map $H$).}
Let $\mu,\mu'\in U_\mu$, $\tau,\tau'\ge 0$ then
$
\|H(\mu,\tau)-H(\mu',\tau)\|_{\infty,2}\le \frac{5\tau}{9}\,\|\mu-\mu'\|_{\infty,2},
\ \text{ and } \
\|H(\mu,\tau)-H(\mu,\tau')\|_{\infty,2}\le r\,|\tau-\tau'|.
$

\medskip
\noindent\textbf{Theorem 1 (Local convergence under bounded temperature).}\label{thm:localconvergence}
Assume A.1, A.2 and let $\mu_0\in U_\mu$.
If the temperature bound $\bar\tau$ satisfies $\bar q =\frac{5\bar\tau}{9}\Big(1+\frac{e^\eta-1}{\alpha}\Big)\;<\;1,$
then the \textsc{K-Sil} iterates satisfy:
\textbf{(i)} (local invariance) $\mu_t\in U_\mu$ and $c(\mu_t)=c_\star$ for all $t$;
\textbf{(ii)} (linear decay of step sizes) with $d_t:=\|\mu_t-\mu_{t-1}\|_{\infty,2}$,
$d_{t+1}\le \bar q\, d_t\quad \text{for all }t\ge 1;$
\textbf{(iii)} (convergence) $\mu_t\to\mu_\infty\in U_\mu$, $S_t\to S_\infty:=S(\mu_\infty)$, and $\tau_t\to\tau_\infty\in[\tau_{\min},\bar\tau]$;

\medskip
\noindent
Under Theorem 1, labels remain fixed ($c_t=c_\star$), so $s_i(\mu_t)\to s_i(\mu_\infty)$ for each $i$.
Since $\mu_t\to\mu_\infty$ and $\tau_t\to\tau_\infty$, we have $w_i{(\mu_t, \tau_{t+1})}\to w_i{(\mu_\infty, \tau_\infty)}$.
In particular, the limiting centers satisfy
$
\mu_{\infty,j}
=\frac{\sum_{i:c_\star(i)=j} w_i{(\mu_\infty, \tau_\infty)} x_i}{\sum_{i:c_\star(i)=j} w_i{(\mu_\infty, \tau_\infty)}},
$
so $\mu_\infty$ is a fixed point of a weighted $k$-means update with weights induced by the limiting silhouette-confidence profile.

\subsection{Computational complexity}\label{subsec:complexity}
Let $n$ be the number of points, $d$ the dimension, and $k$ the number of clusters.
The dominant cost is computing distances from each point to the $k$ centroids. For each $x_i$, we need (\textit{i}) its assigned centroid distance $\|x_i-\mu_{c(i)}\|$ and
(\textit{ii}) the nearest other centroid distance $\min_{j\neq c(i)}\|x_i-\mu_j\|$, which can be obtained by tracking the smallest and second-smallest centroid distances in one pass.
This requires $O(kd)$ work per point, hence $O(nkd)$ per iteration.
Given these distances, computing silhouettes $s_i$ and the macro score $S(\mu)$ is $O(n)$, the temperature update is $O(k)$ (via $m_{\max}$), and computing weights plus the weighted centroid update
is $O(nd)$ using per-cluster accumulators (one weighted sum and one weight sum per cluster). The label reassignment step is already available from the same distance pass (or costs another $O(nkd)$
if recomputed from scratch). Overall, the per-iteration time is
$O(nkd) \ \text{(dominant term)},$ matching the standard $k$-means iteration order, with only constant-factor overhead.
For $T_{\max}$ iterations, the total runtime is $O(T_{\max} nkd)$ (plus initialization, e.g.\ random seeding).

\section{Empirical Validation}\label{sec:empirical}

\subsection{Datasets}\label{subsec:datasets}
\begin{figure}[h]
    \centering
    \includegraphics[width=\textwidth, trim=10mm 12mm 0 12mm, clip]{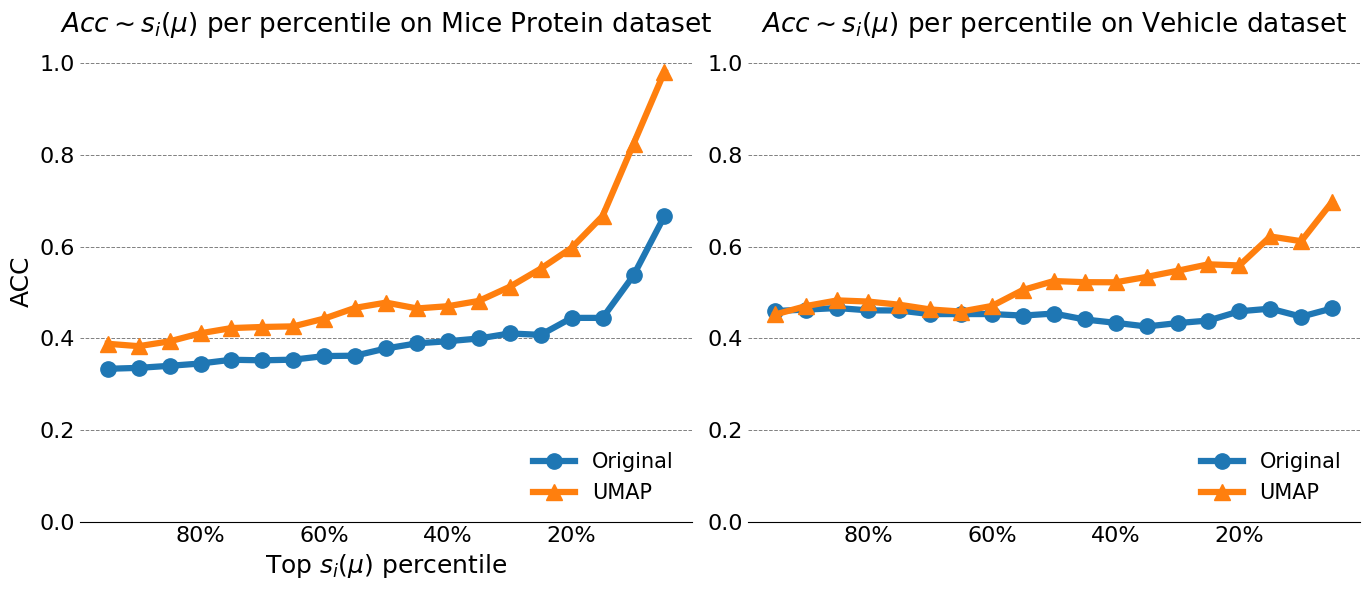}
    \caption{Hungarian-matched accuracy across \emph{top} $s_i(\mu)$ percentiles (from a reference $k$-means partition), comparing the \textcolor{MidnightBlue}{original feature space} and the \textcolor{orange}{UMAP representation} on \textsc{Mip} (left) and \textsc{Vcl} (right).}
    \label{fig:acc_vs_sil_umap}
      \vspace{-1.5em}
\end{figure}

Our empirical study consists of 15 real-world datasets from biomedical and health data, structured tabular and sensor benchmarks, natural-language corpora for topic/intent classification, and an image benchmark. All datasets are available via OpenML, UCI, and Hugging Face Datasets. We summarize each dataset by its sample size $n$, number of classes $k$, and post-preprocessing dimensionality $d$.
\textbf{Leukemia} (\textsc{Lkm}): cancer gene-expression profiles; $n{=}72$, $k{=}2$, $d{=}7129$,
\textbf{Vehicle} (\textsc{Vcl}): vehicle silhouette features; $n{=}846$, $k{=}4$, $d{=}15$,
\textbf{Ionosphere} (\textsc{Inr}): radar returns; $n{=}351$, $k{=}2$, $d{=}34$,
\textbf{Mice Protein} (\textsc{Mip}): protein expression measurements; $n{=}1{,}080$, $k{=}8$, $d{=}15$,
\textbf{Diabetes} (\textsc{Dbt}): patient clinical measurements; $n{=}768$, $k{=}2$, $d{=}8$,
\textbf{Wine} (\textsc{Wne}): wine chemistry measurements; $n{=}178$, $k{=}3$, $d{=}13$,
\textbf{Breast Cancer} (\textsc{BrC}): tumor diagnostic measurements; $n{=}569$, $k{=}2$, $d{=}30$,
\textbf{SMS Spam} (\textsc{Sms}): short text messages; $n{=}5{,}574$, $k{=}2$, $d{=}50$,
\textbf{MINDS-14} (\textsc{Mds}): spoken user requests; $n{=}567$, $k{=}14$, $d{=}50$,
\textbf{HTRU2} (\textsc{Htr}): pulsar candidate features; $n{=}17{,}898$, $k{=}2$, $d{=}8$,
\textbf{BBC News} (\textsc{Bbc}): news articles by topic; $n{=}2{,}225$, $k{=}5$, $d{=}50$,
\textbf{R8} (\textsc{Re8}): Reuters newswire topics; $n{=}4{,}941$, $k{=}8$, $d{=}50$,
\textbf{Banking77} (\textsc{B77}): banking user queries; $n{=}13{,}083$, $k{=}77$, $d{=}50$,
\textbf{CLINC150} (\textsc{Clc}): assistant user utterances; $n{=}23{,}800$, $k{=}151$, $d{=}50$,
and \textbf{STL10} (\textsc{Stl}): natural images; $n{=}13{,}000$, $k{=}10$, $d{=}100$.
Preprocessing follows the data modality. For tabular datasets we apply mean imputation when needed, followed by z-score standardization; Leukemia is treated analogously but remains high-dimensional after scaling. For the text datasets (\textsc{Sms}, \textsc{Mds}, \textsc{Bbc}, \textsc{Re8}, \textsc{B77}, \textsc{Clc}), each document is embedded with MiniLM (sentence-level Transformer encoder), yielding representations that capture semantic similarity; we then project to $d{=}50$ with PCA and apply $\ell_2$ normalization. For \textsc{Stl}, we extract CLIP (ViT-B/32) image embeddings and apply PCA to $d{=}100$ followed by $\ell_2$ normalization (Appendix~\ref{app:empirical}, Table~\ref{tab:sil_exact_vs_approx}, reports per-dataset correlation/MAE between exact \texttt{sklearn} silhouette scores and our centroid-margin proxy \S\ref{subsec:sil}, along with timing speedups). For \textsc{Mip} and \textsc{Vcl} we use a UMAP representation with $d{=}10$. On these two datasets, the original feature space shows weaker alignment between silhouette-based internal separation and label-consistent external structure, which can make silhouette-based confidence weighting less informative. In the UMAP space, $s_i(\mu)$ aligns more clearly with external agreement; this is reflected in a clearer relationship between $s_i(\mu)$ and Hungarian-matched clustering accuracy when we compare the original space to UMAP embedding across $s_i(\mu)$ percentiles in Fig.~\ref{fig:acc_vs_sil_umap}.

\subsection{Evaluation}\label{subsec:eval}

\noindent\textbf{Empirical convergence.}

\noindent To complement our local convergence analysis (\S\ref{subsec:localconv}), we empirically examine the iterative behavior of \textsc{K-Sil}. For each dataset in \S\ref{subsec:datasets}, we applied \textsc{K-Sil} from a random initialization with a maximum number of iterations $T_{\max}=100$, and at each iteration $t$, we  record two diagnostics: the average centroid movement $\Delta_t$ (Eq.~\ref{eq:avgmov}) and the Pearson correlation $\rho_t$ between the per-instance weight vectors (Eq.~\ref{eq:weights}), computed at consecutive iterates $w(\mu_{t})$ and $w(\mu_{t+1})$:
$$\Delta_t = \frac{1}{k}\sum_{j=1}^{k}\|\mu_{t+1,j} - \mu_{t,j}\|, \quad
\rho_t \;=\; \mathrm{corr}\!\left(w(\mu_{t+1}),\, w(\mu_{t})\right).$$
Across datasets, $\Delta_t$ decreases rapidly with $t$, while $\rho_t$ quickly rises toward $1$, indicating that the confidence weights stabilize after only a few iterations and subsequent updates mainly refine the centroids under near-fixed weights. This behavior is consistent with our theory: once assignments and weights evolve smoothly across iterations, the update behaves like a locally contractive weighted k-means step, yielding fast local convergence to a fixed point. Fig.~\ref{fig:convergence} and Figs.~\ref{fig:convergence_app_a}--\ref{fig:convergence_app_d} in Appendix~\ref{app:empirical} visualize these trends (see also Appendix Fig.~\ref{fig:sse_sil} for additional iteration diagnostics on representative datasets, including SSE and silhouette score over \textsc{K-Sil} iterations).

\begin{figure}[H]
    \centering

    \begin{subfigure}[t]{\linewidth}
        \centering
        \includegraphics[width=0.75\linewidth]{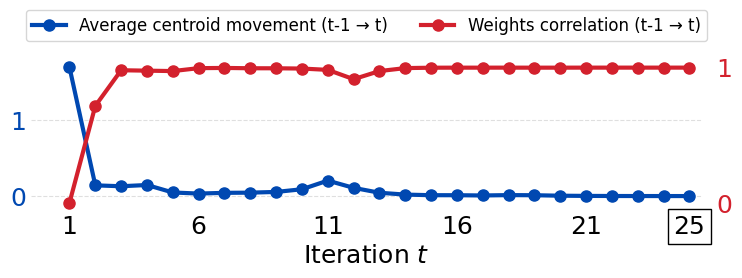}
        \caption{Mice Protein (\textsc{Mip})}
    \end{subfigure}\vspace{0.1em}

    \begin{subfigure}[t]{\linewidth}
        \centering
        \includegraphics[width=0.75\linewidth,trim=0 0 0 15mm,clip]{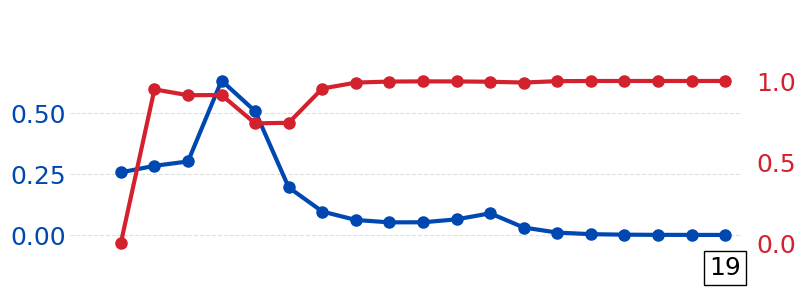}
        \caption{Vehicle (\textsc{Vcl})}
    \end{subfigure}\vspace{0.1em}

    \begin{subfigure}[t]{\linewidth}
        \centering
        \includegraphics[width=0.75\linewidth,trim=0 0 0 15mm,clip]{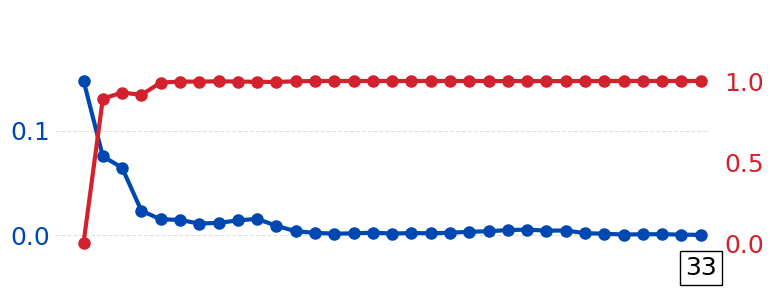}
        \caption{STL10 (\textsc{Stl})}
    \end{subfigure}

    \caption{\textcolor{blue}{Average centroid movement} and \textcolor{red}{Pearson correlation between weights} across \textsc{K-Sil} iterations $t$ (\S\ref{subsec:eval}) for \textsc{Mip}, \textsc{Vcl} and \textsc{Stl}. The final iteration (convergence) is marked by $\square$. The remaining datasets (\S\ref{subsec:datasets}) are shown in Appendix~\ref{app:empirical} (Figs.~\ref{fig:convergence_app_a}--\ref{fig:convergence_app_d}).}
    \label{fig:convergence}
\end{figure}
\vspace{-2em}

\noindent\textbf{Baselines and comparison.}

\noindent We compare \textsc{K-Sil} against established instance-weighted $k$-means variants.
As a reference, we use standard $k$-means.
To account for outliers via density-based instance weighting, we include LOF$k$-means and iLOF$k$-means (iterative LOF$k$-means)~\cite{lofkmeans}, two instance-weighted $k$-means variants in which Local Outlier Factor (LOF) scores are converted into sample weights that reduce the influence of low-density (outlying) points. LOF$k$-means computes LOF weights once and then fits weighted $k$-means, while iLOF$k$-means recomputes LOF weights within the current clusters across iterations (a cluster-wise LOF reweighting loop) before performing weighted centroid updates. 
We also include an object-weighting $k$-means baseline OW$k$-means~\cite{owkmeans}, which introduces silhouette per-point weights into the within-cluster objective; weights are obtained by rescaling each point’s silhouette width so that low-silhouette points receive larger penalties, while well-separated points receive smaller penalties. Since the original OW$k$-means is formulated in a relocation-style update with repeated weight/centroid updates, we use a computationally streamlined variant that retains the same silhouette-to-weight mapping but computes silhouettes using \textsc{K-Sil}’s centroid-distance proxy (\S\ref{subsec:sil}) and alternates reweighting with k-means weighted updates.
For each dataset, we set the number of clusters $k$ to the ground-truth number of classes reported in \S\ref{subsec:datasets}. We then evaluate each method over $10$ independent runs with different random initializations (defaults parameters for each method). For a given run (seed/random state), all methods are initialized from the same set of randomly chosen initial centroids, ensuring a fair comparison and isolating the effect of the update rules from favorable initializations.
To assess clustering performance, we compute the silhouette score (SIL), clustering accuracy (ACC), obtained after optimally matching cluster labels to class labels via the Hungarian algorithm, normalized mutual information (NMI), and the adjusted Rand index (ARI). Tab.~\ref{tab:all_metrics} reports mean values over runs. Full results, including 95\% Student-$t$ confidence intervals and mean runtime comparisons in seconds (s), are provided in Appendix~\ref{app:empirical} Tab.~\ref{app:tab:all_metrics}, with additional Adjusted Mutual Information and Davies--Bouldin results in Appendix Tab.~\ref{app:tab:all_metrics_ami_db}.

\medskip
\noindent\textbf{\emph{k}-means${++}$ initialization.}

\begin{figure}[h] 
\centering 
\vspace{-4mm}
\includegraphics[width=\linewidth]{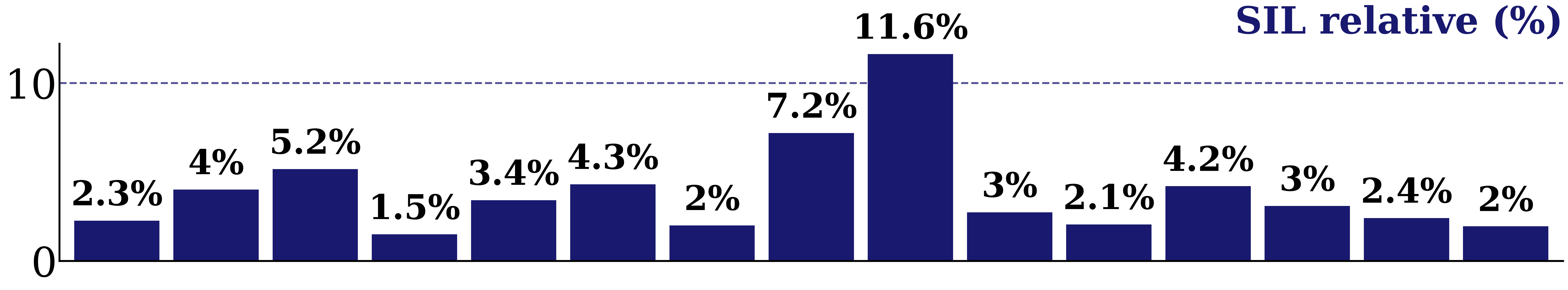}\par\vspace{2mm} \includegraphics[width=\linewidth]{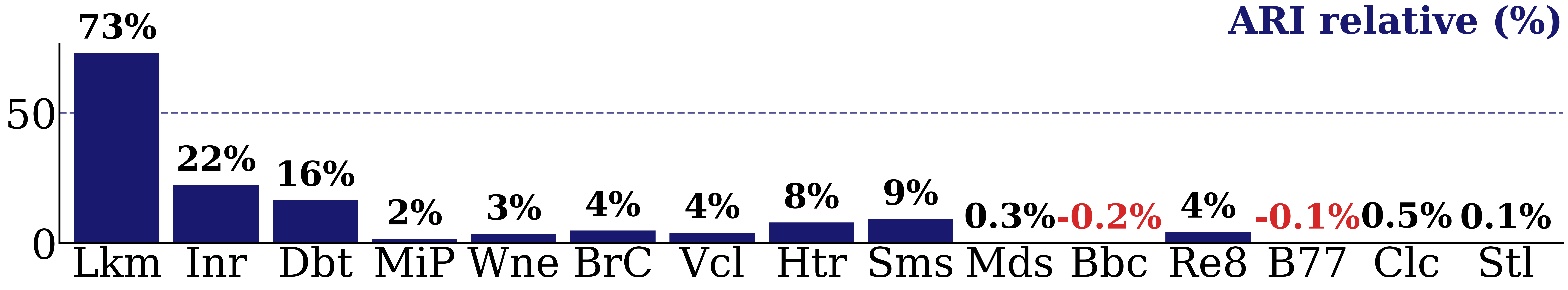}\par\vspace{2mm} \includegraphics[width=\linewidth]{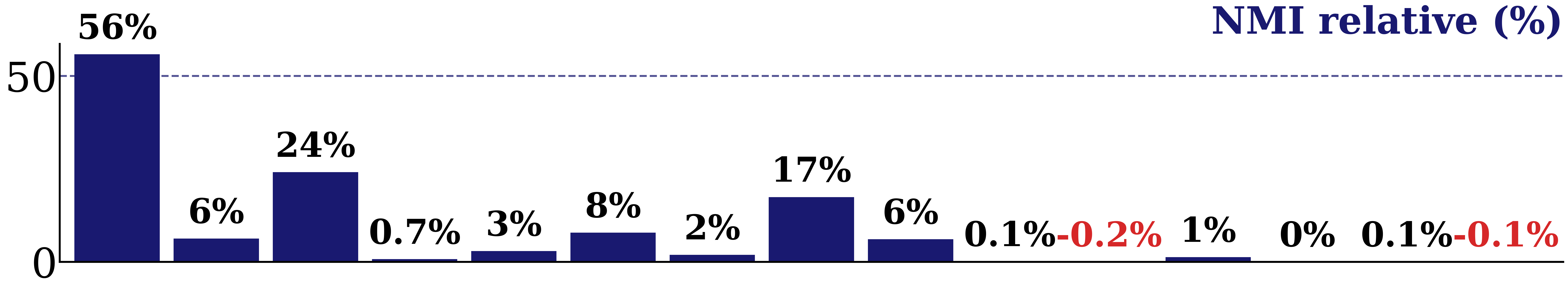}
\caption{\textcolor{Blue}{Mean relative change} (\% \textcolor{black}{$\uparrow$}  \textcolor{red}{$\downarrow$}) in SIL, ARI, NMI of \textsc{K-Sil} over \emph{k}-means (with $k$-means${++}$ initialization) over 120 independent runs across datasets (\S\ref{subsec:datasets}).} \label{fig:rel_improvplus} 
\vspace{-4mm}
\end{figure}

\noindent We additionally compare \textsc{K-Sil} to $k$-means in the $k$-means${++}$ seeding regime, reporting the relative change in mean SIL/ARI/NMI over 120 independent random states (Fig.~\ref{fig:rel_improvplus}). Since $k$-means++ substantially improves the quality of the $k$-means solution, improvements are expected to be more moderate; nevertheless, \textsc{K-Sil} is favorable on nearly all datasets and the few exceptions are negligible. Overall, improvements occur more often and are typically larger in magnitude than the rare, very small regressions.

\medskip
\noindent\textbf{Suboptimal \emph{k}.}

\noindent We examine the robustness of \textsc{K-Sil} when the number of clusters is misspecified: for each dataset with ground-truth $k^{\star}$, we run \textsc{K-Sil} with $k \in \{k^{\star}-5,\ldots,k^{\star}+5\}$ (restricting to $k\ge 2$ when $k^{\star}=2$), and repeat each setting over multiple random initializations/seeds to account for stochasticity. We then average the resulting scores across runs and datasets. We report external agreement via ARI and internal structure via the silhouette score (SIL) (Fig.~\ref{fig:suboptimal-k}; ACC in Appendix Fig.~\ref{fig:accsub}).

\begin{figure}[h]
\vspace{-2mm}
  \centering
  \includegraphics[width=0.9\linewidth]{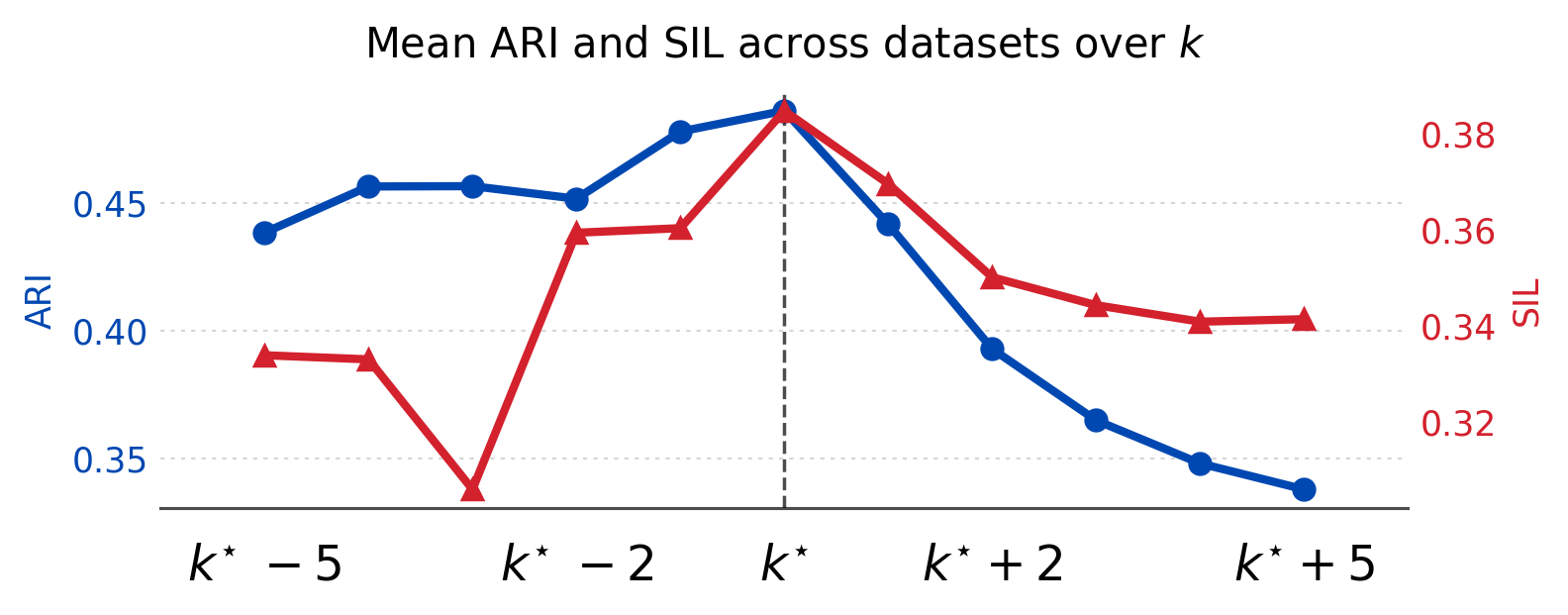}
  \caption{Mean \textcolor{blue}{ARI} and \textcolor{red}{SIL} across multiple random initializations and datasets (\S\ref{subsec:datasets}) as a function of the number of clusters $k\in\{k^\star-5, \ k^\star-4, \dots, \ k^\star, \dots, k^\star+4, \ k^\star+5 \}$.}
  \label{fig:suboptimal-k}
\end{figure}

\begingroup
\setlength{\tabcolsep}{1.5pt}          
\renewcommand{\arraystretch}{0.}

\scriptsize
\setlength{\LTcapwidth}{\linewidth}  

\begin{longtable}{@{} l l *{5}{>{\centering\arraybackslash}p{1.95cm}} @{}}
\caption[Clustering results]{\small{Mean SIL/ACC/NMI/ARI values of $k$-means, LOF$k$-means, iLOF$k$-means, OW$k$-means, and \textsc{K-Sil} over runs (95\% Student-$t$ confidence intervals and mean runtime comparisons in seconds in Appendix Tab.~\ref{app:tab:all_metrics}). Highest mean values in \textbf{bold}.}}
\label{tab:all_metrics}\\
\toprule
Dataset & Metric
& \makecell[c]{$k$-means}
& \makecell[c]{LOF\\$k$-means}
& \makecell[c]{iLOF\\$k$-means}
& \makecell[c]{OW\\$k$-means}
& \makecell[c]{\textsc{K-Sil}} \\
\midrule
\endfirsthead

\toprule
Dataset & Metric
& \makecell[c]{$k$-means}
& \makecell[c]{LOF\\$k$-means}
& \makecell[c]{iLOF\\$k$-means}
& \makecell[c]{OW\\$k$-means}
& \makecell[c]{\textsc{K-Sil}} \\
\midrule
\endhead

\bottomrule
\endlastfoot

\multirow{4}{*}{\textsc{Mip}}
& SIL & \makecell[c]{0.564}
      & \makecell[c]{0.560}
      & \makecell[c]{0.559}
      & \makecell[c]{0.556}
      & \makecell[c]{\textbf{0.612}} \\

& ACC & \makecell[c]{0.382}
      & \makecell[c]{0.381}
      & \makecell[c]{0.388}
      & \makecell[c]{0.383}
      & \makecell[c]{\textbf{0.393}} \\

& NMI & \makecell[c]{0.421}
      & \makecell[c]{0.420}
      & \makecell[c]{0.429}
      & \makecell[c]{0.419}
      & \makecell[c]{\textbf{0.444}} \\

& ARI & \makecell[c]{0.223}
      & \makecell[c]{0.221}
      & \makecell[c]{0.230}
      & \makecell[c]{0.222}
      & \makecell[c]{\textbf{0.243}} \\

\cmidrule(lr){2-7}

\multirow{4}{*}{\textsc{Wne}}
& SIL & \makecell[c]{0.411}
      & \makecell[c]{0.410}
      & \makecell[c]{0.418}
      & \makecell[c]{0.397}
      & \makecell[c]{\textbf{0.430}} \\

& ACC & \makecell[c]{0.931}
      & \makecell[c]{0.941}
      & \makecell[c]{0.933}
      & \makecell[c]{0.938}
      & \makecell[c]{\textbf{0.972}} \\

& NMI & \makecell[c]{0.839}
      & \makecell[c]{0.855}
      & \makecell[c]{0.843}
      & \makecell[c]{0.835}
      & \makecell[c]{\textbf{0.893}} \\

& ARI & \makecell[c]{0.846}
      & \makecell[c]{0.868}
      & \makecell[c]{0.854}
      & \makecell[c]{0.844}
      & \makecell[c]{\textbf{0.914}} \\

\cmidrule(lr){2-7}

\multirow{4}{*}{\textsc{BrC}}
& SIL & \makecell[c]{0.471}
      & \makecell[c]{0.467}
      & \makecell[c]{0.467}
      & \makecell[c]{0.453}
      & \makecell[c]{\textbf{0.492}} \\

& ACC & \makecell[c]{0.908}
      & \makecell[c]{0.909}
      & \makecell[c]{0.908}
      & \makecell[c]{0.907}
      & \makecell[c]{\textbf{0.919}} \\

& NMI & \makecell[c]{0.542}
      & \makecell[c]{0.546}
      & \makecell[c]{0.542}
      & \makecell[c]{0.542}
      & \makecell[c]{\textbf{0.594}} \\

& ARI & \makecell[c]{0.662}
      & \makecell[c]{0.666}
      & \makecell[c]{0.664}
      & \makecell[c]{0.660}
      & \makecell[c]{\textbf{0.700}} \\

\cmidrule(lr){2-7}

\multirow{4}{*}{\textsc{Htr}}
& SIL & \makecell[c]{0.684}
      & \makecell[c]{0.679}
      & \makecell[c]{0.684}
      & \makecell[c]{0.653}
      & \makecell[c]{\textbf{0.749}} \\

& ACC & \makecell[c]{0.943}
      & \makecell[c]{0.940}
      & \makecell[c]{0.943}
      & \makecell[c]{0.933}
      & \makecell[c]{\textbf{0.965}} \\

& NMI & \makecell[c]{0.442}
      & \makecell[c]{0.423}
      & \makecell[c]{0.442}
      & \makecell[c]{0.391}
      & \makecell[c]{\textbf{0.572}} \\

& ARI & \makecell[c]{0.635}
      & \makecell[c]{0.621}
      & \makecell[c]{0.635}
      & \makecell[c]{0.590}
      & \makecell[c]{\textbf{0.722}} \\

\cmidrule(lr){2-7}

\multirow{4}{*}{\textsc{Vcl}}
& SIL & \makecell[c]{0.587}
      & \makecell[c]{0.589}
      & \makecell[c]{0.588}
      & \makecell[c]{0.575}
      & \makecell[c]{\textbf{0.602}} \\

& ACC & \makecell[c]{0.413}
      & \makecell[c]{0.418}
      & \makecell[c]{0.413}
      & \makecell[c]{0.412}
      & \makecell[c]{\textbf{0.428}} \\

& NMI & \makecell[c]{0.163}
      & \makecell[c]{0.168}
      & \makecell[c]{0.163}
      & \makecell[c]{0.168}
      & \makecell[c]{\textbf{0.172}} \\

& ARI & \makecell[c]{0.116}
      & \makecell[c]{0.120}
      & \makecell[c]{0.117}
      & \makecell[c]{0.114}
      & \makecell[c]{\textbf{0.125}} \\

\cmidrule(lr){2-7}

\multirow{4}{*}{\textsc{Sms}}
& SIL & \makecell[c]{0.099}
      & \makecell[c]{0.099}
      & \makecell[c]{0.099}
      & \makecell[c]{0.093}
      & \makecell[c]{\textbf{0.111}} \\

& ACC & \makecell[c]{0.845}
      & \makecell[c]{0.845}
      & \makecell[c]{0.845}
      & \makecell[c]{0.834}
      & \makecell[c]{\textbf{0.860}} \\

& NMI & \makecell[c]{0.343}
      & \makecell[c]{0.343}
      & \makecell[c]{0.343}
      & \makecell[c]{0.328}
      & \makecell[c]{\textbf{0.364}} \\

& ARI & \makecell[c]{0.414}
      & \makecell[c]{0.414}
      & \makecell[c]{0.414}
      & \makecell[c]{0.387}
      & \makecell[c]{\textbf{0.452}} \\

\cmidrule(lr){2-7}

\multirow{4}{*}{\textsc{Mds}}
& SIL & \makecell[c]{0.338}
      & \makecell[c]{0.344}
      & \makecell[c]{0.347}
      & \makecell[c]{0.326}
      & \makecell[c]{\textbf{0.379}} \\

& ACC & \makecell[c]{0.801}
      & \makecell[c]{0.817}
      & \makecell[c]{0.793}
      & \makecell[c]{0.800}
      & \makecell[c]{\textbf{0.845}} \\

& NMI & \makecell[c]{0.876}
      & \makecell[c]{0.886}
      & \makecell[c]{0.878}
      & \makecell[c]{0.877}
      & \makecell[c]{\textbf{0.894}} \\

& ARI & \makecell[c]{0.767}
      & \makecell[c]{0.780}
      & \makecell[c]{0.759}
      & \makecell[c]{0.768}
      & \makecell[c]{\textbf{0.803}} \\

\cmidrule(lr){2-7}

\multirow{4}{*}{\textsc{Inr}}
& SIL & \makecell[c]{0.563}
      & \makecell[c]{0.575}
      & \makecell[c]{0.570}
      & \makecell[c]{0.548}
      & \makecell[c]{\textbf{0.606}} \\

& ACC & \makecell[c]{0.601}
      & \makecell[c]{0.632}
      & \makecell[c]{0.600}
      & \makecell[c]{0.638}
      & \makecell[c]{\textbf{0.684}} \\

& NMI & \makecell[c]{0.093}
      & \makecell[c]{0.094}
      & \makecell[c]{0.092}
      & \makecell[c]{0.100}
      & \makecell[c]{\textbf{0.104}} \\

& ARI & \makecell[c]{0.051}
      & \makecell[c]{0.080}
      & \makecell[c]{0.048}
      & \makecell[c]{0.089}
      & \makecell[c]{\textbf{0.133}} \\

\cmidrule(lr){2-7}

\multirow{4}{*}{\textsc{Dbt}}
& SIL & \makecell[c]{0.303}
      & \makecell[c]{0.306}
      & \makecell[c]{0.302}
      & \makecell[c]{0.269}
      & \makecell[c]{\textbf{0.316}} \\

& ACC & \makecell[c]{0.684}
      & \makecell[c]{0.684}
      & \makecell[c]{0.683}
      & \makecell[c]{0.688}
      & \makecell[c]{\textbf{0.699}} \\

& NMI & \makecell[c]{0.082}
      & \makecell[c]{0.083}
      & \makecell[c]{0.087}
      & \makecell[c]{0.091}
      & \makecell[c]{\textbf{0.108}} \\

& ARI & \makecell[c]{0.127}
      & \makecell[c]{0.127}
      & \makecell[c]{0.127}
      & \makecell[c]{0.139}
      & \makecell[c]{\textbf{0.150}} \\

\cmidrule(lr){2-7}

\multirow{4}{*}{\textsc{Bbc}}
& SIL & \makecell[c]{0.184}
      & \makecell[c]{0.182}
      & \makecell[c]{0.184}
      & \makecell[c]{0.178}
      & \makecell[c]{\textbf{0.188}} \\

& ACC & \makecell[c]{0.886}
      & \makecell[c]{0.872}
      & \makecell[c]{0.877}
      & \makecell[c]{0.876}
      & \makecell[c]{\textbf{0.896}} \\

& NMI & \makecell[c]{0.837}
      & \makecell[c]{0.822}
      & \makecell[c]{0.833}
      & \makecell[c]{0.828}
      & \makecell[c]{\textbf{0.839}} \\

& ARI & \makecell[c]{0.830}
      & \makecell[c]{0.810}
      & \makecell[c]{0.822}
      & \makecell[c]{0.815}
      & \makecell[c]{\textbf{0.837}} \\

\cmidrule(lr){2-7}

\multirow{4}{*}{\textsc{Lkm}}
& SIL & \makecell[c]{0.204}
      & \makecell[c]{0.174}
      & \makecell[c]{0.194}
      & \makecell[c]{0.198}
      & \makecell[c]{\textbf{0.226}} \\

& ACC & \makecell[c]{0.672}
      & \makecell[c]{0.672}
      & \makecell[c]{0.674}
      & \makecell[c]{0.639}
      & \makecell[c]{\textbf{0.717}} \\

& NMI & \makecell[c]{0.066}
      & \makecell[c]{0.111}
      & \makecell[c]{0.098}
      & \makecell[c]{0.048}
      & \makecell[c]{\textbf{0.123}} \\

& ARI & \makecell[c]{0.097}
      & \makecell[c]{0.123}
      & \makecell[c]{0.121}
      & \makecell[c]{0.064}
      & \makecell[c]{\textbf{0.171}} \\

\cmidrule(lr){2-7}

\multirow{4}{*}{\textsc{Re8}}
& SIL & \makecell[c]{0.242}
      & \makecell[c]{0.244}
      & \makecell[c]{0.242}
      & \makecell[c]{0.224}
      & \makecell[c]{\textbf{0.252}} \\

& ACC & \makecell[c]{0.491}
      & \makecell[c]{0.491}
      & \makecell[c]{0.482}
      & \makecell[c]{0.460}
      & \makecell[c]{\textbf{0.504}} \\

& NMI & \makecell[c]{0.519}
      & \makecell[c]{0.522}
      & \makecell[c]{0.520}
      & \makecell[c]{0.509}
      & \makecell[c]{\textbf{0.527}} \\

& ARI & \makecell[c]{0.349}
      & \makecell[c]{0.352}
      & \makecell[c]{0.344}
      & \makecell[c]{0.320}
      & \makecell[c]{\textbf{0.359}} \\

\cmidrule(lr){2-7}

\multirow{4}{*}{\textsc{B77}}
& SIL & \makecell[c]{0.304}
      & \makecell[c]{0.300}
      & \makecell[c]{0.304}
      & \makecell[c]{0.288}
      & \makecell[c]{\textbf{0.314}} \\

& ACC & \makecell[c]{0.546}
      & \makecell[c]{0.542}
      & \makecell[c]{0.545}
      & \makecell[c]{0.538}
      & \makecell[c]{\textbf{0.553}} \\

& NMI & \makecell[c]{0.734}
      & \makecell[c]{0.733}
      & \makecell[c]{\textbf{0.735}}
      & \makecell[c]{0.732}
      & \makecell[c]{\textbf{0.735}} \\

& ARI & \makecell[c]{0.444}
      & \makecell[c]{0.438}
      & \makecell[c]{\textbf{0.446}}
      & \makecell[c]{0.440}
      & \makecell[c]{\textbf{0.446}} \\

\cmidrule(lr){2-7}

\multirow{4}{*}{\textsc{Clc}}
& SIL & \makecell[c]{0.286}
      & \makecell[c]{0.288}
      & \makecell[c]{0.287}
      & \makecell[c]{0.274}
      & \makecell[c]{\textbf{0.294}} \\

& ACC & \makecell[c]{0.687}
      & \makecell[c]{0.687}
      & \makecell[c]{0.684}
      & \makecell[c]{0.677}
      & \makecell[c]{\textbf{0.693}} \\

& NMI & \makecell[c]{0.858}
      & \makecell[c]{0.858}
      & \makecell[c]{\textbf{0.859}}
      & \makecell[c]{0.855}
      & \makecell[c]{\textbf{0.859}} \\

& ARI & \makecell[c]{0.548}
      & \makecell[c]{0.548}
      & \makecell[c]{0.548}
      & \makecell[c]{0.540}
      & \makecell[c]{\textbf{0.552}} \\

\cmidrule(lr){2-7}

\multirow{4}{*}{\textsc{Stl}}
& SIL & \makecell[c]{0.257}
      & \makecell[c]{0.264}
      & \makecell[c]{0.258}
      & \makecell[c]{0.250}
      & \makecell[c]{\textbf{0.265}} \\

& ACC & \makecell[c]{0.843}
      & \makecell[c]{0.858}
      & \makecell[c]{0.840}
      & \makecell[c]{0.832}
      & \makecell[c]{\textbf{0.859}} \\

& NMI & \makecell[c]{0.894}
      & \makecell[c]{0.893}
      & \makecell[c]{0.894}
      & \makecell[c]{0.888}
      & \makecell[c]{\textbf{0.899}} \\

& ARI & \makecell[c]{0.820}
      & \makecell[c]{0.829}
      & \makecell[c]{0.819}
      & \makecell[c]{0.806}
      & \makecell[c]{\textbf{0.836}} \\

\end{longtable}
\endgroup

\noindent\textbf{Stress tests.}

\begin{figure}[h]
  \centering

  \begin{subfigure}[t]{0.49\linewidth}
    \centering
    \includegraphics[width=\linewidth,trim=0 0 0 28,clip]{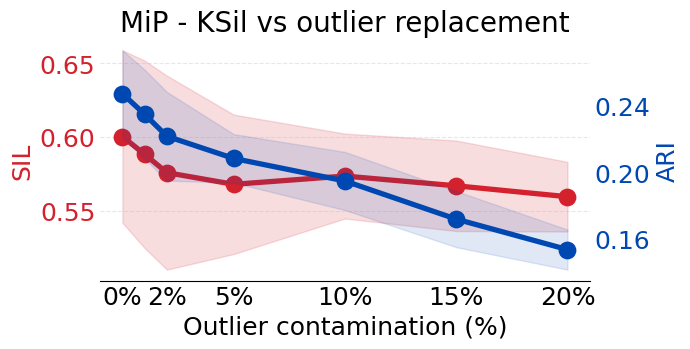}
    \caption{\textsc{MiP} (outlier replacement)}
    \label{fig:stress:mip:stor}
  \end{subfigure}\hfill
  \begin{subfigure}[t]{0.49\linewidth}
    \centering
    \includegraphics[width=\linewidth,trim=0 0 0 28,clip]{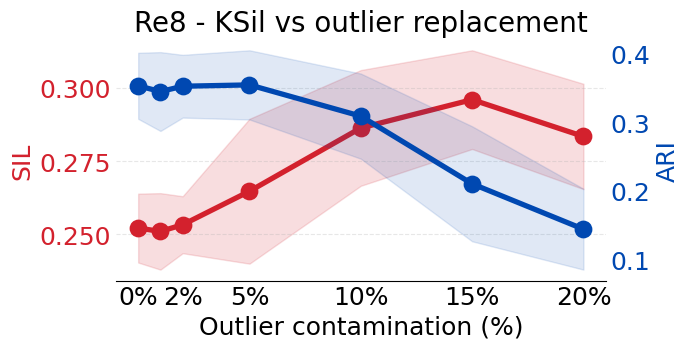}
    \caption{\textsc{Re8} (outlier replacement)}
    \label{fig:stress:re8:stor}
  \end{subfigure}

  \vspace{2mm}

  \begin{subfigure}[t]{0.49\linewidth}
    \centering
    \includegraphics[width=\linewidth,trim=0 0 0 28,clip]{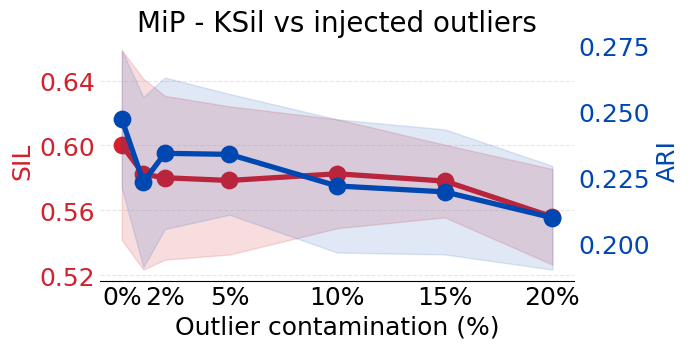}
    \caption{\textsc{MiP} (outlier injection)}
    \label{fig:stress:mip:stio}
  \end{subfigure}\hfill
  \begin{subfigure}[t]{0.49\linewidth}
    \centering
    \includegraphics[width=\linewidth,trim=0 0 0 28,clip]{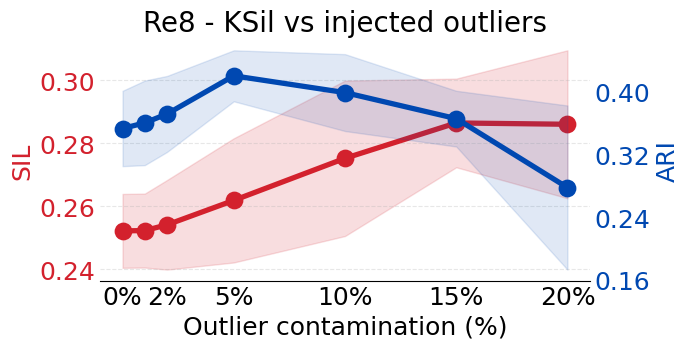}
    \caption{\textsc{Re8} (outlier injection)}
    \label{fig:stress:re8:stio}
  \end{subfigure}

  \caption{Stress tests on \textsc{MiP} and \textsc{Re8}. Top row (a), (b): outlier replacement; bottom row (c), (d): outlier injection. Mean \textcolor{red}{SIL} (left axis) and \textcolor{blue}{ARI} (right axis) vs. outlier contamination over runs \%; shaded areas indicate $\pm $ std over runs.}
  \label{fig:stress:all}
  \vspace{-5mm}
\end{figure}

\noindent To probe robustness to outlier ``contamination,'' we stress-test \textsc{K-Sil} on two representative datasets, \textsc{MiP} and \textsc{Re8}. We fix $k=k^{\star}$ and rerun \textsc{K-Sil} for increasing outlier levels $\varepsilon \in \{1,2,5,10,15,20\}\%$, averaging over multiple random initializations; for each $\varepsilon$ we report mean$\pm$std of SIL and ARI. We consider two protocols: (\textit{i}) \emph{outlier replacement}, which replaces an $\varepsilon$ fraction of original points with synthetic outliers (injecting noise while removing informative samples); and (\textit{ii}) \emph{outlier injection}, which appends $\varepsilon$ synthetic outliers while keeping all original points. In (\textit{ii}), ARI is computed on the original points only, directly measuring preservation of the clean-label structure. The resulting trends are summarized in Fig.~\ref{fig:stress:all} (\ref{fig:stress:mip:stor}, \ref{fig:stress:re8:stor}:  replacement; \ref{fig:stress:mip:stio}, \ref{fig:stress:re8:stio}: injection).
 
\noindent\textbf{Ablations.}

\begin{figure}[h]
\vspace{-6mm}
    \centering
    \begin{subfigure}[t]{0.49\linewidth}
        \centering
        \includegraphics[width=\linewidth]{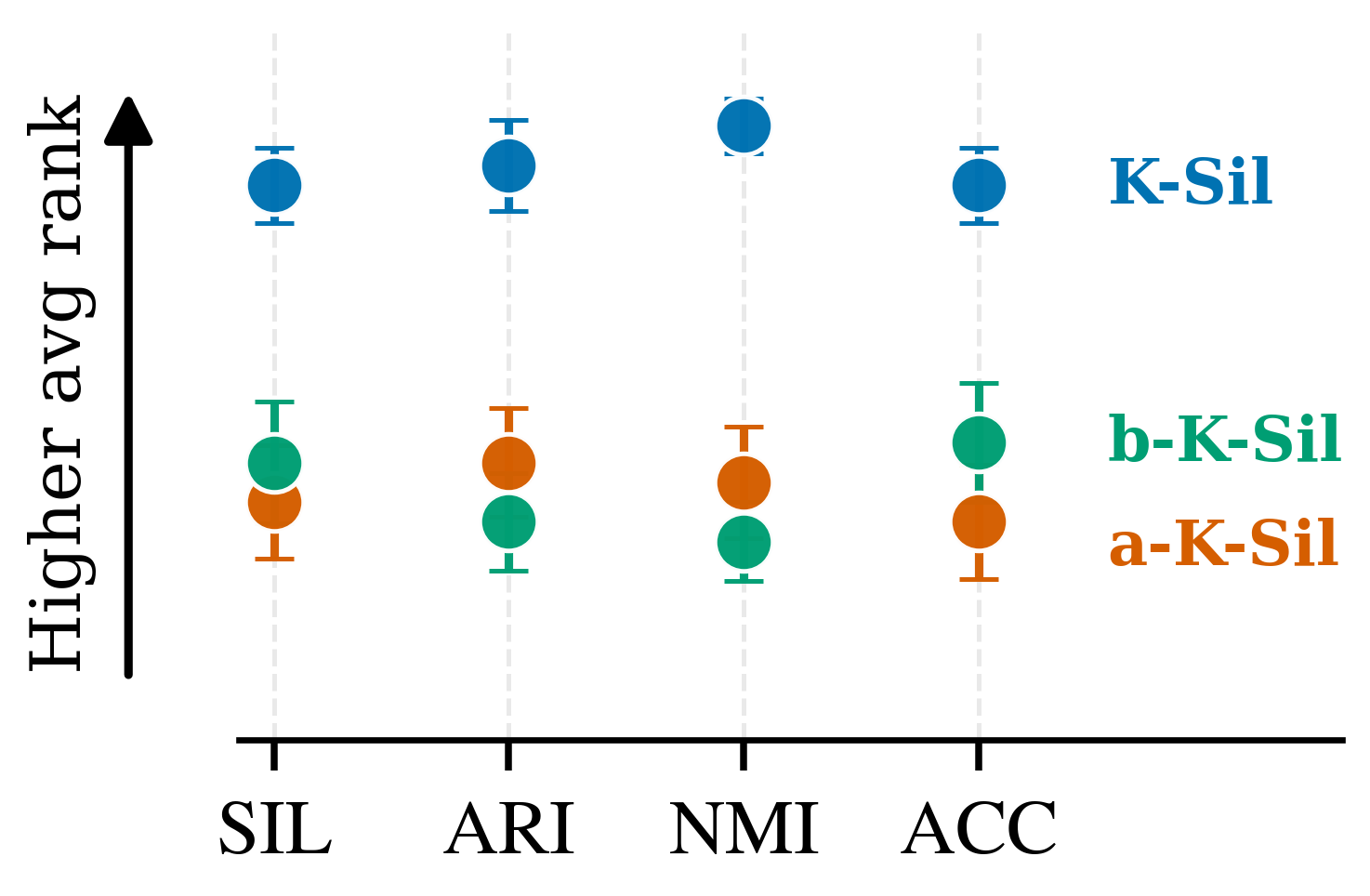}
        \caption{Weighting signal ablation.}
        \label{fig:ablation_i}
    \end{subfigure}\hfill
    \begin{subfigure}[t]{0.49\linewidth}
        \centering
        \includegraphics[width=\linewidth]{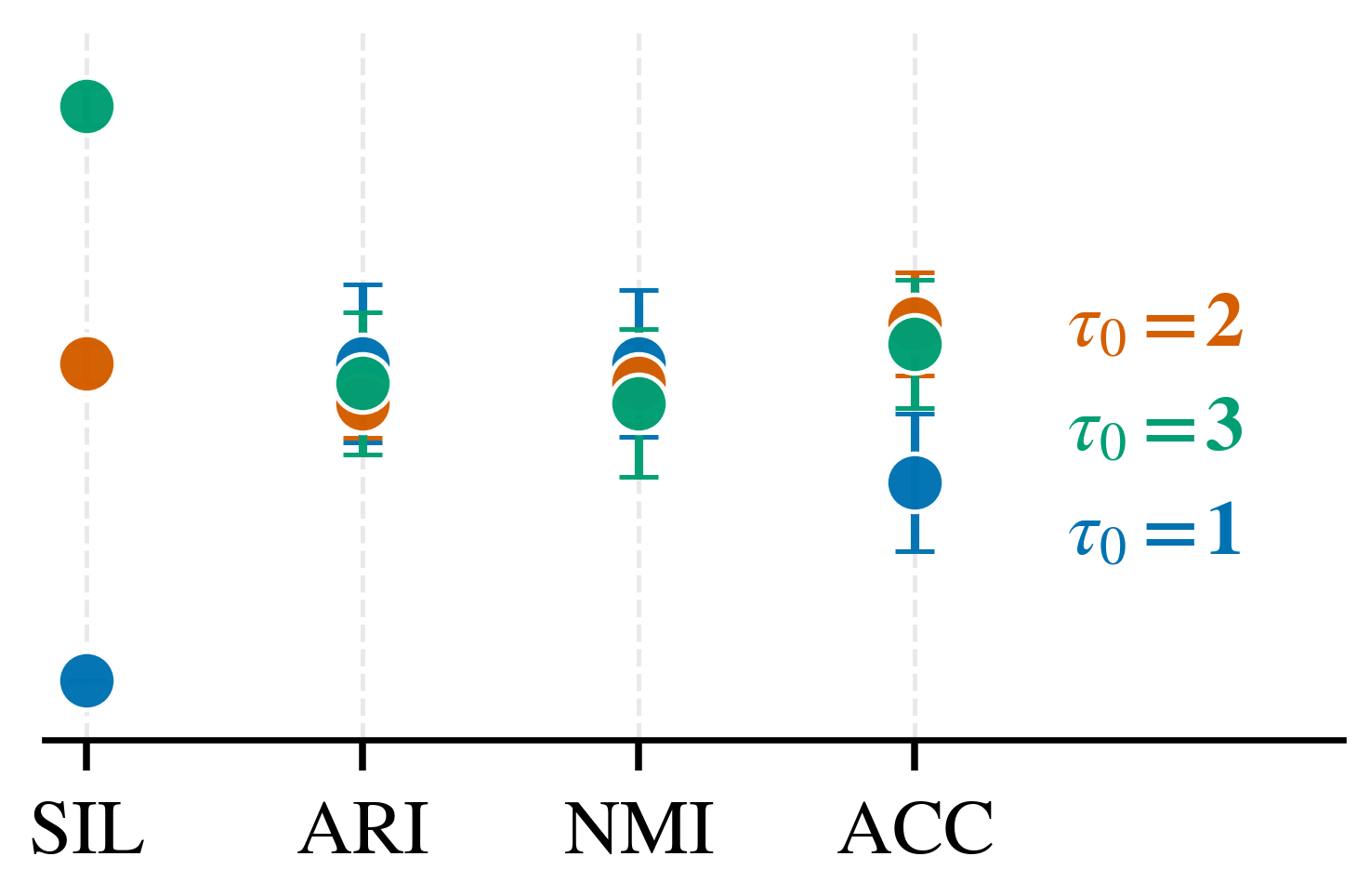}
        \caption{Initial temperature $\tau_0$ (\S\ref{subsec:temperature}) sweep.}
        \label{fig:ablation_ii}
    \end{subfigure}
    \caption{Dots show the average per-dataset rank (higher rank $\approx$ more frequent method wins) over $\mathrm{SIL}$, $\mathrm{ARI}$, $\mathrm{NMI}$, and $\mathrm{ACC}$; error bars denote SEM across multiple seeds.}
    \label{fig:ksil_ablations}
\end{figure}

\noindent We perform ablations to assess the effect of each design choice in \textsc{K-Sil}, modifying one component at a time while keeping the rest unchanged: (a) \textsc{K-Sil} uses weights $w_i \propto \exp(\tau s_i)$. We keep the same weighting form but replace the signal $s_i$ with its building blocks $a_i, b_i$ (Eq.~\ref{eq:weights}): a-\textsc{K-Sil} replaces $s_i$ with the intra-cluster term $a_i$ (using $1/(a_i+\varepsilon)$), emphasizing compactness, while b-\textsc{K-Sil} replaces $s_i$ with the inter-cluster term $b_i$, emphasizing separation (Fig.~\ref{fig:ablation_i}); (b) we sweep the initial temperature (\S\ref{subsec:temperature}) $\tau_0 \in \{1,2,3\}$ (Fig.~\ref{fig:ablation_ii}). An additional ablation over the learning rate $\eta$ (Eq.~\ref{eq:tauraw}) is deferred to Appendix Fig.~\ref{fig:ablationlr}.

\subsection{Results}\label{subsec:results}
\textsc{K-Sil} converges across all datasets and runs. The average centroid movement quickly decays and then plateaus, while the instance-weights become almost perfectly correlated across successive iterations (Fig.~\ref{fig:convergence}; Appendix Figs.~\ref{fig:convergence_app_a}--\ref{fig:convergence_app_d}).
\textsc{K-Sil} improves internal separation and typically strengthens external agreement. It yields consistent improvements over baselines across datasets and metrics (SIL/ACC/NMI/ARI in Tab.~\ref{tab:all_metrics} and AMI/DB in Appendix Tab.~\ref{app:tab:all_metrics_ami_db}; typically with confidence intervals that are tighter and centered at higher values Appendix Tab.~\ref{app:tab:all_metrics}).  
Gains are particularly pronounced on several datasets (e.g., \textsc{BrC}, \textsc{Htr}, \textsc{Wne}, \textsc{Mds}). On more challenging regimes, improvements are naturally more modest, but \textsc{K-Sil} remains stable and competitive, including \textsc{Lkm} ($d\!\gg\!n$) and large-$k$ benchmarks such as \textsc{B77} ($k{=}77$) and \textsc{Clc} ($k{=}151$). Runtime-wise, K-Sil adds modest overhead versus \emph{k}-means, yet is typically faster than iLOF-based reweighting (Appendix Tab.~\ref{app:tab:all_metrics}).
When comparing \textsc{K-Sil} to \emph{k}-means in the \emph{k}-means$++$ seeding regime, improvements largely persist despite the stronger baseline (Fig.~\ref{fig:rel_improvplus}). \textsc{K-Sil} improves on nearly every dataset, and the few declines are tiny (e.g., $\approx-0.1\%$ NMI on Bbc and Stl). Across datasets, gains are both more frequent and larger than the rare, minor regressions.
Under misspecified cluster counts, SIL and ARI peak around the ground-truth $k^\star$ across datasets (Fig.~\ref{fig:suboptimal-k}, with the same trend for ACC Appendix Fig.~\ref{fig:accsub}), indicating that performance is higher near $k^\star$, without a systematic bias toward over- or under-clustering. The robustness experiments (Fig.~\ref{fig:stress:all}) highlight complementary behaviors. Under outlier replacement (Fig.~\ref{fig:stress:mip:stor}), ARI typically degrades more strongly than SIL, showing that \textsc{K-Sil} can maintain geometric separation even when informative signal is progressively destroyed; on \textsc{Re8} we also observe silhouette inflation, illustrating that separation-based criteria may become overly optimistic under strong destructive corruption. Under outlier injection (Fig.~\ref{fig:stress:mip:stio}), \textsc{K-Sil} is markedly more stable, while SIL declines, reflecting that appended outliers do not fit the learned structure. Overall, \textsc{K-Sil} is more reliable under additive contamination, and the tests show that internal separation and label agreement may diverge under replacement noise.
The rank-based ablations (Fig.~\ref{fig:ablation_i}) show that \textsc{K-Sil} ranks best overall, consistently winning against compactness-only ($a_i$) and separation-only ($b_i$) variants across datasets and metrics (SIL/ARI/NMI/ACC), supporting the centroid-margin silhouette proxy as a balanced confidence signal.
The initial temperature ablation (Fig.~\ref{fig:ablation_ii}) shows that higher $\tau_0$ can make the early weighting more selective and may inflate SIL, while improvements in external metrics depend on the degree to which internal geometry aligns with label structure; overall, $\tau_0{=}1$ is a safe default for external agreement (our rule for choosing $\tau_0\!\in\!\{1,2\}$ is reported in Appendix~\ref{app:inittau}).

\section{Conclusion}
We introduced \textsc{K-Sil}, a silhouette-driven instance-weighted $k$-means variant that uses a centroid-margin silhouette proxy to turn geometric assignment confidence into a per-cluster instance-weighting distribution at each iteration.
By updating each centroid as a softmax-weighted mean of its assigned points and calibrating the sharpness of reweighting through an adaptive temperature guided by a macro-averaged silhouette criterion, \textsc{K-Sil} emphasizes reliable instances and suppresses ambiguous or noisy influence.
On the theory side, we established local convergence under standard separation conditions. Empirically, \textsc{K-Sil} consistently improves clustering quality across diverse datasets and remains competitive under strong baselines and stress settings.
These results highlight a broadly useful principle for improving centroid-based clustering: use within-iteration geometric signals to steer centroid updates toward reliably assigned structure. Future work includes principled selection or learning of representations and embedding spaces where silhouette-based geometry aligns most strongly with external-label structure, making \textsc{K-Sil}'s weighting most informative.

\section*{Declarations}

\begin{itemize}
\item Funding: This work was supported by the Archimedes Research Unit, Athena Research Center, through the project ``ARCHIMEDES Unit: Research in Artificial Intelligence, Data Science, and Algorithms'', implemented within the framework of the National Recovery and Resilience Plan ``Greece 2.0'' and funded by the European Union -- NextGenerationEU.
\item Conflict of interest/Competing interests: `Not applicable'
\item Ethics approval and consent to participate: `Not applicable'
\item Consent for publication: `Not applicable' 
\item Data availability: `Not applicable' 
\item Materials availability: `Not applicable' 
\item Code availability: Available at \url{https://github.com/semoglou/ksil_clustering}.
\item Author contribution: AS: Methodology; Formal Analysis; Software; Writing (original draft), AL: Methodology; Supervision; Writing (review/editing), JP: Conceptualization; Supervision; Writing (review/editing).
\end{itemize}

\bibliography{references}

@article{jain1999data,
  author = {Jain, A.K. and Murty, M.N. and Flynn, P.J.},
  title = {Data Clustering: A Review},
  journal = {ACM Computing Surveys},
  volume = {31},
  number = {3},
  pages = {264--323},
  year = {1999}
}

@article{xu2005survey,
  title={Survey of clustering algorithms},
  author={Xu, Rui and Wunsch, Donald},
  journal={IEEE Transactions on Neural Networks},
  volume={16},
  number={3},
  pages={645--678},
  year={2005},
  publisher={IEEE}
}

@article{rousseeuw1987,
  author = {Rousseeuw, P.J.},
  title = {Silhouettes: A graphical aid to the interpretation and validation of cluster analysis},
  journal = {Journal of Computational and Applied Mathematics},
  volume = {20},
  pages = {53--65},
  year = {1987},
  publisher = {Elsevier}
}

@InProceedings{pavlopoulos2024,
author="Pavlopoulos, John
and Vardakas, Georgios
and Likas, Aristidis",
title="Revisiting Silhouette Aggregation",
booktitle="Discovery Science",
year="2025",
publisher="Springer Nature Switzerland",
address="Cham",
pages="354--368",
isbn="978-3-031-78977-9"
}

@inproceedings{dudek2020silhouette,
  title={Silhouette index as clustering evaluation tool},
  author={Dudek, Andrzej},
  booktitle={Classification and Data Analysis: Theory and Applications 28},
  pages={19--33},
  year={2020},
  organization={Springer}
}

@inproceedings{lofkmeans,
author="Moggridge, Paul
and Helian, Na
and Sun, Yi
and Lilley, Mariana
and Veneziano, Vito",
editor="Iliadis, Lazaros
and Angelov, Plamen Parvanov
and Jayne, Chrisina
and Pimenidis, Elias",
title="Instance Weighted Clustering: Local Outlier Factor and K-Means",
booktitle="Proceedings of the 21st EANN (Engineering Applications of Neural Networks) 2020 Conference",
year="2020",
publisher="Springer International Publishing",
address="Cham",
pages="435--446",
isbn="978-3-030-48791-1"
}

@article{owkmeans,
author = {Gondeau, Alexandre and Aouabed, Zahia and Hijri, Mohamed and Peres-Neto, Pedro and Makarenkov, Vladimir},
year = {2021},
month = {03},
pages = {633-643},
title = {Object Weighting: A New Clustering Approach to Deal with Outliers and Cluster Overlap in Computational Biology},
volume = {18},
journal = {IEEE/ACM transactions on computational biology and bioinformatics / IEEE, ACM}
}

@inproceedings{macqueen1967,
  author = {MacQueen, J.B.},
  title = {Some methods for classification and analysis of multivariate observations},
  booktitle = {Proceedings of the 5th Berkeley Symposium on Mathematical Statistics and Probability},
  pages = {281--297},
  year = {1967}
}

@InProceedings{wangsilapprox,
author="Wang, Fei
and Franco-Penya, Hector-Hugo
and Kelleher, John D.
and Pugh, John
and Ross, Robert",
editor="Perner, Petra",
title="An Analysis of the Application of Simplified Silhouette to the Evaluation of k-means Clustering Validity",
booktitle="Machine Learning and Data Mining in Pattern Recognition",
year="2017",
publisher="Springer International Publishing",
address="Cham",
pages="291--305",
isbn="978-3-319-62416-7"
}

@inproceedings{kaufman2009finding,
  title={Finding Groups in Data: An Introduction to Cluster Analysis},
  author={Leonard Kaufman and Peter J. Rousseeuw},
  year={1990}
}

@article{bubeck2009,
author = {Bubeck, Sebastien and Meila, Marina and Luxburg, Ulrike},
year = {2009},
month = {07},
pages = {},
title = {How the initialization affects the stability of the k-means algorithm},
volume = {16},
journal = {ESAIM: Probability and Statistics}
}

@article{ikotun2022,
  title={K-means clustering algorithms: A comprehensive review, variants analysis, and advances in the era of big data},
  author={Abiodun Motunrayo Ikotun and Ezugwu E. Absalom and Laith Mohammad Abualigah and Belal Abuhaija and Heming Jia},
  journal={Inf. Sci.},
  year={2022},
  volume={622},
  pages={178-210}
}

@inproceedings{UCI,
  title={UCI Repository of machine learning databases},
  author={Catherine Blake},
  year={1998}
}

@article{Jain2008DataC5,
  title={Data clustering: 50 years beyond K-means},
  author={Anil K. Jain},
  journal={Pattern Recognit. Lett.},
  year={2008},
  volume={31},
  pages={651-666}
}

@inproceedings{ester1996,
  author = {Ester, M. and Kriegel, H.-P. and Sander, J. and Xu, X.},
  title = {A Density-Based Algorithm for Discovering Clusters in Large Spatial Databases with Noise},
  booktitle = {Proceedings of the 2nd International Conference on Knowledge Discovery and Data Mining (KDD)},
  pages = {226--231},
  year = {1996}
}

@article{mullner2011modern,
  author = {Müllner, D.},
  title = {Modern hierarchical, agglomerative clustering algorithms},
  journal = {arXiv preprint arXiv:1109.2378},
  year = {2011}
}

@article{reynolds2009gaussian,
  author = {Reynolds, D.A.},
  title = {Gaussian Mixture Models},
  journal = {Encyclopedia of Biometrics},
  pages = {659--663},
  year = {2009}
}

@inproceedings{ng2002spectral,
  author = {Ng, A.Y. and Jordan, M.I. and Weiss, Y.},
  title = {On Spectral Clustering: Analysis and an algorithm},
  booktitle = {Advances in Neural Information Processing Systems (NeurIPS)},
  volume = {14},
  year = {2002}
}

@article{likas2003global,
  title={The global k-means clustering algorithm},
  author={Likas, Aristidis and Vlassis, Nikos and Verbeek, Jakob J},
  journal={Pattern recognition},
  volume={36},
  number={2},
  pages={451--461},
  year={2003},
  publisher={Elsevier}
}

@article{vardakas2022global,
  title={Global $k$-means$++$: an effective relaxation of the global $k$-means clustering algorithm},
  author={Vardakas, Georgios and Likas, Aristidis},
  journal={arXiv preprint arXiv:2211.12271},
  year={2022}
}

@inproceedings{arthur2007kmeans++,
  title={K-means++ the advantages of careful seeding},
  author={Arthur, David and Vassilvitskii, Sergei},
  booktitle={Proceedings of the eighteenth annual ACM-SIAM symposium on Discrete algorithms},
  pages={1027--1035},
  year={2007}
}

@inproceedings{bachem2016afkmc2,
  author    = {Bachem, Olivier and Lucic, Mario and Krause, Andreas},
  title     = {Fast and Provably Good Seedings for k-Means},
  booktitle = {Advances in Neural Information Processing Systems (NeurIPS)},
  year      = {2016},
  pages     = {55--63}
}

@article{arbelaitz2013extensive,
  title={An extensive comparative study of cluster validity indices},
  author={Arbelaitz, Olatz and Gurrutxaga, Ibai and Muguerza, Javier and P{\'e}rez, Jes{\'u}s M and Perona, I{\~n}igo},
  journal={Pattern recognition},
  volume={46},
  number={1},
  pages={243--256},
  year={2013},
  publisher={Elsevier}
}

@article{bombina2024,
    author = {Bombina, Polina AND Tally, Dwayne AND Abrams, Zachary B. AND Coombes, Kevin R.},
    journal = {PLOS ONE},
    publisher = {Public Library of Science},
    title = {SillyPutty: Improved clustering by optimizing the silhouette width},
    year = {2024},
    month = {06},
    volume = {19},
    pages = {1-17},
    number = {6},
}

@article{batool2021clustering,
  title={Clustering with the average silhouette width},
  author={Batool, Fatima and Hennig, Christian},
  journal={Computational Statistics \& Data Analysis},
  volume={158},
  pages={107190},
  year={2021},
  publisher={Elsevier}
}

@article{huang2005,
  author = {Huang, Joshua Zhexue and Ng, Michael K. and Rong, Hongqiang and Li, Zichen},
  title = {Automated Variable Weighting in k-Means Type Clustering},
  journal = {IEEE Transactions on Pattern Analysis and Machine Intelligence},
  volume = {27},
  number = {5},
  pages = {657--668},
  year = {2005},
  publisher = {IEEE}
}

@article{jing2007,
  author = {Jing, Liping and Ng, Michael K. and Huang, Joshua Zhexue},
  title = {An entropy weighting k-means algorithm for subspace clustering of high-dimensional sparse data},
  journal = {IEEE Transactions on Knowledge and Data Engineering},
  volume = {19},
  number = {8},
  pages = {1026--1041},
  year = {2007},
  publisher = {IEEE}
}

@article{lai2024,
author = {Lai, Huixia and Huang, Tao and Lu, BinLong and Zhang, Shi and Xiaog, Ruliang},
year = {2024},
month = {12},
pages = {3061-3075},
title = {Silhouette coefficient-based weighting k-means algorithm},
volume = {37},
journal = {Neural Computing and Applications}
}

@article{chan2004optimization,
  author = {Chan, E. Y. and Ching, W. K. and Ng, Michael K. and Huang, Joshua Zhexue},
  title = {An optimization algorithm for clustering using weighted dissimilarity measures},
  journal = {Pattern Recognition},
  volume = {37},
  number = {5},
  pages = {943--952},
  year = {2004},
  publisher = {Elsevier}
}

@InProceedings{ilof,
author="Moggridge, Paul
and Helian, Na
and Sun, Yi
and Lilley, Mariana
and Veneziano, Vito",
editor="Iliadis, Lazaros
and Angelov, Plamen Parvanov
and Jayne, Chrisina
and Pimenidis, Elias",
title="Instance Weighted Clustering: Local Outlier Factor and K-Means",
booktitle="Proceedings of the 21st EANN (Engineering Applications of Neural Networks) 2020 Conference",
year="2020",
publisher="Springer International Publishing",
address="Cham",
pages="435--446"
}

\newpage

\begin{appendices}

\section{Theoretical results}\label{app:theory}

\noindent\textbf{Proof of L.1 (Stability near $\mathbf{\mu_\star}$).}

\begin{proof}
Fix $j$ and $x_i$ with $c_\star(i)=j$. Then $\|x_i - \mu_{\star,j}\| \le r$.
For any $\mu$ with $\|\mu - \mu_\star\|_{\infty,2} \le r$, we have $\|\mu_j - \mu_{\star,j}\| \le \|\mu - \mu_\star\|_{\infty,2} \le r$, hence
$$ \|x_i - \mu_j\|
\le \|x_i - \mu_{\star,j}\| + \|\mu_j - \mu_{\star,j}\|
\le r + r = 2r.$$
For any other cluster $\ell \neq j$,
$$\|x_i - \mu_\ell\|
\ge \|\mu_{\star,\ell} - \mu_{\star,j}\|
- \|x_i - \mu_{\star,j}\|
- \|\mu_\ell - \mu_{\star,\ell}\|
\ge 5r - r - r = 3r.$$
Thus $\mu_j$ remains the nearest centroid to $x_i$ for all $\mu \in U_\mu$, and therefore $c(\mu) = c_\star$.
Since $H_j(\mu,\tau)$ is a convex combination of $x_i$ for $i \in C_j = \{i:c_\star(i)=j\}$ and $\|x_i-\mu_{\star,j}\|\le r$ for $i\in C_j$ (A.1),
we have $\|H_j(\mu,\tau)-\mu_{\star,j}\|\le r$ $\forall j$ $\Rightarrow$ $\|H(\mu,\tau)-\mu_\star\|_{\infty,2}\le r$ and $H(\mu,\tau)\in U_\mu$.
\end{proof}

\medskip
\noindent\textbf{Proof of L.2 (Silhouette Lipschitz bound).}

\begin{proof}
Let $i$: $c_\star(i)=j$. By L.1: 
$$a_i(\mu)=\|x_i-\mu_j\|\le 2r,\ b_i(\mu)=\min_{\ell\neq j}\|x_i-\mu_\ell\|\ge 3r$$ for any $\mu\in U_\mu$. Also, by definition,
$$|a_i(\mu)-a_i(\mu')|\le \|\mu-\mu'\|_{\infty,2},\
|b_i(\mu)-b_i(\mu')|\le \|\mu-\mu'\|_{\infty,2}.$$
Since $s_i(\mu)= (b_i(\mu) - a_i(\mu))/b_i(\mu)=1-a_i(\mu)/b_i(\mu)$ (\S\ref{subsec:sil}) on $U_\mu$, we bound the ratio:
$$
\Big|\frac{a}{b}-\frac{a'}{b'}\Big|
\le \frac{|a-a'|}{b}+\frac{a'|b-b'|}{bb'}
\le \frac{1}{3r}\|\mu-\mu'\|_{\infty,2}+\frac{2r}{(3r)^2}\|\mu-\mu'\|_{\infty,2}
=\frac{5}{9r}\|\mu-\mu'\|_{\infty,2}.
$$
This proves the bound for $s_i$ (Eq.~\ref{eq:silhouette_score}). $S(\mu)$ (Eq.~\ref{eq:macro_silhouette}) is an average of the $s_i(\mu)$ with fixed nonnegative weights
(since clusters are fixed on $U_\mu$), so the same Lipschitz constant applies.
\end{proof}

\medskip
\noindent\textbf{Proof of L.3 (Softmax Lipschitz).}

\begin{proof}
Write $f(u)=\mathrm{softmax}(u)$, so $f:\mathbb{R}^m\to\mathbb{R}^m$ is a vector-valued map.
Fix $z\in\mathbb{R}^m$ and set $$p=f(z)=(p_1,\dots,p_m), \ \text{ i.e. } p_i=\exp(z_i)/\sum_{\ell=1}^m\exp(z_\ell)$$ The derivative of $f$ at $z$ is its Jacobian matrix
$J(z)\in\mathbb{R}^{m\times m}$ with entries $J_{i\ell}(z)=\partial p_i/\partial z_\ell$.
A direct computation gives the standard identity
$J_{i\ell}(z)=p_i(\delta_{i\ell}-p_\ell)$, which in matrix form is $J(z)=\mathrm{diag}(p)-pp^\top$,
where $\mathrm{diag}(p)$ is the diagonal matrix with diagonal $(p_1,\dots,p_m)$ and $(pp^\top)_{i\ell}=p_ip_\ell$.
To bound the operator norm $\|J(z)\|_{\infty\to1}$, let $v\in\mathbb{R}^m$ with $\|v\|_\infty\le1$.
Define the $p$--weighted average $\mu=\langle p,v\rangle=\sum_{i=1}^m p_i v_i$.
Using $J(z)=\mathrm{diag}(p)-pp^\top$ we obtain, for each coordinate $i$,
$$(J(z)v)_i = p_i v_i - p_i\sum_{\ell=1}^m p_\ell v_\ell
= p_i\bigl(v_i-\mu\bigr).$$
Hence
$\|J(z)v\|_1=\sum_{i=1}^m |(J(z)v)_i|
=\sum_{i=1}^m p_i\,|v_i-\mu|.$
We now maximize this quantity over the cube $\{v:\|v\|_\infty\le1\}$.
The map $v\mapsto \sum_i p_i|v_i-\langle p,v\rangle|$ is convex in $v$
(absolute value is convex and $\langle p,v\rangle$ is linear), so its maximum over a compact convex set
is attained at an extreme point of the set; here the extreme points are exactly the sign vectors
$v\in\{\pm1\}^m$. Fix such a $v\in\{\pm1\}^m$ and write
$p_+=\sum_{i:v_i=1}p_i$ and $p_-=\sum_{i:v_i=-1}p_i=1-p_+$.
Then the weighted average is $\mu=\langle p,v\rangle=p_+-p_-=2p_+-1$.
Moreover, for indices with $v_i=1$ we have $|v_i-\mu|=|1-\mu|$, and for $v_i=-1$ we have $|v_i-\mu|=|-1-\mu|$,
so
\[
\sum_{i=1}^m p_i|v_i-\mu|
= p_+|1-\mu| + p_-|-1-\mu|
= p_+\cdot 2(1-p_+) + p_-\cdot 2p_+
=4p_+p_- \le 1,
\]
since $p_+p_-\le \frac14$. Therefore $\|J(z)v\|_1\le 1$ for all $\|v\|_\infty\le1$, i.e.
$\|J(z)\|_{\infty\to1}\le1$ for all $z$.
Finally, apply the mean value theorem for vector-valued functions along the line segment from $u'$ to $u$ and taking $\ell_1$ norms and using the operator-norm bound gives:
$$f(u)-f(u')=\int_0^1 J\bigl(u'+t(u-u')\bigr)(u-u')\,dt.$$
$$\|f(u)-f(u')\|_1
\le \int_0^1 \|J(u'+t(u-u'))\|_{\infty\to1}\,dt\;\|u-u'\|_\infty
\le \|u-u'\|_\infty,$$
\end{proof}

\medskip
\noindent\textbf{Proof of L.4 (Lipschitz bounds for the centroid map $H$).}

\begin{proof}
Fix a cluster $j$. Since $\mu,\mu'\in U_\mu$, Lemma L.1 implies the labels are constant on $U_\mu$,
hence $C_j(\mu)=C_j(\mu')=:C_j$. Write the weighted update as
\[
H_j(\mu,\tau)=\sum_{i\in C_j} p_i(\mu,\tau)\,x_i,
\qquad
p(\mu,\tau)=\mathrm{softmax}\!\Big(\tau\,[s_i(\mu)]_{i\in C_j}\Big),
\]
so that $\sum_{i\in C_j}p_i(\mu,\tau)=1$.

\smallskip
\noindent(i) Lipschitz in $\mu$ (fixed $\tau$).
Using $\sum_{i\in C_j}(p_i(\mu,\tau)-p_i(\mu',\tau))=0$ and centering at $\mu_{\star,j}$,
\[
H_j(\mu,\tau)-H_j(\mu',\tau)
=\sum_{i\in C_j}\big(p_i(\mu,\tau)-p_i(\mu',\tau)\big)\,(x_i-\mu_{\star,j}).
\]
By A.1, $\|x_i-\mu_{\star,j}\|\le r$ for all $i\in C_j$, hence
\[
\|H_j(\mu,\tau)-H_j(\mu',\tau)\|
\le \sum_{i\in C_j}|p_i(\mu,\tau)-p_i(\mu',\tau)|\,\|x_i-\mu_{\star,j}\|
\le r\,\|p(\mu,\tau)-p(\mu',\tau)\|_1.
\]
Let $u(\mu)=\tau\,[s_i(\mu)]_{i\in C_j}$ be the logits. Lemma L.3 gives
\[
\|p(\mu,\tau)-p(\mu',\tau)\|_1
\le \|u(\mu)-u(\mu')\|_\infty
\le \tau\,\max_{i\in C_j}|s_i(\mu)-s_i(\mu')|.
\]
Applying Lemma L.2 yields
$\max_{i\in C_j}|s_i(\mu)-s_i(\mu')|\le \frac{5}{9r}\|\mu-\mu'\|_{\infty,2}$, so
\[
\|H_j(\mu,\tau)-H_j(\mu',\tau)\|
\le r\cdot \tau\cdot \frac{5}{9r}\,\|\mu-\mu'\|_{\infty,2}
=\frac{5\tau}{9}\,\|\mu-\mu'\|_{\infty,2}.
\]
Taking $\max_{1\le j\le k}$ gives the first claim:
$\|H(\mu,\tau)-H(\mu',\tau)\|_{\infty,2}\le \frac{5\tau}{9}\|\mu-\mu'\|_{\infty,2}$.

\smallskip
\noindent(ii) Lipschitz in $\tau$ (fixed $\mu$).
Fix $\mu$ and consider
\[
H_j(\mu,\tau)-H_j(\mu,\tau')
=\sum_{i\in C_j}\big(p_i(\mu,\tau)-p_i(\mu,\tau')\big)\,(x_i-\mu_{\star,j}),
\]
hence $\|H_j(\mu,\tau)-H_j(\mu,\tau')\|\le r\,\|p(\mu,\tau)-p(\mu,\tau')\|_1$ as above.
With logits $u=\tau\,[s_i(\mu)]_{i\in C_j}$ and $u'=\tau'\,[s_i(\mu)]_{i\in C_j}$, Lemma L.3 gives
\[
\|p(\mu,\tau)-p(\mu,\tau')\|_1 \le \|u-u'\|_\infty
=|\tau-\tau'|\,\|[s_i(\mu)]_{i\in C_j}\|_\infty
\le |\tau-\tau'|,
\]
since $s_i(\mu)\in[0,1]$. Therefore
$$\|H_j(\mu,\tau)-H_j(\mu,\tau')\|\le r|\tau-\tau'|,$$ and taking $\max_j$ gives
$$\|H(\mu,\tau)-H(\mu,\tau')\|_{\infty,2}\le r|\tau-\tau'|.$$
\end{proof}

\medskip
\noindent\textbf{Proof of Theorem (Local convergence under bounded temperature).}
\begin{proof}
(i) Invariance and fixed labels.\\
Since $\mu_0\in U_\mu$, Lemma~L.1 gives $c(\mu_0)=c_\star$ and $\mu_1=H(\mu_0,\tau_1)\in U_\mu$.\\
Repeating the same argument inductively yields $\mu_t\in U_\mu$ and $c(\mu_t)=c_\star$ for all $t$.\\
In particular, the index sets $C_j(\mu_t)$ (hence the sizes $n_j(\mu_t)$ and the bound $\tau_{\max}(t)$) are constant on $U_\mu$.\\
(ii) One-step contraction for the centroid increments.\\
Using $\mu_{t+1}=H(\mu_t,\tau_{t+1})$ and $\mu_t=H(\mu_{t-1},\tau_t)$,
$$
H(\mu_t,\tau_{t+1})-H(\mu_{t-1},\tau_t)
=\underbrace{\big[H(\mu_t,\tau_{t+1})-H(\mu_{t-1},\tau_{t+1})\big]}_{(A)}
+\underbrace{\big[H(\mu_{t-1},\tau_{t+1})-H(\mu_{t-1},\tau_t)\big]}_{(B)}.
$$
By Lemma~L.4 and $\tau_{t+1}\le \bar\tau$ (A.2(i)): 
$$ \ \|(A)\|_{\infty,2}\le \frac{5\tau_{t+1}}{9}\,\|\mu_t-\mu_{t-1}\|_{\infty,2}
\le \frac{5\bar\tau}{9}\,d_t.$$
Also by Lemma~L.4, $$\|(B)\|_{\infty,2}\le r\,|\tau_{t+1}-\tau_t|.$$
We write the raw multiplicative update (Eq.~\ref{eq:tauraw}) as
$\tau_{t+1}^{\mathrm{raw}}=\tau_t\exp\{\eta\,R(S_t,S_{t-1})\}$,
and recall $\tau_{t+1}=\mathrm{clip}_{[\tau_{\min},\,\tau_{\max}(t)]}(\tau_{t+1}^{\mathrm{raw}})$ (Eq.~\ref{eq:Psi_map}).\\
Since $\tau_t\in[\tau_{\min},\tau_{\max}(t)]$ (A.2(i)), lower clipping is non-expansive, so
$$
|\tau_{t+1}-\tau_t|
\le |\tau_{t+1}^{\mathrm{raw}}-\tau_t|
=\tau_t\big|\exp\{\eta\,R(S_t,S_{t-1})\}-1\big|.
$$
Using convexity of $r\mapsto e^{\eta r}$ on $[-1,1]$, for all $|r|\le 1$ we have
$|e^{\eta r}-1|\le (e^\eta-1)|r|$.\\ 
Since $|R(S_t,S_{t-1})|\le 1$, it follows that
$$|\tau_{t+1}-\tau_t|
\le \bar\tau\,(e^\eta-1)\,|R(S_t,S_{t-1})|.$$
Under A.2(ii)--(iii), $R(S_t,S_{t-1})=\tilde r(S_t,S_{t-1})$, hence
$$
|R(S_t,S_{t-1})|
=\frac{|S_t-S_{t-1}|}{1-S_{t-1}+\epsilon}
\le \frac{1}{\alpha}\,|S(\mu_t)-S(\mu_{t-1})|
\le \frac{L_s}{\alpha}\,d_t,
$$
where we used Lemma~L.2. Combining the bounds and $L_s=\frac{5}{9r}$ gives
$$
d_{t+1}\le \Big[\frac{5\bar\tau}{9}+r\,\bar\tau(e^\eta-1)\frac{L_s}{\alpha}\Big]d_t
=\frac{5\bar\tau}{9}\Big(1+\frac{e^\eta-1}{\alpha}\Big)d_t
=\bar q\,d_t.
$$
Since, $\bar q =\frac{5\bar\tau}{9}\Big(1+\frac{e^\eta-1}{\alpha}\Big)\;<\;1$, proving (ii).\\
(iii) Convergence of $\mu_t$, $S_t$, and $\tau_t$.
From (ii), $d_t\le \bar q^{\,t-1}d_1$, hence $\sum_{t\ge 1} d_t<\infty$ and $(\mu_t)$ is Cauchy in $\|\cdot\|_{\infty,2}$.
Since $U_\mu$ is closed, $\mu_t\to\mu_\infty\in U_\mu$.
By continuity of $S(\mu)$ on $U_\mu$ (labels fixed by (i)), we have $S_t=S(\mu_t)\to S(\mu_\infty)=S_\infty$.\\
For the temperature, the bound above yields
$|\tau_{t+1}-\tau_t|
\le \bar\tau (e^\eta-1)\frac{L_s}{\alpha}\,d_t,$
and the right-hand side is summable since $\sum_t d_t<\infty$. Thus $\sum_t|\tau_{t+1}-\tau_t|<\infty$,
so $(\tau_t)$ is Cauchy and $\tau_t\to\tau_\infty\in[\tau_{\min},\bar\tau]$.
Moreover, by continuity of $H$ and $\Psi_\eta$ on $U_\mu\times[0,\bar\tau]$ (labels fixed by (i)),
the limits satisfy $\mu_\infty=H(\mu_\infty,\tau_\infty)$ and $\tau_\infty=\Psi_\eta(\tau_\infty,S_\infty,S_\infty)$.

\end{proof}

\section{Empty-cluster re-initialization strategy}\label{app:empty}

After the nearest-centroid reassignment (Eq.~\ref{eq:assignment}) at iteration $t+1$, define the index set and size of cluster $j$ as
$C_j(\mu_{t+1})=\{i: c(\mu_{t+1})(i)=j\},
\ \
n_j(\mu_{t+1})=|C_j(\mu_{t+1})|.$
If a cluster becomes empty, i.e.\ $n_j(\mu_{t+1})=0$, we re-initialize its centroid by selecting a point
from the currently largest cluster (by size) that lies farthest from that cluster’s centroid.
Formally, let
$j_{\max}\in \arg\max_{\ell\in\{1,\dots,k\}} n_\ell(\mu_{t+1})$
denote an index of a largest cluster, and let
$i^\star \in \arg\max_{i\in C_{j_{\max}}(\mu_{t+1})}\|x_i-\mu_{t+1,j_{\max}}\|$
be the index of the point farthest from its centroid within that largest cluster. We then re-seed the empty cluster by setting
$\mu_{t+1,j} \leftarrow x_{i^\star}$ for each $j$ such that $n_j(\mu_{t+1})=0$.
Optionally (and in our implementation), we also assign this point to the re-initialized cluster,
$c(\mu_{t+1})(i^\star)\leftarrow j$, to ensure the cluster is non-empty immediately.
In addition to truly empty clusters, our implementation applies the \emph{same} re-initialization rule to clusters that are \emph{too small}, i.e.\ those satisfying $n_j(\mu_{t+1})<3$. This is a safeguard introduced to maintain the standing assumption used in \S\ref{subsec:temperature} that cluster sizes remain at least three, $n_j(\mu_t)\ge 3$ for all $j,t$: when a cluster has size $1$ or $2$, its per-cluster silhouette aggregation (Eq.~\ref{eq:macro_silhouette})
and the temperature dynamics driven by it can become overly noisy and unrepresentative. Thus, whenever $n_j(\mu_{t+1})<3$ occurs, we treat the cluster as effectively empty and re-seed it using the procedure above.
This strategy is based on the intuition that points farthest from the dominant cluster’s center are likely to represent under-modeled structure or outlying regions; re-seeding at such a point helps restore the
intended number of clusters while exploring uncovered regions. In all experiments reported in \S\ref{sec:empirical}, no cluster became empty and no cluster satisfied
$n_j(\mu_t)<3$ at any iteration; the above procedure is included purely as a safeguard.

\section{Empirical Results}\label{app:empirical}

\begingroup
\setlength{\tabcolsep}{2.5pt}          
\renewcommand{\arraystretch}{1.2}
\scriptsize
\setlength{\LTcapwidth}{\linewidth}  

\begin{longtable}{@{} l l *{5}{>{\centering\arraybackslash}p{1.95cm}} @{}}
\caption[Clustering results]{Mean SIL/ACC/NMI/ARI values with 95\% $t$-confidence intervals, along with mean runtime in seconds (s) per dataset over runs. Gray ranges (\textcolor{black!65}{\(\cdot\)--\(\cdot\)}) show confidence intervals; highest mean values are in \textbf{bold}. \\All experiments were run in a cloud-hosted notebook with 51\,GB RAM.}
\label{app:tab:all_metrics}\\
\toprule
Dataset & Metric
& \makecell[c]{$k$-means}
& \makecell[c]{LOF\\$k$-means}
& \makecell[c]{iLOF\\$k$-means}
& \makecell[c]{OW\\$k$-means}
& \makecell[c]{\textsc{K-Sil}} \\
\midrule
\endfirsthead

\toprule
Dataset & Metric
& \makecell[c]{$k$-means}
& \makecell[c]{LOF\\$k$-means}
& \makecell[c]{iLOF\\$k$-means}
& \makecell[c]{OW\\$k$-means}
& \makecell[c]{\textsc{K-Sil}} \\
\midrule
\endhead

\bottomrule
\endlastfoot

\multirow{5}{*}{\textsc{Mip}}
& SIL & \makecell[c]{0.564\\{\scriptsize\textcolor{black!65}{0.539--0.589}}}
      & \makecell[c]{0.560\\{\scriptsize\textcolor{black!65}{0.534--0.585}}}
      & \makecell[c]{0.559\\{\scriptsize\textcolor{black!65}{0.533--0.586}}}
      & \makecell[c]{0.556\\{\scriptsize\textcolor{black!65}{0.531--0.581}}}
      & \makecell[c]{\textbf{0.612}\\{\scriptsize\textcolor{black!65}{0.593--0.631}}} \\

& ACC & \makecell[c]{0.382\\{\scriptsize\textcolor{black!65}{0.373--0.390}}}
      & \makecell[c]{0.381\\{\scriptsize\textcolor{black!65}{0.370--0.392}}}
      & \makecell[c]{0.388\\{\scriptsize\textcolor{black!65}{0.377--0.398}}}
      & \makecell[c]{0.383\\{\scriptsize\textcolor{black!65}{0.374--0.391}}}
      & \makecell[c]{\textbf{0.393}\\{\scriptsize\textcolor{black!65}{0.385--0.400}}} \\

& NMI & \makecell[c]{0.421\\{\scriptsize\textcolor{black!65}{0.409--0.433}}}
      & \makecell[c]{0.420\\{\scriptsize\textcolor{black!65}{0.406--0.433}}}
      & \makecell[c]{0.429\\{\scriptsize\textcolor{black!65}{0.419--0.440}}}
      & \makecell[c]{0.419\\{\scriptsize\textcolor{black!65}{0.406--0.432}}}
      & \makecell[c]{\textbf{0.444}\\{\scriptsize\textcolor{black!65}{0.434--0.454}}} \\

& ARI & \makecell[c]{0.223\\{\scriptsize\textcolor{black!65}{0.211--0.234}}}
      & \makecell[c]{0.221\\{\scriptsize\textcolor{black!65}{0.209--0.232}}}
      & \makecell[c]{0.230\\{\scriptsize\textcolor{black!65}{0.218--0.241}}}
      & \makecell[c]{0.222\\{\scriptsize\textcolor{black!65}{0.209--0.234}}}
      & \makecell[c]{\textbf{0.243}\\{\scriptsize\textcolor{black!65}{0.232--0.254}}}
      \\
      
& Runtime (s)
      & \makecell[c]{0.002}
      & \makecell[c]{0.006}
      & \makecell[c]{0.064}
      & \makecell[c]{0.020}
      & \makecell[c]{0.042} \\

\cmidrule(lr){2-7}

\multirow{5}{*}{\textsc{Wne}}
& SIL & \makecell[c]{0.411\\{\scriptsize\textcolor{black!65}{0.402--0.420}}}
      & \makecell[c]{0.410\\{\scriptsize\textcolor{black!65}{0.394--0.427}}}
      & \makecell[c]{0.418\\{\scriptsize\textcolor{black!65}{0.409--0.427}}}
      & \makecell[c]{0.397\\{\scriptsize\textcolor{black!65}{0.389--0.406}}}
      & \makecell[c]{\textbf{0.430}\\{\scriptsize\textcolor{black!65}{0.430--0.430}}} \\

& ACC & \makecell[c]{0.931\\{\scriptsize\textcolor{black!65}{0.885--0.978}}}
      & \makecell[c]{0.941\\{\scriptsize\textcolor{black!65}{0.899--0.983}}}
      & \makecell[c]{0.933\\{\scriptsize\textcolor{black!65}{0.886--0.981}}}
      & \makecell[c]{0.938\\{\scriptsize\textcolor{black!65}{0.901--0.973}}}
      & \makecell[c]{\textbf{0.972}\\{\scriptsize\textcolor{black!65}{0.972--0.973}}} \\

& NMI & \makecell[c]{0.839\\{\scriptsize\textcolor{black!65}{0.791--0.886}}}
      & \makecell[c]{0.855\\{\scriptsize\textcolor{black!65}{0.809--0.900}}}
      & \makecell[c]{0.843\\{\scriptsize\textcolor{black!65}{0.796--0.890}}}
      & \makecell[c]{0.835\\{\scriptsize\textcolor{black!65}{0.800--0.870}}}
      & \makecell[c]{\textbf{0.893}\\{\scriptsize\textcolor{black!65}{0.891--0.894}}} \\

& ARI & \makecell[c]{0.846\\{\scriptsize\textcolor{black!65}{0.784--0.909}}}
      & \makecell[c]{0.868\\{\scriptsize\textcolor{black!65}{0.811--0.926}}}
      & \makecell[c]{0.854\\{\scriptsize\textcolor{black!65}{0.792--0.916}}}
      & \makecell[c]{0.844\\{\scriptsize\textcolor{black!65}{0.798--0.889}}}
      & \makecell[c]{\textbf{0.914}\\{\scriptsize\textcolor{black!65}{0.913--0.916}}} \\
      
& Runtime (s)
      & \makecell[c]{0.001}
      & \makecell[c]{0.003}
      & \makecell[c]{0.023}
      & \makecell[c]{0.013}
      & \makecell[c]{0.018} \\
      
\cmidrule(lr){2-7}

\multirow{5}{*}{\textsc{BrC}}
& SIL & \makecell[c]{0.471\\{\scriptsize\textcolor{black!65}{0.470--0.472}}}
      & \makecell[c]{0.467\\{\scriptsize\textcolor{black!65}{0.466--0.467}}}
      & \makecell[c]{0.467\\{\scriptsize\textcolor{black!65}{0.467--0.467}}}
      & \makecell[c]{0.453\\{\scriptsize\textcolor{black!65}{0.453--0.454}}}
      & \makecell[c]{\textbf{0.492}\\{\scriptsize\textcolor{black!65}{0.492--0.493}}} \\

& ACC & \makecell[c]{0.908\\{\scriptsize\textcolor{black!65}{0.905--0.910}}}
      & \makecell[c]{0.909\\{\scriptsize\textcolor{black!65}{0.908--0.910}}}
      & \makecell[c]{0.908\\{\scriptsize\textcolor{black!65}{0.907--0.909}}}
      & \makecell[c]{0.907\\{\scriptsize\textcolor{black!65}{0.905--0.910}}}
      & \makecell[c]{\textbf{0.919}\\{\scriptsize\textcolor{black!65}{0.919--0.919}}} \\

& NMI & \makecell[c]{0.542\\{\scriptsize\textcolor{black!65}{0.533--0.552}}}
      & \makecell[c]{0.546\\{\scriptsize\textcolor{black!65}{0.542--0.549}}}
      & \makecell[c]{0.542\\{\scriptsize\textcolor{black!65}{0.540--0.544}}}
      & \makecell[c]{0.542\\{\scriptsize\textcolor{black!65}{0.531--0.552}}}
      & \makecell[c]{\textbf{0.594}\\{\scriptsize\textcolor{black!65}{0.594--0.594}}} \\

& ARI & \makecell[c]{0.662\\{\scriptsize\textcolor{black!65}{0.654--0.669}}}
      & \makecell[c]{0.666\\{\scriptsize\textcolor{black!65}{0.664--0.669}}}
      & \makecell[c]{0.664\\{\scriptsize\textcolor{black!65}{0.661--0.666}}}
      & \makecell[c]{0.660\\{\scriptsize\textcolor{black!65}{0.652--0.669}}}
      & \makecell[c]{\textbf{0.700}\\{\scriptsize\textcolor{black!65}{0.700--0.700}}} \\
      
& Runtime (s)
      & \makecell[c]{0.002}
      & \makecell[c]{0.004}
      & \makecell[c]{0.022}
      & \makecell[c]{0.016}
      & \makecell[c]{0.021} \\
      
\cmidrule(lr){2-7}

\multirow{5}{*}{\textsc{Htr}}
& SIL & \makecell[c]{0.684\\{\scriptsize\textcolor{black!65}{0.667--0.701}}}
      & \makecell[c]{0.679\\{\scriptsize\textcolor{black!65}{0.666--0.691}}}
      & \makecell[c]{0.684\\{\scriptsize\textcolor{black!65}{0.667--0.701}}}
      & \makecell[c]{0.653\\{\scriptsize\textcolor{black!65}{0.653--0.654}}}
      & \makecell[c]{\textbf{0.749}\\{\scriptsize\textcolor{black!65}{0.749--0.750}}} \\

& ACC & \makecell[c]{0.943\\{\scriptsize\textcolor{black!65}{0.933--0.952}}}
      & \makecell[c]{0.940\\{\scriptsize\textcolor{black!65}{0.933--0.947}}}
      & \makecell[c]{0.943\\{\scriptsize\textcolor{black!65}{0.934--0.952}}}
      & \makecell[c]{0.933\\{\scriptsize\textcolor{black!65}{0.931--0.934}}}
      & \makecell[c]{\textbf{0.965}\\{\scriptsize\textcolor{black!65}{0.965--0.965}}} \\

& NMI & \makecell[c]{0.442\\{\scriptsize\textcolor{black!65}{0.387--0.497}}}
      & \makecell[c]{0.423\\{\scriptsize\textcolor{black!65}{0.382--0.465}}}
      & \makecell[c]{0.442\\{\scriptsize\textcolor{black!65}{0.387--0.497}}}
      & \makecell[c]{0.391\\{\scriptsize\textcolor{black!65}{0.386--0.395}}}
      & \makecell[c]{\textbf{0.572}\\{\scriptsize\textcolor{black!65}{0.571--0.573}}} \\

& ARI & \makecell[c]{0.635\\{\scriptsize\textcolor{black!65}{0.593--0.677}}}
      & \makecell[c]{0.621\\{\scriptsize\textcolor{black!65}{0.589--0.652}}}
      & \makecell[c]{0.635\\{\scriptsize\textcolor{black!65}{0.593--0.676}}}
      & \makecell[c]{0.590\\{\scriptsize\textcolor{black!65}{0.585--0.595}}}
      & \makecell[c]{\textbf{0.722}\\{\scriptsize\textcolor{black!65}{0.721--0.723}}} \\
      
& Runtime (s)
      & \makecell[c]{0.006}
      & \makecell[c]{0.261}
      & \makecell[c]{6.251}
      & \makecell[c]{0.072}
      & \makecell[c]{0.151} \\
      
\cmidrule(lr){2-7}

\multirow{5}{*}{\textsc{Vcl}}
& SIL & \makecell[c]{0.587\\{\scriptsize\textcolor{black!65}{0.555--0.618}}}
      & \makecell[c]{0.589\\{\scriptsize\textcolor{black!65}{0.560--0.619}}}
      & \makecell[c]{0.588\\{\scriptsize\textcolor{black!65}{0.557--0.619}}}
      & \makecell[c]{0.575\\{\scriptsize\textcolor{black!65}{0.547--0.603}}}
      & \makecell[c]{\textbf{0.602}\\{\scriptsize\textcolor{black!65}{0.572--0.632}}} \\

& ACC & \makecell[c]{0.413\\{\scriptsize\textcolor{black!65}{0.397--0.428}}}
      & \makecell[c]{0.418\\{\scriptsize\textcolor{black!65}{0.402--0.434}}}
      & \makecell[c]{0.413\\{\scriptsize\textcolor{black!65}{0.398--0.429}}}
      & \makecell[c]{0.412\\{\scriptsize\textcolor{black!65}{0.399--0.425}}}
      & \makecell[c]{\textbf{0.428}\\{\scriptsize\textcolor{black!65}{0.414--0.442}}} \\

& NMI & \makecell[c]{0.163\\{\scriptsize\textcolor{black!65}{0.150--0.176}}}
      & \makecell[c]{0.168\\{\scriptsize\textcolor{black!65}{0.154--0.183}}}
      & \makecell[c]{0.163\\{\scriptsize\textcolor{black!65}{0.150--0.177}}}
      & \makecell[c]{0.168\\{\scriptsize\textcolor{black!65}{0.156--0.179}}}
      & \makecell[c]{\textbf{0.172}\\{\scriptsize\textcolor{black!65}{0.158--0.185}}} \\

& ARI & \makecell[c]{0.116\\{\scriptsize\textcolor{black!65}{0.109--0.124}}}
      & \makecell[c]{0.120\\{\scriptsize\textcolor{black!65}{0.112--0.128}}}
      & \makecell[c]{0.117\\{\scriptsize\textcolor{black!65}{0.109--0.124}}}
      & \makecell[c]{0.114\\{\scriptsize\textcolor{black!65}{0.105--0.123}}}
      & \makecell[c]{\textbf{0.125}\\{\scriptsize\textcolor{black!65}{0.118--0.132}}} \\
      
& Runtime (s)
      & \makecell[c]{0.002}
      & \makecell[c]{0.006}
      & \makecell[c]{0.058}
      & \makecell[c]{0.018}
      & \makecell[c]{0.024} \\
      
\cmidrule(lr){2-7}

\multirow{5}{*}{\textsc{Sms}}
& SIL & \makecell[c]{0.099\\{\scriptsize\textcolor{black!65}{0.099--0.099}}}
      & \makecell[c]{0.099\\{\scriptsize\textcolor{black!65}{0.099--0.099}}}
      & \makecell[c]{0.099\\{\scriptsize\textcolor{black!65}{0.099--0.099}}}
      & \makecell[c]{0.093\\{\scriptsize\textcolor{black!65}{0.093--0.093}}}
      & \makecell[c]{\textbf{0.111}\\{\scriptsize\textcolor{black!65}{0.111--0.111}}} \\

& ACC & \makecell[c]{0.845\\{\scriptsize\textcolor{black!65}{0.845--0.845}}}
      & \makecell[c]{0.845\\{\scriptsize\textcolor{black!65}{0.845--0.845}}}
      & \makecell[c]{0.845\\{\scriptsize\textcolor{black!65}{0.845--0.845}}}
      & \makecell[c]{0.834\\{\scriptsize\textcolor{black!65}{0.833--0.834}}}
      & \makecell[c]{\textbf{0.860}\\{\scriptsize\textcolor{black!65}{0.860--0.860}}} \\

& NMI & \makecell[c]{0.343\\{\scriptsize\textcolor{black!65}{0.343--0.343}}}
      & \makecell[c]{0.343\\{\scriptsize\textcolor{black!65}{0.343--0.343}}}
      & \makecell[c]{0.343\\{\scriptsize\textcolor{black!65}{0.343--0.343}}}
      & \makecell[c]{0.328\\{\scriptsize\textcolor{black!65}{0.327--0.329}}}
      & \makecell[c]{\textbf{0.364}\\{\scriptsize\textcolor{black!65}{0.364--0.364}}} \\

& ARI & \makecell[c]{0.414\\{\scriptsize\textcolor{black!65}{0.414--0.414}}}
      & \makecell[c]{0.414\\{\scriptsize\textcolor{black!65}{0.414--0.414}}}
      & \makecell[c]{0.414\\{\scriptsize\textcolor{black!65}{0.414--0.414}}}
      & \makecell[c]{0.387\\{\scriptsize\textcolor{black!65}{0.385--0.389}}}
      & \makecell[c]{\textbf{0.452}\\{\scriptsize\textcolor{black!65}{0.452--0.452}}} \\
      
& Runtime (s)
      & \makecell[c]{0.011}
      & \makecell[c]{0.105}
      & \makecell[c]{0.963}
      & \makecell[c]{0.079}
      & \makecell[c]{0.107} \\
      
\cmidrule(lr){2-7}

\multirow{5}{*}{\textsc{Mds}}
& SIL & \makecell[c]{0.338\\{\scriptsize\textcolor{black!65}{0.328--0.348}}}
      & \makecell[c]{0.344\\{\scriptsize\textcolor{black!65}{0.336--0.351}}}
      & \makecell[c]{0.347\\{\scriptsize\textcolor{black!65}{0.336--0.357}}}
      & \makecell[c]{0.326\\{\scriptsize\textcolor{black!65}{0.316--0.336}}}
      & \makecell[c]{\textbf{0.379}\\{\scriptsize\textcolor{black!65}{0.371--0.387}}} \\

& ACC & \makecell[c]{0.801\\{\scriptsize\textcolor{black!65}{0.775--0.826}}}
      & \makecell[c]{0.817\\{\scriptsize\textcolor{black!65}{0.797--0.837}}}
      & \makecell[c]{0.793\\{\scriptsize\textcolor{black!65}{0.763--0.822}}}
      & \makecell[c]{0.800\\{\scriptsize\textcolor{black!65}{0.772--0.827}}}
      & \makecell[c]{\textbf{0.845}\\{\scriptsize\textcolor{black!65}{0.815--0.874}}} \\

& NMI & \makecell[c]{0.876\\{\scriptsize\textcolor{black!65}{0.864--0.887}}}
      & \makecell[c]{0.886\\{\scriptsize\textcolor{black!65}{0.877--0.895}}}
      & \makecell[c]{0.878\\{\scriptsize\textcolor{black!65}{0.867--0.889}}}
      & \makecell[c]{0.877\\{\scriptsize\textcolor{black!65}{0.864--0.889}}}
      & \makecell[c]{\textbf{0.894}\\{\scriptsize\textcolor{black!65}{0.882--0.906}}} \\

& ARI & \makecell[c]{0.767\\{\scriptsize\textcolor{black!65}{0.741--0.792}}}
      & \makecell[c]{0.780\\{\scriptsize\textcolor{black!65}{0.758--0.801}}}
      & \makecell[c]{0.759\\{\scriptsize\textcolor{black!65}{0.733--0.785}}}
      & \makecell[c]{0.768\\{\scriptsize\textcolor{black!65}{0.741--0.796}}}
      & \makecell[c]{\textbf{0.803}\\{\scriptsize\textcolor{black!65}{0.775--0.832}}} \\
      
& Runtime (s)
      & \makecell[c]{0.003}
      & \makecell[c]{0.006}
      & \makecell[c]{0.088}
      & \makecell[c]{0.022}
      & \makecell[c]{0.046} \\
      
\cmidrule(lr){2-7}

\multirow{5}{*}{\textsc{Inr}}
& SIL & \makecell[c]{0.563\\{\scriptsize\textcolor{black!65}{0.543--0.584}}}
      & \makecell[c]{0.575\\{\scriptsize\textcolor{black!65}{0.556--0.594}}}
      & \makecell[c]{0.570\\{\scriptsize\textcolor{black!65}{0.554--0.586}}}
      & \makecell[c]{0.548\\{\scriptsize\textcolor{black!65}{0.534--0.563}}}
      & \makecell[c]{\textbf{0.606}\\{\scriptsize\textcolor{black!65}{0.606--0.606}}} \\

& ACC & \makecell[c]{0.601\\{\scriptsize\textcolor{black!65}{0.539--0.664}}}
      & \makecell[c]{0.632\\{\scriptsize\textcolor{black!65}{0.576--0.688}}}
      & \makecell[c]{0.600\\{\scriptsize\textcolor{black!65}{0.538--0.661}}}
      & \makecell[c]{0.638\\{\scriptsize\textcolor{black!65}{0.579--0.698}}}
      & \makecell[c]{\textbf{0.684}\\{\scriptsize\textcolor{black!65}{0.684--0.684}}} \\

& NMI & \makecell[c]{0.093\\{\scriptsize\textcolor{black!65}{0.088--0.099}}}
      & \makecell[c]{0.094\\{\scriptsize\textcolor{black!65}{0.090--0.099}}}
      & \makecell[c]{0.092\\{\scriptsize\textcolor{black!65}{0.087--0.097}}}
      & \makecell[c]{0.100\\{\scriptsize\textcolor{black!65}{0.095--0.104}}}
      & \makecell[c]{\textbf{0.104}\\{\scriptsize\textcolor{black!65}{0.104--0.104}}} \\

& ARI & \makecell[c]{0.051\\{\scriptsize\textcolor{black!65}{0.000--0.113}}}
      & \makecell[c]{0.080\\{\scriptsize\textcolor{black!65}{0.025--0.136}}}
      & \makecell[c]{0.048\\{\scriptsize\textcolor{black!65}{0.000--0.109}}}
      & \makecell[c]{0.089\\{\scriptsize\textcolor{black!65}{0.0029--0.149}}}
      & \makecell[c]{\textbf{0.133}\\{\scriptsize\textcolor{black!65}{0.133--0.133}}} \\
      
& Runtime (s)
      & \makecell[c]{0.002}
      & \makecell[c]{0.003}
      & \makecell[c]{0.017}
      & \makecell[c]{0.018}
      & \makecell[c]{0.018} \\
      
\cmidrule(lr){2-7}

\multirow{5}{*}{\textsc{Dbt}}
& SIL & \makecell[c]{0.303\\{\scriptsize\textcolor{black!65}{0.262--0.345}}}
      & \makecell[c]{0.306\\{\scriptsize\textcolor{black!65}{0.262--0.350}}}
      & \makecell[c]{0.302\\{\scriptsize\textcolor{black!65}{0.259--0.346}}}
      & \makecell[c]{0.269\\{\scriptsize\textcolor{black!65}{0.262--0.276}}}
      & \makecell[c]{\textbf{0.316}\\{\scriptsize\textcolor{black!65}{0.277--0.356}}} \\

& ACC & \makecell[c]{0.684\\{\scriptsize\textcolor{black!65}{0.666--0.702}}}
      & \makecell[c]{0.684\\{\scriptsize\textcolor{black!65}{0.666--0.702}}}
      & \makecell[c]{0.683\\{\scriptsize\textcolor{black!65}{0.666--0.701}}}
      & \makecell[c]{0.688\\{\scriptsize\textcolor{black!65}{0.671--0.704}}}
      & \makecell[c]{\textbf{0.699}\\{\scriptsize\textcolor{black!65}{0.683--0.714}}} \\

& NMI & \makecell[c]{0.082\\{\scriptsize\textcolor{black!65}{0.053--0.112}}}
      & \makecell[c]{0.083\\{\scriptsize\textcolor{black!65}{0.054--0.111}}}
      & \makecell[c]{0.087\\{\scriptsize\textcolor{black!65}{0.058--0.116}}}
      & \makecell[c]{0.091\\{\scriptsize\textcolor{black!65}{0.067--0.115}}}
      & \makecell[c]{\textbf{0.108}\\{\scriptsize\textcolor{black!65}{0.077--0.138}}} \\

& ARI & \makecell[c]{0.127\\{\scriptsize\textcolor{black!65}{0.089--0.165}}}
      & \makecell[c]{0.127\\{\scriptsize\textcolor{black!65}{0.090--0.164}}}
      & \makecell[c]{0.127\\{\scriptsize\textcolor{black!65}{0.090--0.163}}}
      & \makecell[c]{0.139\\{\scriptsize\textcolor{black!65}{0.112--0.165}}}
      & \makecell[c]{\textbf{0.150}\\{\scriptsize\textcolor{black!65}{0.113--0.188}}} \\
      
& Runtime (s)
      & \makecell[c]{0.002}
      & \makecell[c]{0.009}
      & \makecell[c]{0.111}
      & \makecell[c]{0.018}
      & \makecell[c]{0.043} \\
      
\cmidrule(lr){2-7}

\multirow{5}{*}{\textsc{Bbc}}
& SIL & \makecell[c]{0.184\\{\scriptsize\textcolor{black!65}{0.180--0.187}}}
      & \makecell[c]{0.182\\{\scriptsize\textcolor{black!65}{0.177--0.186}}}
      & \makecell[c]{0.184\\{\scriptsize\textcolor{black!65}{0.180--0.188}}}
      & \makecell[c]{0.178\\{\scriptsize\textcolor{black!65}{0.174--0.182}}}
      & \makecell[c]{\textbf{0.188}\\{\scriptsize\textcolor{black!65}{0.185--0.191}}} \\

& ACC & \makecell[c]{0.886\\{\scriptsize\textcolor{black!65}{0.837--0.935}}}
      & \makecell[c]{0.872\\{\scriptsize\textcolor{black!65}{0.821--0.923}}}
      & \makecell[c]{0.877\\{\scriptsize\textcolor{black!65}{0.827--0.927}}}
      & \makecell[c]{0.876\\{\scriptsize\textcolor{black!65}{0.827--0.925}}}
      & \makecell[c]{\textbf{0.896}\\{\scriptsize\textcolor{black!65}{0.850--0.942}}} \\

& NMI & \makecell[c]{0.837\\{\scriptsize\textcolor{black!65}{0.811--0.862}}}
      & \makecell[c]{0.822\\{\scriptsize\textcolor{black!65}{0.790--0.854}}}
      & \makecell[c]{0.833\\{\scriptsize\textcolor{black!65}{0.807--0.859}}}
      & \makecell[c]{0.828\\{\scriptsize\textcolor{black!65}{0.799--0.856}}}
      & \makecell[c]{\textbf{0.839}\\{\scriptsize\textcolor{black!65}{0.816--0.863}}} \\

& ARI & \makecell[c]{0.830\\{\scriptsize\textcolor{black!65}{0.781--0.879}}}
      & \makecell[c]{0.810\\{\scriptsize\textcolor{black!65}{0.756--0.864}}}
      & \makecell[c]{0.822\\{\scriptsize\textcolor{black!65}{0.771--0.872}}}
      & \makecell[c]{0.815\\{\scriptsize\textcolor{black!65}{0.763--0.866}}}
      & \makecell[c]{\textbf{0.837}\\{\scriptsize\textcolor{black!65}{0.791--0.884}}} \\
      
& Runtime (s)
      & \makecell[c]{0.003}
      & \makecell[c]{0.020}
      & \makecell[c]{0.159}
      & \makecell[c]{0.025}
      & \makecell[c]{0.043} \\
      
\cmidrule(lr){2-7}

\multirow{5}{*}{\textsc{Lkm}}
& SIL & \makecell[c]{0.204\\{\scriptsize\textcolor{black!65}{0.163--0.246}}}
      & \makecell[c]{0.174\\{\scriptsize\textcolor{black!65}{0.142--0.206}}}
      & \makecell[c]{0.194\\{\scriptsize\textcolor{black!65}{0.147--0.240}}}
      & \makecell[c]{0.198\\{\scriptsize\textcolor{black!65}{0.158--0.238}}}
      & \makecell[c]{\textbf{0.226}\\{\scriptsize\textcolor{black!65}{0.150--0.302}}} \\

& ACC & \makecell[c]{0.672\\{\scriptsize\textcolor{black!65}{0.647--0.698}}}
      & \makecell[c]{0.672\\{\scriptsize\textcolor{black!65}{0.615--0.729}}}
      & \makecell[c]{0.674\\{\scriptsize\textcolor{black!65}{0.614--0.733}}}
      & \makecell[c]{0.639\\{\scriptsize\textcolor{black!65}{0.589--0.689}}}
      & \makecell[c]{\textbf{0.717}\\{\scriptsize\textcolor{black!65}{0.692--0.742}}} \\

& NMI & \makecell[c]{0.066\\{\scriptsize\textcolor{black!65}{0.034--0.097}}}
      & \makecell[c]{0.111\\{\scriptsize\textcolor{black!65}{0.060--0.162}}}
      & \makecell[c]{0.098\\{\scriptsize\textcolor{black!65}{0.046--0.151}}}
      & \makecell[c]{0.048\\{\scriptsize\textcolor{black!65}{0.018--0.079}}}
      & \makecell[c]{\textbf{0.123}\\{\scriptsize\textcolor{black!65}{0.095--0.152}}} \\

& ARI & \makecell[c]{0.097\\{\scriptsize\textcolor{black!65}{0.055--0.140}}}
      & \makecell[c]{0.123\\{\scriptsize\textcolor{black!65}{0.048--0.197}}}
      & \makecell[c]{0.121\\{\scriptsize\textcolor{black!65}{0.036--0.205}}}
      & \makecell[c]{0.064\\{\scriptsize\textcolor{black!65}{0.010--0.118}}}
      & \makecell[c]{\textbf{0.171}\\{\scriptsize\textcolor{black!65}{0.121--0.222}}} \\
      
& Runtime (s)
      & \makecell[c]{0.010}
      & \makecell[c]{0.019}
      & \makecell[c]{0.081}
      & \makecell[c]{0.067}
      & \makecell[c]{0.096} \\
      
\cmidrule(lr){2-7}

\multirow{5}{*}{\textsc{Re8}}
& SIL & \makecell[c]{0.242\\{\scriptsize\textcolor{black!65}{0.234--0.250}}}
      & \makecell[c]{0.244\\{\scriptsize\textcolor{black!65}{0.237--0.251}}}
      & \makecell[c]{0.242\\{\scriptsize\textcolor{black!65}{0.235--0.250}}}
      & \makecell[c]{0.224\\{\scriptsize\textcolor{black!65}{0.215--0.233}}}
      & \makecell[c]{\textbf{0.252}\\{\scriptsize\textcolor{black!65}{0.243--0.261}}} \\

& ACC & \makecell[c]{0.491\\{\scriptsize\textcolor{black!65}{0.441--0.540}}}
      & \makecell[c]{0.491\\{\scriptsize\textcolor{black!65}{0.466--0.516}}}
      & \makecell[c]{0.482\\{\scriptsize\textcolor{black!65}{0.448--0.516}}}
      & \makecell[c]{0.460\\{\scriptsize\textcolor{black!65}{0.419--0.501}}}
      & \makecell[c]{\textbf{0.504}\\{\scriptsize\textcolor{black!65}{0.458--0.550}}} \\

& NMI & \makecell[c]{0.519\\{\scriptsize\textcolor{black!65}{0.498--0.539}}}
      & \makecell[c]{0.522\\{\scriptsize\textcolor{black!65}{0.503--0.542}}}
      & \makecell[c]{0.520\\{\scriptsize\textcolor{black!65}{0.499--0.540}}}
      & \makecell[c]{0.509\\{\scriptsize\textcolor{black!65}{0.483--0.535}}}
      & \makecell[c]{\textbf{0.527}\\{\scriptsize\textcolor{black!65}{0.506--0.548}}} \\

& ARI & \makecell[c]{0.349\\{\scriptsize\textcolor{black!65}{0.313--0.385}}}
      & \makecell[c]{0.352\\{\scriptsize\textcolor{black!65}{0.331--0.373}}}
      & \makecell[c]{0.344\\{\scriptsize\textcolor{black!65}{0.315--0.373}}}
      & \makecell[c]{0.320\\{\scriptsize\textcolor{black!65}{0.283--0.358}}}
      & \makecell[c]{\textbf{0.359}\\{\scriptsize\textcolor{black!65}{0.326--0.392}}} \\
      
& Runtime (s)
      & \makecell[c]{0.008}
      & \makecell[c]{0.049}
      & \makecell[c]{0.845}
      & \makecell[c]{0.042}
      & \makecell[c]{0.143} \\
      
\cmidrule(lr){2-7}

\multirow{5}{*}{\textsc{B77}}
& SIL & \makecell[c]{0.304\\{\scriptsize\textcolor{black!65}{0.299--0.309}}}
      & \makecell[c]{0.300\\{\scriptsize\textcolor{black!65}{0.297--0.303}}}
      & \makecell[c]{0.304\\{\scriptsize\textcolor{black!65}{0.298--0.309}}}
      & \makecell[c]{0.288\\{\scriptsize\textcolor{black!65}{0.282--0.294}}}
      & \makecell[c]{\textbf{0.314}\\{\scriptsize\textcolor{black!65}{0.309--0.319}}} \\

& ACC & \makecell[c]{0.546\\{\scriptsize\textcolor{black!65}{0.535--0.556}}}
      & \makecell[c]{0.542\\{\scriptsize\textcolor{black!65}{0.534--0.550}}}
      & \makecell[c]{0.545\\{\scriptsize\textcolor{black!65}{0.534--0.557}}}
      & \makecell[c]{0.538\\{\scriptsize\textcolor{black!65}{0.525--0.551}}}
      & \makecell[c]{\textbf{0.553}\\{\scriptsize\textcolor{black!65}{0.542--0.563}}} \\

& NMI & \makecell[c]{0.734\\{\scriptsize\textcolor{black!65}{0.730--0.739}}}
      & \makecell[c]{0.733\\{\scriptsize\textcolor{black!65}{0.731--0.735}}}
      & \makecell[c]{\textbf{0.735}\\{\scriptsize\textcolor{black!65}{0.731--0.739}}}
      & \makecell[c]{0.732\\{\scriptsize\textcolor{black!65}{0.728--0.736}}}
      & \makecell[c]{\textbf{0.735}\\{\scriptsize\textcolor{black!65}{0.731--0.739}}} \\

& ARI & \makecell[c]{0.444\\{\scriptsize\textcolor{black!65}{0.435--0.452}}}
      & \makecell[c]{0.438\\{\scriptsize\textcolor{black!65}{0.430--0.446}}}
      & \makecell[c]{\textbf{0.446}\\{\scriptsize\textcolor{black!65}{0.436--0.455}}}
      & \makecell[c]{0.440\\{\scriptsize\textcolor{black!65}{0.431--0.449}}}
      & \makecell[c]{\textbf{0.446}\\{\scriptsize\textcolor{black!65}{0.438--0.454}}} \\
      
& Runtime (s)
      & \makecell[c]{0.057}
      & \makecell[c]{0.220}
      & \makecell[c]{3.190}
      & \makecell[c]{0.656}
      & \makecell[c]{2.248} \\
      
\cmidrule(lr){2-7}

\multirow{5}{*}{\textsc{Clc}}
& SIL & \makecell[c]{0.286\\{\scriptsize\textcolor{black!65}{0.284--0.287}}}
      & \makecell[c]{0.288\\{\scriptsize\textcolor{black!65}{0.286--0.289}}}
      & \makecell[c]{0.287\\{\scriptsize\textcolor{black!65}{0.286--0.289}}}
      & \makecell[c]{0.274\\{\scriptsize\textcolor{black!65}{0.273--0.276}}}
      & \makecell[c]{\textbf{0.294}\\{\scriptsize\textcolor{black!65}{0.293--0.295}}} \\

& ACC & \makecell[c]{0.687\\{\scriptsize\textcolor{black!65}{0.682--0.693}}}
      & \makecell[c]{0.687\\{\scriptsize\textcolor{black!65}{0.680--0.694}}}
      & \makecell[c]{0.684\\{\scriptsize\textcolor{black!65}{0.679--0.689}}}
      & \makecell[c]{0.677\\{\scriptsize\textcolor{black!65}{0.671--0.682}}}
      & \makecell[c]{\textbf{0.693}\\{\scriptsize\textcolor{black!65}{0.687--0.698}}} \\

& NMI & \makecell[c]{0.858\\{\scriptsize\textcolor{black!65}{0.857--0.860}}}
      & \makecell[c]{0.858\\{\scriptsize\textcolor{black!65}{0.857--0.859}}}
      & \makecell[c]{\textbf{0.859}\\{\scriptsize\textcolor{black!65}{0.858--0.860}}}
      & \makecell[c]{0.855\\{\scriptsize\textcolor{black!65}{0.853--0.856}}}
      & \makecell[c]{\textbf{0.859}\\{\scriptsize\textcolor{black!65}{0.858--0.860}}} \\

& ARI & \makecell[c]{0.548\\{\scriptsize\textcolor{black!65}{0.545--0.552}}}
      & \makecell[c]{0.548\\{\scriptsize\textcolor{black!65}{0.544--0.552}}}
      & \makecell[c]{0.548\\{\scriptsize\textcolor{black!65}{0.545--0.551}}}
      & \makecell[c]{0.540\\{\scriptsize\textcolor{black!65}{0.536--0.543}}}
      & \makecell[c]{\textbf{0.552}\\{\scriptsize\textcolor{black!65}{0.549--0.555}}} \\
      
& Runtime (s)
      & \makecell[c]{0.248}
      & \makecell[c]{1.019}
      & \makecell[c]{6.209}
      & \makecell[c]{1.282}
      & \makecell[c]{4.247} \\
      
\cmidrule(lr){2-7}

\multirow{5}{*}{\textsc{Stl}}
& SIL & \makecell[c]{0.257\\{\scriptsize\textcolor{black!65}{0.245--0.268}}}
      & \makecell[c]{0.264\\{\scriptsize\textcolor{black!65}{0.255--0.273}}}
      & \makecell[c]{0.258\\{\scriptsize\textcolor{black!65}{0.247--0.270}}}
      & \makecell[c]{0.250\\{\scriptsize\textcolor{black!65}{0.238--0.262}}}
      & \makecell[c]{\textbf{0.265}\\{\scriptsize\textcolor{black!65}{0.254--0.276}}} \\

& ACC & \makecell[c]{0.843\\{\scriptsize\textcolor{black!65}{0.763--0.922}}}
      & \makecell[c]{0.858\\{\scriptsize\textcolor{black!65}{0.794--0.921}}}
      & \makecell[c]{0.840\\{\scriptsize\textcolor{black!65}{0.760--0.921}}}
      & \makecell[c]{0.832\\{\scriptsize\textcolor{black!65}{0.754--0.909}}}
      & \makecell[c]{\textbf{0.859}\\{\scriptsize\textcolor{black!65}{0.789--0.930}}} \\

& NMI & \makecell[c]{0.894\\{\scriptsize\textcolor{black!65}{0.863--0.926}}}
      & \makecell[c]{0.893\\{\scriptsize\textcolor{black!65}{0.869--0.918}}}
      & \makecell[c]{0.894\\{\scriptsize\textcolor{black!65}{0.862--0.925}}}
      & \makecell[c]{0.888\\{\scriptsize\textcolor{black!65}{0.858--0.917}}}
      & \makecell[c]{\textbf{0.899}\\{\scriptsize\textcolor{black!65}{0.872--0.925}}} \\

& ARI & \makecell[c]{0.820\\{\scriptsize\textcolor{black!65}{0.741--0.900}}}
      & \makecell[c]{0.829\\{\scriptsize\textcolor{black!65}{0.771--0.887}}}
      & \makecell[c]{0.819\\{\scriptsize\textcolor{black!65}{0.739--0.899}}}
      & \makecell[c]{0.806\\{\scriptsize\textcolor{black!65}{0.730--0.882}}}
      & \makecell[c]{\textbf{0.836}\\{\scriptsize\textcolor{black!65}{0.767--0.904}}} \\
      
& Runtime (s)
      & \makecell[c]{0.083}
      & \makecell[c]{0.493}
      & \makecell[c]{2.335}
      & \makecell[c]{0.613}
      & \makecell[c]{1.244} \\

\end{longtable}
\endgroup

\begin{figure}[h]
  \centering
  \includegraphics[width=\linewidth]{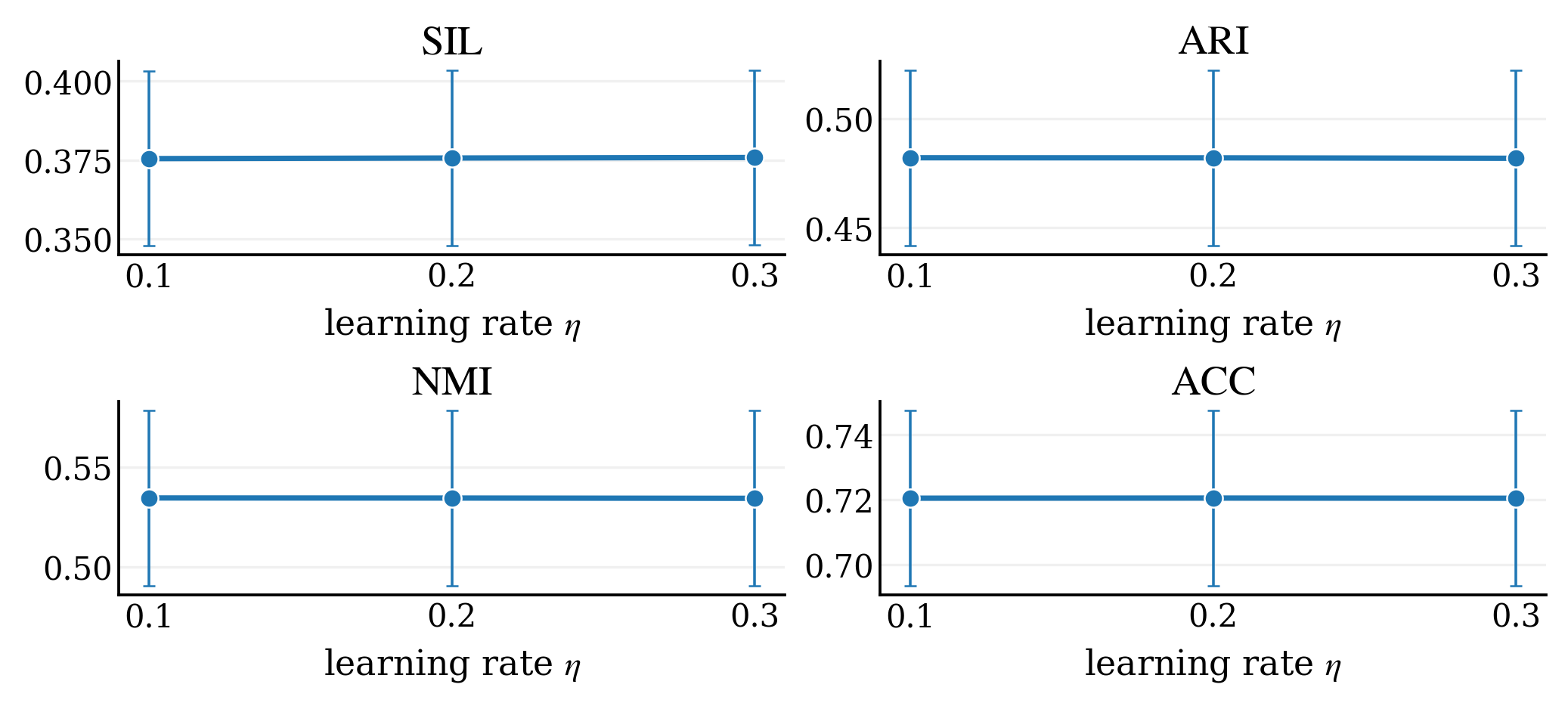}
  \caption{\textbf{Learning-rate ablation.} We examine sensitivity of \textsc{K-Sil} to the learning rate $\eta$ used in the multiplicative update of the temperature parameter that converts silhouette signals into sample weights (Eq.~\ref{eq:tauraw},\ref{eq:Psi_map}). For each dataset, we fix $k$ to the ground-truth number of classes $k^\star$ (\S\ref{subsec:datasets}) and run \textsc{K-Sil} with random initialization for each tested $\eta \in \{0.1, 0.2, 0.3\}$ over multiple random seeds; we report the mean performance with error bars summarizing variability across runs and datasets. Across all four criteria (SIL, ARI, NMI, and clustering accuracy ACC), performance is essentially flat within the error bars over the tested range, indicating that given the temperature bounds presented in \S\ref{subsec:temperature}, the adaptive-temperature mechanism is stable and does not require careful tuning of $\eta$ (we therefore use a single default value $\eta=0.2$ in the main experiments in \S\ref{sec:empirical}).}
  \label{fig:ablationlr}
\end{figure}

\begin{table}[h]
\centering
\caption{Exact vs.\ centroid-based approximate silhouette scores computed on a reference $k$-means run for each dataset. ``Exact'' denotes \texttt{sklearn.metrics.silhouette\_samples} (pairwise point distances), while ``Approx'' denotes our centroid-margin proxy. Pearson correlation and mean absolute error (MAE) are computed between per-point exact and approximate silhouette values. Reported times are seconds for computing per-point silhouette scores; Speedup is the ratio Time$_{\text{Exact}}$/Time$_{\text{Approx}}$.}
\label{tab:sil_exact_vs_approx}
\normalsize
\begin{tabular}{lrrrrr}
\toprule
Dataset & Pearson & MAE & Time Exact (s) & Time Approx (s) & Speedup \\
\midrule
\textsc{Lkm} & 0.738 & 0.088 & 0.0110 & 0.0022 & 4.8x \\
\textsc{Inr} & 0.951 & 0.103 & 0.0090 & 0.0006 & 15.0x \\
\textsc{Dbt} & 0.938 & 0.118 & 0.0140 & 0.0006 & 23.2x \\
\textsc{Mip} & 0.924 & 0.105 & 0.0240 & 0.0009 & 27.7x \\
\textsc{Wne} & 0.925 & 0.134 & 0.0020 & 0.0004 & 3.9x \\
\textsc{Brc} & 0.850 & 0.136 & 0.0110 & 0.0007 & 17.2x \\
\textsc{Vcl} & 0.926 & 0.098 & 0.0190 & 0.0007 & 27.6x \\
\textsc{Htr} & 0.894 & 0.088 & 2.6680 & 0.0022 & 1219.6x \\
\textsc{Sms} & 0.849 & 0.056 & 0.3380 & 0.0019 & 177.0x \\
\textsc{Mds} & 0.964 & 0.124 & 0.0070 & 0.0007 & 9.9x \\
\textsc{Bbc} & 0.935 & 0.077 & 0.0550 & 0.0011 & 48.7x \\
\textsc{Re8} & 0.947 & 0.094 & 0.1980 & 0.0014 & 138.5x \\
\textsc{B77} & 0.928 & 0.116 & 1.2460 & 0.0136 & 91.5x \\
\textsc{Clc} & 0.925 & 0.115 & 4.3530 & 0.0279 & 156.0x \\
\textsc{Stl} & 0.946 & 0.103 & 1.1570 & 0.0074 & 156.3x \\
\bottomrule
\end{tabular}
\end{table}

\begin{figure}[h]
  \centering
  \includegraphics[width=0.9\linewidth]{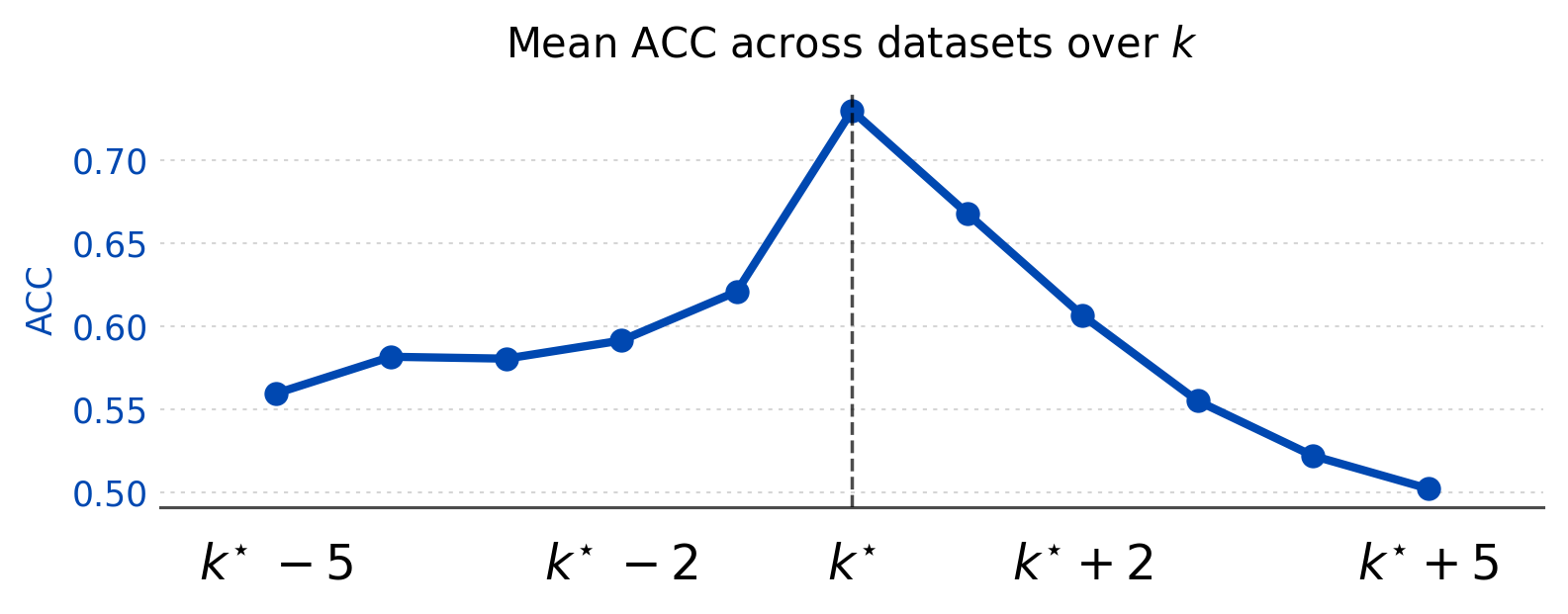}
  \caption{Mean \textcolor{blue}{ACC} across multiple random initializations and datasets (\S\ref{subsec:datasets}) as a function of the number of clusters $k\in\{k^\star-5, \ k^\star-4, \dots, \ k^\star, \dots, k^\star+4, \ k^\star+5 \}$.}
  \label{fig:accsub}
\end{figure}

\clearpage

\begingroup
\setlength{\tabcolsep}{2.5pt}          
\renewcommand{\arraystretch}{1.2}
\scriptsize
\setlength{\LTcapwidth}{\linewidth}  

\begin{longtable}{@{} l l *{5}{>{\centering\arraybackslash}p{1.95cm}} @{}}
\caption[Clustering results]{Mean Adjusted Mutual Information (AMI) and Davies--Bouldin index (DB) values with 95\% $t$-confidence intervals per dataset over runs. Gray ranges (\textcolor{black!65}{\(\cdot\)--\(\cdot\)}) show confidence intervals; best mean values are in \textbf{bold} (max for AMI, min for DB).}
\label{app:tab:all_metrics_ami_db}\\
\toprule
Dataset & Metric
& \makecell[c]{$k$-means}
& \makecell[c]{LOF\\$k$-means}
& \makecell[c]{iLOF\\$k$-means}
& \makecell[c]{OW\\$k$-means}
& \makecell[c]{\textsc{K-Sil}} \\
\midrule
\endfirsthead

\toprule
Dataset & Metric
& \makecell[c]{$k$-means}
& \makecell[c]{LOF\\$k$-means}
& \makecell[c]{iLOF\\$k$-means}
& \makecell[c]{OW\\$k$-means}
& \makecell[c]{\textsc{K-Sil}} \\
\midrule
\endhead

\bottomrule
\endlastfoot

\multirow{2}{*}{\textsc{Mip}}
& AMI & \makecell[c]{0.414\\{\scriptsize\textcolor{black!65}{0.401--0.426}}}
      & \makecell[c]{0.412\\{\scriptsize\textcolor{black!65}{0.399--0.426}}}
      & \makecell[c]{0.422\\{\scriptsize\textcolor{black!65}{0.412--0.433}}}
      & \makecell[c]{0.411\\{\scriptsize\textcolor{black!65}{0.398--0.425}}}
      & \makecell[c]{\textbf{0.437}\\{\scriptsize\textcolor{black!65}{0.427--0.447}}} \\

& DB  & \makecell[c]{0.801\\{\scriptsize\textcolor{black!65}{0.728--0.873}}}
      & \makecell[c]{0.819\\{\scriptsize\textcolor{black!65}{0.739--0.900}}}
      & \makecell[c]{0.839\\{\scriptsize\textcolor{black!65}{0.758--0.920}}}
      & \makecell[c]{0.773\\{\scriptsize\textcolor{black!65}{0.712--0.833}}}
      & \makecell[c]{\textbf{0.740}\\{\scriptsize\textcolor{black!65}{0.674--0.805}}} \\

\cmidrule(lr){2-7}

\multirow{2}{*}{\textsc{Wne}}
& AMI & \makecell[c]{0.837\\{\scriptsize\textcolor{black!65}{0.789--0.885}}}
      & \makecell[c]{0.853\\{\scriptsize\textcolor{black!65}{0.807--0.899}}}
      & \makecell[c]{0.841\\{\scriptsize\textcolor{black!65}{0.793--0.889}}}
      & \makecell[c]{0.833\\{\scriptsize\textcolor{black!65}{0.798--0.868}}}
      & \makecell[c]{\textbf{0.892}\\{\scriptsize\textcolor{black!65}{0.890--0.893}}} \\

& DB  & \makecell[c]{1.417\\{\scriptsize\textcolor{black!65}{1.345--1.490}}}
      & \makecell[c]{1.448\\{\scriptsize\textcolor{black!65}{1.365--1.531}}}
      & \makecell[c]{1.414\\{\scriptsize\textcolor{black!65}{1.345--1.484}}}
      & \makecell[c]{1.422\\{\scriptsize\textcolor{black!65}{1.354--1.489}}}
      & \makecell[c]{\textbf{1.389}\\{\scriptsize\textcolor{black!65}{1.389--1.390}}} \\

\cmidrule(lr){2-7}

\multirow{2}{*}{\textsc{BrC}}
& AMI & \makecell[c]{0.542\\{\scriptsize\textcolor{black!65}{0.532--0.551}}}
      & \makecell[c]{0.545\\{\scriptsize\textcolor{black!65}{0.542--0.548}}}
      & \makecell[c]{0.541\\{\scriptsize\textcolor{black!65}{0.539--0.543}}}
      & \makecell[c]{0.541\\{\scriptsize\textcolor{black!65}{0.531--0.552}}}
      & \makecell[c]{\textbf{0.593}\\{\scriptsize\textcolor{black!65}{0.593--0.593}}} \\

& DB  & \makecell[c]{1.317\\{\scriptsize\textcolor{black!65}{1.314--1.321}}}
      & \makecell[c]{1.320\\{\scriptsize\textcolor{black!65}{1.317--1.322}}}
      & \makecell[c]{1.321\\{\scriptsize\textcolor{black!65}{1.321--1.322}}}
      & \makecell[c]{1.316\\{\scriptsize\textcolor{black!65}{1.313--1.320}}}
      & \makecell[c]{\textbf{1.304}\\{\scriptsize\textcolor{black!65}{1.304--1.304}}} \\

\cmidrule(lr){2-7}

\multirow{2}{*}{\textsc{Htr}}
& AMI & \makecell[c]{0.442\\{\scriptsize\textcolor{black!65}{0.387--0.497}}}
      & \makecell[c]{0.423\\{\scriptsize\textcolor{black!65}{0.382--0.465}}}
      & \makecell[c]{0.442\\{\scriptsize\textcolor{black!65}{0.387--0.497}}}
      & \makecell[c]{0.391\\{\scriptsize\textcolor{black!65}{0.386--0.395}}}
      & \makecell[c]{\textbf{0.572}\\{\scriptsize\textcolor{black!65}{0.571--0.573}}} \\

& DB  & \makecell[c]{0.992\\{\scriptsize\textcolor{black!65}{0.896--1.088}}}
      & \makecell[c]{1.024\\{\scriptsize\textcolor{black!65}{0.951--1.096}}}
      & \makecell[c]{0.991\\{\scriptsize\textcolor{black!65}{0.895--1.088}}}
      & \makecell[c]{1.072\\{\scriptsize\textcolor{black!65}{1.067--1.077}}}
      & \makecell[c]{\textbf{0.690}\\{\scriptsize\textcolor{black!65}{0.689--0.691}}} \\

\cmidrule(lr){2-7}

\multirow{2}{*}{\textsc{Vcl}}
& AMI & \makecell[c]{0.160\\{\scriptsize\textcolor{black!65}{0.146--0.173}}}
      & \makecell[c]{0.165\\{\scriptsize\textcolor{black!65}{0.150--0.179}}}
      & \makecell[c]{0.160\\{\scriptsize\textcolor{black!65}{0.146--0.174}}}
      & \makecell[c]{0.164\\{\scriptsize\textcolor{black!65}{0.153--0.176}}}
      & \makecell[c]{\textbf{0.169}\\{\scriptsize\textcolor{black!65}{0.155--0.182}}} \\

& DB  & \makecell[c]{0.856\\{\scriptsize\textcolor{black!65}{0.769--0.944}}}
      & \makecell[c]{0.848\\{\scriptsize\textcolor{black!65}{0.768--0.927}}}
      & \makecell[c]{0.856\\{\scriptsize\textcolor{black!65}{0.769--0.944}}}
      & \makecell[c]{0.854\\{\scriptsize\textcolor{black!65}{0.778--0.929}}}
      & \makecell[c]{\textbf{0.819}\\{\scriptsize\textcolor{black!65}{0.757--0.880}}} \\

\cmidrule(lr){2-7}

\multirow{2}{*}{\textsc{Sms}}
& AMI & \makecell[c]{0.343\\{\scriptsize\textcolor{black!65}{0.343--0.343}}}
      & \makecell[c]{0.343\\{\scriptsize\textcolor{black!65}{0.343--0.343}}}
      & \makecell[c]{0.343\\{\scriptsize\textcolor{black!65}{0.343--0.343}}}
      & \makecell[c]{0.328\\{\scriptsize\textcolor{black!65}{0.327--0.329}}}
      & \makecell[c]{\textbf{0.364}\\{\scriptsize\textcolor{black!65}{0.364--0.364}}} \\

& DB  & \makecell[c]{4.068\\{\scriptsize\textcolor{black!65}{4.068--4.069}}}
      & \makecell[c]{4.068\\{\scriptsize\textcolor{black!65}{4.068--4.068}}}
      & \makecell[c]{4.068\\{\scriptsize\textcolor{black!65}{4.068--4.068}}}
      & \makecell[c]{4.130\\{\scriptsize\textcolor{black!65}{4.127--4.134}}}
      & \makecell[c]{\textbf{3.984}\\{\scriptsize\textcolor{black!65}{3.983--3.984}}} \\

\cmidrule(lr){2-7}

\multirow{2}{*}{\textsc{Mds}}
& AMI & \makecell[c]{0.867\\{\scriptsize\textcolor{black!65}{0.854--0.879}}}
      & \makecell[c]{0.878\\{\scriptsize\textcolor{black!65}{0.868--0.888}}}
      & \makecell[c]{0.869\\{\scriptsize\textcolor{black!65}{0.857--0.881}}}
      & \makecell[c]{0.868\\{\scriptsize\textcolor{black!65}{0.854--0.881}}}
      & \makecell[c]{\textbf{0.887}\\{\scriptsize\textcolor{black!65}{0.874--0.899}}} \\

& DB  & \makecell[c]{1.904\\{\scriptsize\textcolor{black!65}{1.853--1.955}}}
      & \makecell[c]{1.891\\{\scriptsize\textcolor{black!65}{1.852--1.930}}}
      & \makecell[c]{1.885\\{\scriptsize\textcolor{black!65}{1.834--1.935}}}
      & \makecell[c]{1.911\\{\scriptsize\textcolor{black!65}{1.858--1.963}}}
      & \makecell[c]{\textbf{1.785}\\{\scriptsize\textcolor{black!65}{1.749--1.821}}} \\

\cmidrule(lr){2-7}

\multirow{2}{*}{\textsc{Inr}}
& AMI & \makecell[c]{0.091\\{\scriptsize\textcolor{black!65}{0.085--0.097}}}
      & \makecell[c]{0.092\\{\scriptsize\textcolor{black!65}{0.088--0.097}}}
      & \makecell[c]{0.090\\{\scriptsize\textcolor{black!65}{0.085--0.095}}}
      & \makecell[c]{0.098\\{\scriptsize\textcolor{black!65}{0.092--0.103}}}
      & \makecell[c]{\textbf{0.102}\\{\scriptsize\textcolor{black!65}{0.102--0.102}}} \\

& DB  & \makecell[c]{0.893\\{\scriptsize\textcolor{black!65}{0.854--0.932}}}
      & \makecell[c]{0.871\\{\scriptsize\textcolor{black!65}{0.835--0.908}}}
      & \makecell[c]{0.892\\{\scriptsize\textcolor{black!65}{0.852--0.932}}}
      & \makecell[c]{0.874\\{\scriptsize\textcolor{black!65}{0.844--0.905}}}
      & \makecell[c]{\textbf{0.836}\\{\scriptsize\textcolor{black!65}{0.836--0.836}}} \\

\cmidrule(lr){2-7}

\multirow{2}{*}{\textsc{Dbt}}
& AMI & \makecell[c]{0.081\\{\scriptsize\textcolor{black!65}{0.052--0.111}}}
      & \makecell[c]{0.082\\{\scriptsize\textcolor{black!65}{0.053--0.110}}}
      & \makecell[c]{0.086\\{\scriptsize\textcolor{black!65}{0.057--0.115}}}
      & \makecell[c]{0.090\\{\scriptsize\textcolor{black!65}{0.066--0.114}}}
      & \makecell[c]{\textbf{0.107}\\{\scriptsize\textcolor{black!65}{0.076--0.137}}} \\

& DB  & \makecell[c]{2.002\\{\scriptsize\textcolor{black!65}{1.806--2.198}}}
      & \makecell[c]{\textbf{1.998}\\{\scriptsize\textcolor{black!65}{1.798--2.198}}}
      & \makecell[c]{2.009\\{\scriptsize\textcolor{black!65}{1.811--2.207}}}
      & \makecell[c]{2.090\\{\scriptsize\textcolor{black!65}{2.068--2.113}}}
      & \makecell[c]{2.005\\{\scriptsize\textcolor{black!65}{1.808--2.202}}} \\

\cmidrule(lr){2-7}

\multirow{2}{*}{\textsc{Bbc}}
& AMI & \makecell[c]{0.836\\{\scriptsize\textcolor{black!65}{0.811--0.862}}}
      & \makecell[c]{0.821\\{\scriptsize\textcolor{black!65}{0.789--0.853}}}
      & \makecell[c]{0.833\\{\scriptsize\textcolor{black!65}{0.806--0.859}}}
      & \makecell[c]{0.827\\{\scriptsize\textcolor{black!65}{0.799--0.856}}}
      & \makecell[c]{\textbf{0.839}\\{\scriptsize\textcolor{black!65}{0.816--0.862}}} \\

& DB  & \makecell[c]{2.644\\{\scriptsize\textcolor{black!65}{2.616--2.672}}}
      & \makecell[c]{2.688\\{\scriptsize\textcolor{black!65}{2.631--2.745}}}
      & \makecell[c]{2.651\\{\scriptsize\textcolor{black!65}{2.621--2.681}}}
      & \makecell[c]{2.655\\{\scriptsize\textcolor{black!65}{2.623--2.686}}}
      & \makecell[c]{\textbf{2.636}\\{\scriptsize\textcolor{black!65}{2.611--2.662}}} \\

\cmidrule(lr){2-7}

\multirow{2}{*}{\textsc{Lkm}}
& AMI & \makecell[c]{0.054\\{\scriptsize\textcolor{black!65}{0.020--0.087}}}
      & \makecell[c]{0.101\\{\scriptsize\textcolor{black!65}{0.048--0.153}}}
      & \makecell[c]{0.087\\{\scriptsize\textcolor{black!65}{0.032--0.142}}}
      & \makecell[c]{0.036\\{\scriptsize\textcolor{black!65}{0.004--0.068}}}
      & \makecell[c]{\textbf{0.112}\\{\scriptsize\textcolor{black!65}{0.081--0.144}}} \\

& DB  & \makecell[c]{2.717\\{\scriptsize\textcolor{black!65}{2.433--3.000}}}
      & \makecell[c]{2.911\\{\scriptsize\textcolor{black!65}{2.660--3.163}}}
      & \makecell[c]{2.716\\{\scriptsize\textcolor{black!65}{2.421--3.010}}}
      & \makecell[c]{2.713\\{\scriptsize\textcolor{black!65}{2.403--3.022}}}
      & \makecell[c]{\textbf{2.634}\\{\scriptsize\textcolor{black!65}{2.107--3.160}}} \\

\cmidrule(lr){2-7}

\multirow{2}{*}{\textsc{Re8}}
& AMI & \makecell[c]{0.517\\{\scriptsize\textcolor{black!65}{0.497--0.538}}}
      & \makecell[c]{0.521\\{\scriptsize\textcolor{black!65}{0.501--0.540}}}
      & \makecell[c]{0.518\\{\scriptsize\textcolor{black!65}{0.498--0.539}}}
      & \makecell[c]{0.507\\{\scriptsize\textcolor{black!65}{0.481--0.534}}}
      & \makecell[c]{\textbf{0.525}\\{\scriptsize\textcolor{black!65}{0.504--0.546}}} \\

& DB  & \makecell[c]{2.103\\{\scriptsize\textcolor{black!65}{2.036--2.170}}}
      & \makecell[c]{2.096\\{\scriptsize\textcolor{black!65}{2.036--2.156}}}
      & \makecell[c]{2.110\\{\scriptsize\textcolor{black!65}{2.039--2.182}}}
      & \makecell[c]{2.191\\{\scriptsize\textcolor{black!65}{2.119--2.263}}}
      & \makecell[c]{\textbf{2.048}\\{\scriptsize\textcolor{black!65}{1.982--2.113}}} \\

\cmidrule(lr){2-7}

\multirow{2}{*}{\textsc{B77}}
& AMI & \makecell[c]{0.718\\{\scriptsize\textcolor{black!65}{0.713--0.722}}}
      & \makecell[c]{0.716\\{\scriptsize\textcolor{black!65}{0.714--0.719}}}
      & \makecell[c]{\textbf{0.719}\\{\scriptsize\textcolor{black!65}{0.714--0.723}}}
      & \makecell[c]{0.716\\{\scriptsize\textcolor{black!65}{0.712--0.720}}}
      & \makecell[c]{\textbf{0.719}\\{\scriptsize\textcolor{black!65}{0.715--0.723}}} \\

& DB  & \makecell[c]{1.718\\{\scriptsize\textcolor{black!65}{1.688--1.747}}}
      & \makecell[c]{1.735\\{\scriptsize\textcolor{black!65}{1.720--1.749}}}
      & \makecell[c]{1.728\\{\scriptsize\textcolor{black!65}{1.697--1.759}}}
      & \makecell[c]{1.749\\{\scriptsize\textcolor{black!65}{1.719--1.779}}}
      & \makecell[c]{\textbf{1.692}\\{\scriptsize\textcolor{black!65}{1.667--1.716}}} \\

\cmidrule(lr){2-7}

\multirow{2}{*}{\textsc{Clc}}
& AMI & \makecell[c]{0.841\\{\scriptsize\textcolor{black!65}{0.840--0.843}}}
      & \makecell[c]{0.841\\{\scriptsize\textcolor{black!65}{0.839--0.842}}}
      & \makecell[c]{\textbf{0.842}\\{\scriptsize\textcolor{black!65}{0.841--0.844}}}
      & \makecell[c]{0.837\\{\scriptsize\textcolor{black!65}{0.836--0.839}}}
      & \makecell[c]{\textbf{0.842}\\{\scriptsize\textcolor{black!65}{0.841--0.844}}} \\

& DB  & \makecell[c]{2.027\\{\scriptsize\textcolor{black!65}{2.015--2.039}}}
      & \makecell[c]{2.012\\{\scriptsize\textcolor{black!65}{2.000--2.025}}}
      & \makecell[c]{2.033\\{\scriptsize\textcolor{black!65}{2.021--2.044}}}
      & \makecell[c]{2.055\\{\scriptsize\textcolor{black!65}{2.044--2.066}}}
      & \makecell[c]{\textbf{2.004}\\{\scriptsize\textcolor{black!65}{1.994--2.013}}} \\

\cmidrule(lr){2-7}

\multirow{2}{*}{\textsc{Stl}}
& AMI & \makecell[c]{0.894\\{\scriptsize\textcolor{black!65}{0.863--0.926}}}
      & \makecell[c]{0.893\\{\scriptsize\textcolor{black!65}{0.869--0.918}}}
      & \makecell[c]{0.893\\{\scriptsize\textcolor{black!65}{0.862--0.925}}}
      & \makecell[c]{0.887\\{\scriptsize\textcolor{black!65}{0.857--0.917}}}
      & \makecell[c]{\textbf{0.899}\\{\scriptsize\textcolor{black!65}{0.872--0.925}}} \\

& DB  & \makecell[c]{2.168\\{\scriptsize\textcolor{black!65}{2.075--2.261}}}
      & \makecell[c]{\textbf{2.118}\\{\scriptsize\textcolor{black!65}{2.042--2.193}}}
      & \makecell[c]{2.171\\{\scriptsize\textcolor{black!65}{2.078--2.265}}}
      & \makecell[c]{2.181\\{\scriptsize\textcolor{black!65}{2.078--2.285}}}
      & \makecell[c]{2.127\\{\scriptsize\textcolor{black!65}{2.039--2.214}}} \\

\end{longtable}
\endgroup

\clearpage

\begin{figure}[t]
    \centering

    \begin{subfigure}[t]{\linewidth}
        \centering
        \includegraphics[width=0.92\linewidth]{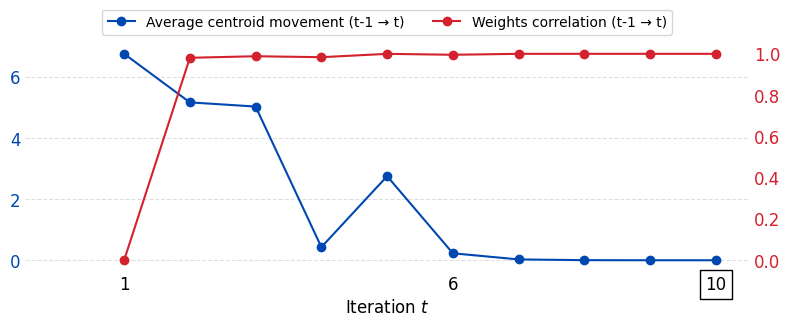}
        \caption{Leukemia (\textsc{Lkm})}
    \end{subfigure}\vspace{0.4em}

    \begin{subfigure}[t]{\linewidth}
        \centering
        \includegraphics[width=0.92\linewidth]{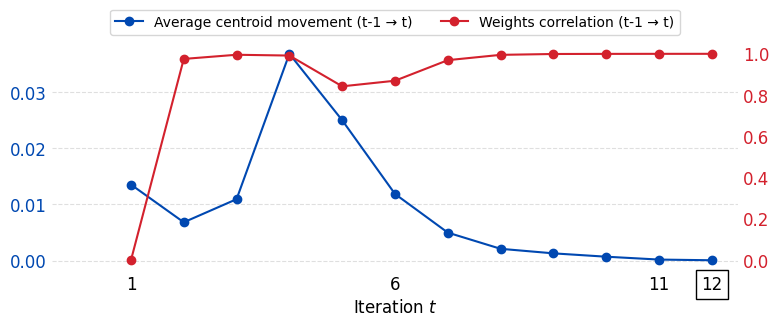}
        \caption{Ionosphere (\textsc{Inr})}
    \end{subfigure}\vspace{0.4em}

    \begin{subfigure}[t]{\linewidth}
        \centering
        \includegraphics[width=0.92\linewidth]{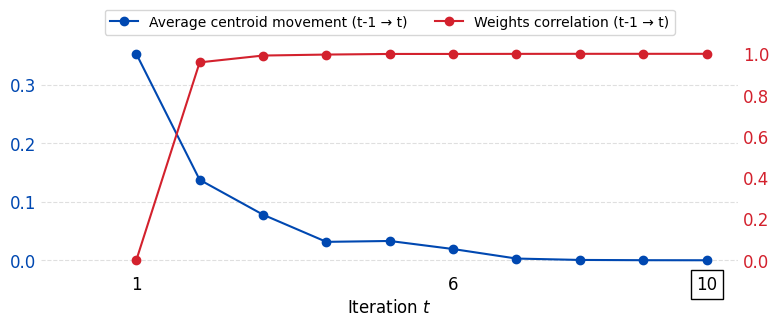}
        \caption{Diabetes (\textsc{Dbt})}
    \end{subfigure}

    \caption{\textcolor{blue}{Average centroid movement} and \textcolor{red}{weights correlation} across \textsc{K-Sil} iterations $t$ (\S\ref{subsec:eval}) for \textsc{Lkm}, \textsc{Inr}, and \textsc{Dbt}. The final iteration is marked by $\square$.}
    \label{fig:convergence_app_a}
\end{figure}

\begin{figure}[t]
    \centering

    \begin{subfigure}[t]{\linewidth}
        \centering
        \includegraphics[width=0.92\linewidth]{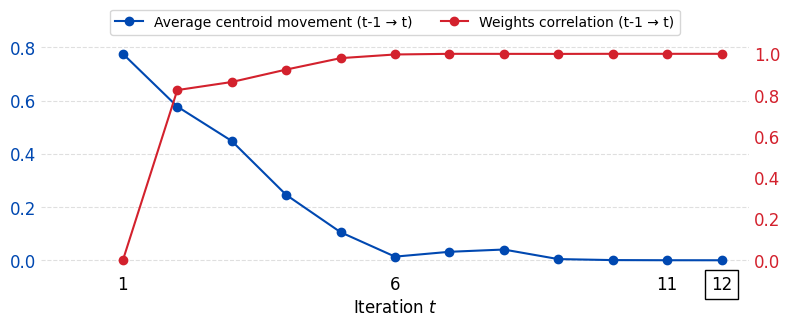}
        \caption{Wine (\textsc{Wne})}
    \end{subfigure}\vspace{0.4em}

    \begin{subfigure}[t]{\linewidth}
        \centering
        \includegraphics[width=0.92\linewidth]{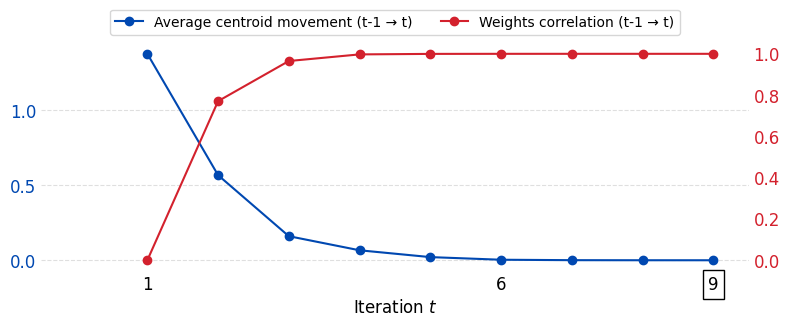}
        \caption{Breast Cancer (\textsc{BrC})}
    \end{subfigure}\vspace{0.4em}

    \begin{subfigure}[t]{\linewidth}
        \centering
        \includegraphics[width=0.92\linewidth]{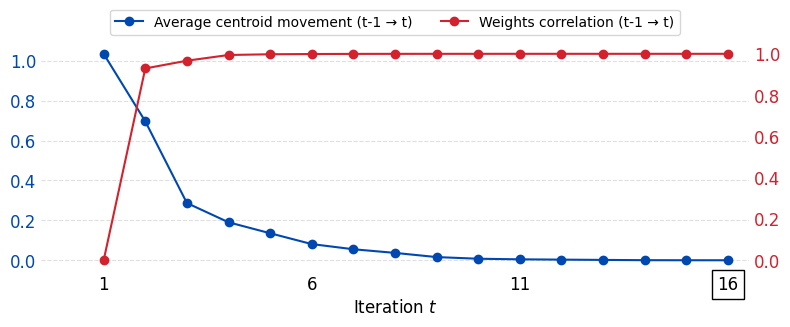}
        \caption{HTRU2 (\textsc{Htr})}
    \end{subfigure}

    \caption{\textcolor{blue}{Average centroid movement} and \textcolor{red}{weights correlation} across \textsc{K-Sil} iterations $t$ (\S\ref{subsec:eval}) for \textsc{Wne}, \textsc{Brc}, and \textsc{Htr}. The final iteration is marked by $\square$.}
    \label{fig:convergence_app_b}
\end{figure}

\begin{figure}[t]
    \centering

    \begin{subfigure}[t]{\linewidth}
        \centering
        \includegraphics[width=0.92\linewidth]{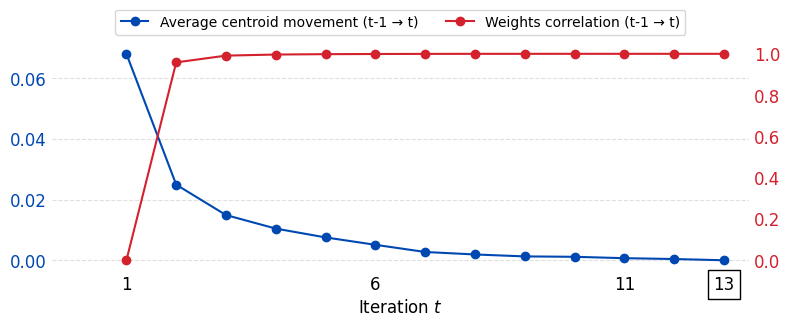}
        \caption{SMS Spam (\textsc{Sms})}
    \end{subfigure}\vspace{0.4em}

    \begin{subfigure}[t]{\linewidth}
        \centering
        \includegraphics[width=0.92\linewidth]{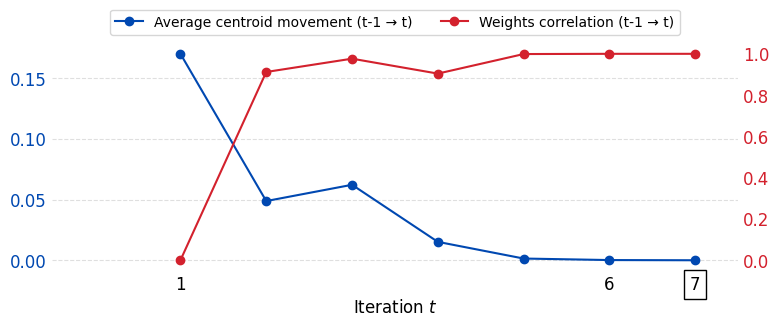}
        \caption{MINDS-14 (\textsc{Mds})}
    \end{subfigure}\vspace{0.4em}

    \begin{subfigure}[t]{\linewidth}
        \centering
        \includegraphics[width=0.92\linewidth]{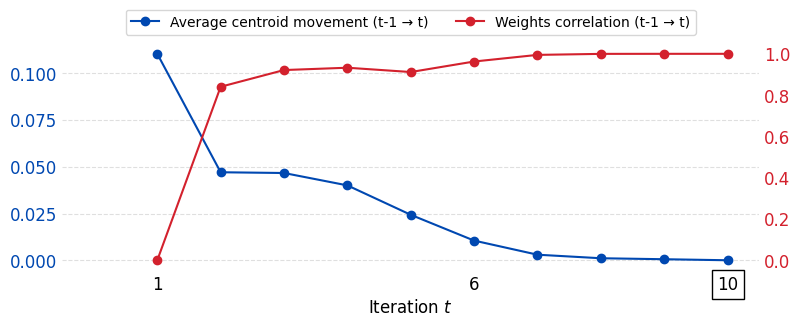}
        \caption{BBC News (\textsc{Bbc})}
    \end{subfigure}

    \caption{\textcolor{blue}{Average centroid movement} and \textcolor{red}{weights correlation} across \textsc{K-Sil} iterations $t$ (\S\ref{subsec:eval}) for \textsc{Sms}, \textsc{Mds}, and \textsc{Bbc}. The final iteration is marked by $\square$.}
    \label{fig:convergence_app_c}
\end{figure}

\begin{figure}[t]
    \centering

    \begin{subfigure}[t]{\linewidth}
        \centering
        \includegraphics[width=0.92\linewidth]{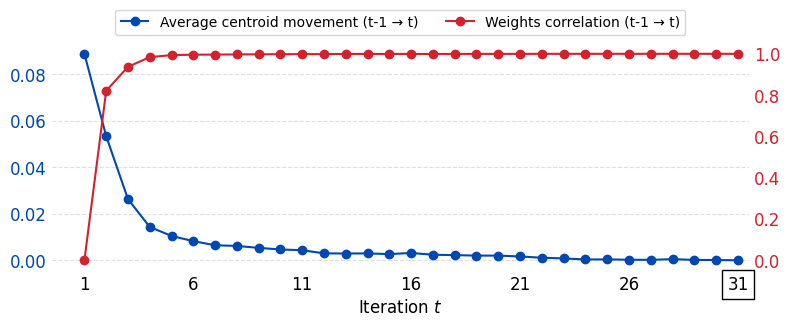}
        \caption{R8 (\textsc{Re8})}
    \end{subfigure}\vspace{0.4em}

    \begin{subfigure}[t]{\linewidth}
        \centering
        \includegraphics[width=0.92\linewidth]{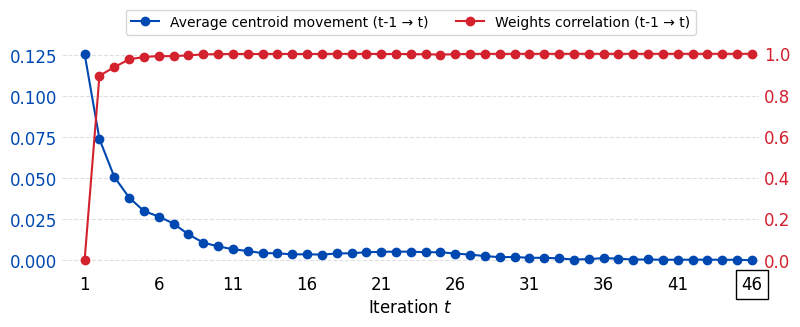}
        \caption{Banking77 (\textsc{B77})}
    \end{subfigure}\vspace{0.4em}

    \begin{subfigure}[t]{\linewidth}
        \centering
        \includegraphics[width=0.92\linewidth]{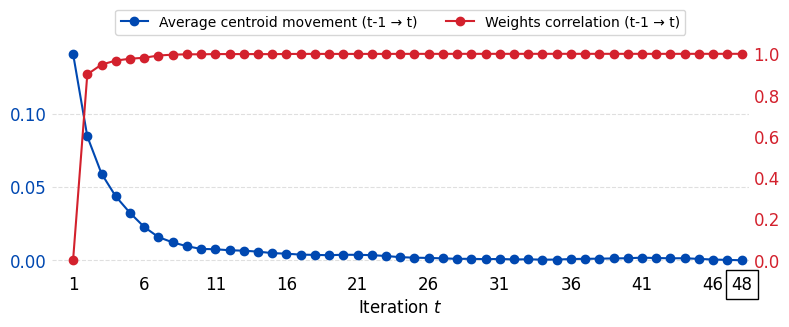}
        \caption{CLINC150 (\textsc{Clc})}
    \end{subfigure}

    \caption{\textcolor{blue}{Average centroid movement} and \textcolor{red}{weights correlation} across \textsc{K-Sil} iterations $t$ (\S\ref{subsec:eval}) for \textsc{Re8}, \textsc{B77}, and \textsc{Clc}. The final iteration is marked by $\square$.}
    \label{fig:convergence_app_d}
\end{figure}

\begin{figure}[t]
    \centering

    \begin{subfigure}[t]{\linewidth}
        \centering
        \includegraphics[width=0.92\linewidth]{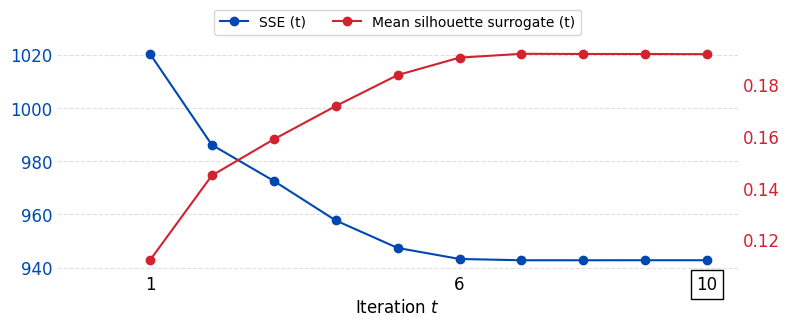}
        \caption{\textsc{Bbc}}
    \end{subfigure}\vspace{0.4em}

    \begin{subfigure}[t]{\linewidth}
        \centering
        \includegraphics[width=0.92\linewidth]{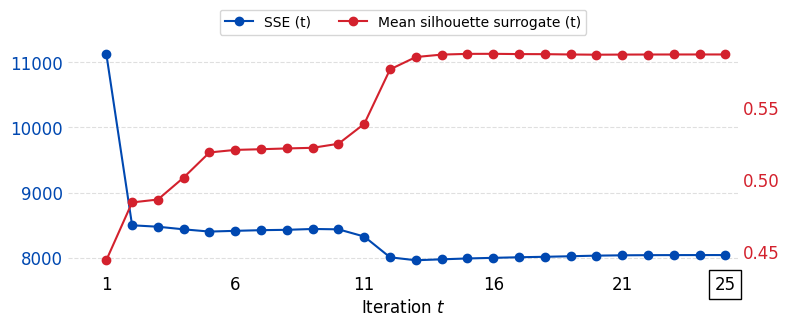}
        \caption{\textsc{Mip}}
    \end{subfigure}\vspace{0.4em}

    \begin{subfigure}[t]{\linewidth}
        \centering
        \includegraphics[width=0.92\linewidth]{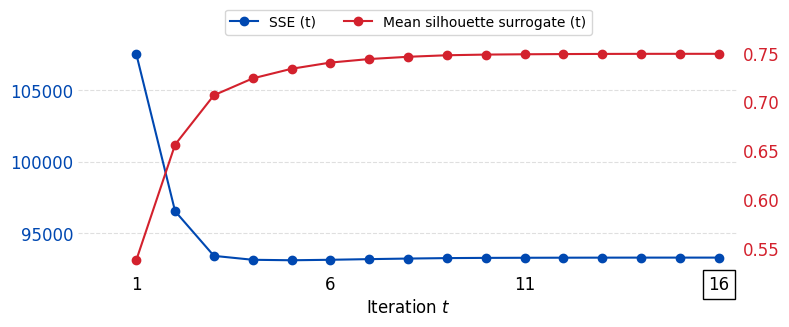}
        \caption{\textsc{Htr}}
    \end{subfigure}

    \caption{\textcolor{blue}{SSE} and \textcolor{red}{Silhouette score} across \textsc{K-Sil} iterations $t$ (\S\ref{subsec:eval}) for representative datasets: \textsc{Bbc}, \textsc{Mip}, and \textsc{Htr}. The final iteration is marked by $\square$.}
    \label{fig:sse_sil}
\end{figure}

\clearpage

\subsection{Initial temperature}\label{app:inittau}
To choose the warm-start temperature $\tau_0$ (used only at the beginning of \textsc{K-Sil}), we inspect the distribution of the initialization silhouette-surrogate scores $s_i(\mu_0)$ obtained right after the random initialization, before any reweighting or center updates. For each dataset we pool $s_i(\mu_0)$ over multiple random seeds and summarize it with: (i) the median and interquartile range (IQR), and (ii) the boundary mass $P[s_i(\mu_0)<0.10]$, i.e., the fraction of points that start in a low-confidence / ambiguous regime.
Intuitively, $\tau_0$ controls how sharply the initial reweighting separates points that look confidently assigned under the initial partition from points that look ambiguous. When $\tau_0$ is larger, the early weights react more strongly to differences in $s_i(\mu_0)$, so points with higher initialization confidence are emphasized more at the warm start; when $\tau_0$ is smaller, the warm start is milder and the method behaves closer to an unweighted start. Since this choice only affects the first stage, we restrict attention to a simple binary decision $\tau_0\in\{1,2\}$.
Heuristically, we set $\tau_0=2$ when the pooled initialization scores indicate that the initial partition contains either (i) a very large ambiguous mass, or (ii) a mixed confidence regime, or (iii) strongly heterogeneous confidence. Concretely, we choose $\tau_0=2$ if any of the following holds:
(i) $P[s_i(\mu_0)<0.10]\ge 0.90$; or
(ii) $\mathrm{median}(s_i(\mu_0))\in[0.15,0.25]$ and $P[s_i(\mu_0)<0.10]\ge 0.20$; or (iii) $\mathrm{IQR}(s_i(\mu_0))\ge 0.25$ and $\mathrm{median}(s_i(\mu_0))\le 0.50$. Otherwise we use the milder warm start $\tau_0=1$.
Table~\ref{tab:init_tau0_diag} reports these pooled initialization summaries and the resulting $\tau_0$ choice for each dataset.

\begin{table}[h]
\centering
\small
\setlength{\tabcolsep}{7pt}
\begin{tabular}{lcccc}
\hline
Dataset & Median & IQR & $P[s<0.10]$ & $\tau_0$ \\
\hline
Sms & 0.036 & 0.043 & 0.95 & 2 \\
Clc & 0.151 & 0.220 & 0.37 & 2 \\
Mds & 0.211 & 0.264 & 0.27 & 2 \\
Lkm & 0.182 & 0.161 & 0.23 & 2 \\
Dbt & 0.270 & 0.262 & 0.17 & 2 \\
Wne & 0.325 & 0.291 & 0.14 & 2 \\
BrC & 0.404 & 0.325 & 0.10 & 2 \\
MiP & 0.490 & 0.370 & 0.07 & 2 \\
Bbc & 0.084 & 0.107 & 0.57 & 1 \\
Stl & 0.131 & 0.160 & 0.40 & 1 \\
Re8 & 0.136 & 0.186 & 0.39 & 1 \\
B77 & 0.148 & 0.210 & 0.37 & 1 \\
Htr & 0.504 & 0.398 & 0.07 & 1 \\
Vcl & 0.563 & 0.325 & 0.06 & 1 \\
Inr & 0.583 & 0.296 & 0.05 & 1 \\
\hline
\end{tabular}
\caption{Initialization diagnostics (pooled over random seeds) and resulting warm-start choice $\tau_0\in\{1,2\}$.}
\label{tab:init_tau0_diag}
\end{table}

\end{appendices}

\end{document}